\documentclass[1p]{elsarticle}

\usepackage{lineno,hyperref}

\biboptions{sort&compress}

\usepackage{array}
\usepackage{subcaption}
\usepackage{algorithmic}
\usepackage{algorithm}
\usepackage{enumitem}
\usepackage{booktabs} 
\usepackage[table]{xcolor}
\usepackage{amssymb}
\usepackage{amsmath}
\usepackage[T1]{fontenc}

\renewcommand*{\today}{03/2020}

\makeatletter
\def\ps@pprintTitle{%
    \let\@oddhead\@empty
    \let\@evenhead\@empty
    \def\@oddfoot{\footnotesize\itshape
         {Preprint accepted by the Journal of Artificial Intelligence Research (JAIR) - extended version} \hfill\today}%
    \let\@evenfoot\@oddfoot
    }
\makeatother

\newcolumntype{x}[1]{>{\centering\arraybackslash\hspace{0pt}}p{#1}}

\DeclareMathOperator*{\argmin}{arg\,min}
\DeclareMathOperator*{\argmax}{arg\,max}

\modulolinenumbers[5]

\journal{the Journal of Artificial Intelligence Research (JAIR)}

\biboptions{authoryear}

\begin{document}

\begin{frontmatter}

\title{Neural Machine Translation: A Review and Survey}

\author{Felix Stahlberg\fnref{myfootnote}}
\address{University of Cambridge, Engineering Department, UK}
\fntext[myfootnote]{Now at Google Research (\href{mailto:fstahlberg@google.com}{fstahlberg@google.com}).}




\begin{abstract}
The field of machine translation (MT), the automatic translation of written text from one natural language into another, has experienced a major paradigm shift in recent years. Statistical MT, which mainly relies on various count-based models and which used to dominate MT research for decades, has largely been superseded by neural machine translation (NMT), which tackles translation with a single neural network. In this work we will trace back the origins of modern NMT architectures to word and sentence embeddings and earlier examples of the encoder-decoder network family. We will conclude with a survey of recent trends in the field.
\end{abstract}


\end{frontmatter}


Various fields in the area of natural language processing (NLP) have been boosted by the rediscovery of neural networks~\citep{nlp-primer}. However, for a long time, the integration of neural nets into machine translation (MT) systems was rather shallow. Early attempts used feedforward neural language models~\citep{nlm,nlm2}\index{Neural language models}\index{Feedforward neural network} for the target language to rerank translation lattices~\citep{nlm-mt}.\index{Lattices} The first neural models which also took the source language into account extended this idea by using the same model with bilingual tuples instead of target language words~\citep{nnjm-tuple}, scoring phrase pairs directly with a feedforward net~\citep{nnjm2}, or adding a source context window\index{Source context} to the neural language model~\citep{nnjm-soul,nnjm-devlin}.\index{Neural joint models} \citet{nnjm-recurrent} and \citet{nnjm-enc-dec} introduced recurrent networks for translation modelling. All those approaches applied neural networks as component in a traditional statistical machine translation system. Therefore, they retained the log-linear model\index{Log-linear models} combination and only exchanged parts in the traditional architecture.

\begin{figure}[tbp!] 
\centering    
\includegraphics[scale=0.68]{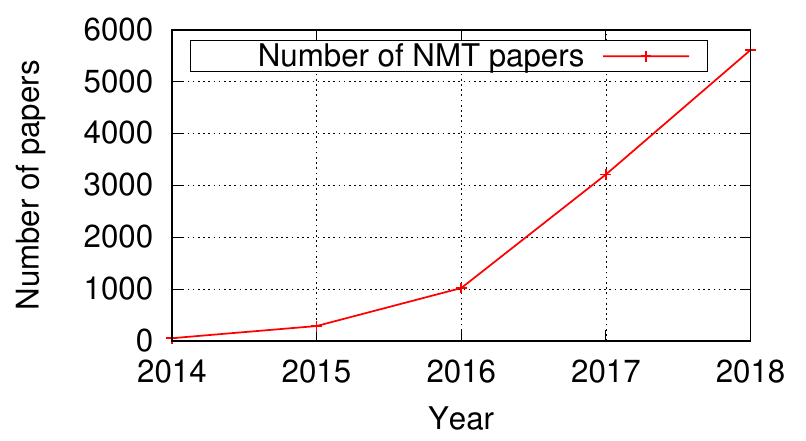}
\caption{Number of papers mentioning ``neural machine translation'' per year according Google Scholar.}
\label{fig:nmt-papers}
\end{figure}

Neural machine translation (NMT) has overcome this separation by using a
single large neural net that directly transforms the source sentence into the target sentence~\citep{nmt-cho,nmt-sutskever,nmt-bahdanau}. The advent of NMT certainly marks one of the major milestones in the history of MT, and has led to a radical and sudden departure of mainstream research from many previous research lines. This is perhaps best reflected by the explosion of scientific publications related to NMT in the past years\footnote{Example Google Scholar search: \url{https://scholar.google.com/scholar?q=\%22neural+machine+translation\%22\&as_ylo=2017\&as_yhi=2017}} (Fig.~\ref{fig:nmt-papers}), and the large number of publicly available NMT toolkits (Tab.~\ref{tab:nmt-toolkits}). NMT has already been widely adopted in the industry~\citep{production-gnmt,production-systran,production-smt-nmt,production-booking} and is deployed in production systems by Google, Microsoft, Facebook, Amazon, SDL, Yandex, and many more. This article will introduce the basic concepts of NMT, and will give a comprehensive overview of current research in the field. For even more insight into the field of neural machine translation, we refer the reader to other overview papers such as \citep{nmt-overview-neubig-tutorial,nmt-overview-recent-trends,nmt-overview-smt-book,nmt-overview-context}.

\begin{table}[t!]
\centering
\footnotesize
\begin{tabular}{l l l l}
\toprule
Name & Citation & Framework &  GitHub\\ 
 &  &  &  Stars \\ 
\midrule
\href{https://github.com/tensorflow/tensor2tensor}{Tensor2Tensor} &	\citet{nmt-tool-t2t} & TensorFlow &	\setlength{\fboxsep}{1pt}\colorbox{gray}{\hspace{1cm}} \\
\href{https://github.com/tensorflow/nmt}{TensorFlow/NMT} &	- & TensorFlow &  \setlength{\fboxsep}{1pt}\colorbox{gray}{\hspace{0.6287262873cm}} \\
\href{https://github.com/pytorch/fairseq}{Fairseq} &	\citet{nmt-tool-fairseq} & PyTorch	& \setlength{\fboxsep}{1pt}\colorbox{gray}{\hspace{0.4337548931cm}} \\
\href{https://github.com/OpenNMT/OpenNMT-py}{OpenNMT-py} &	\citet{nmt-tool-opennmt} & Lua, (Py)Torch, TF	& \setlength{\fboxsep}{1pt}\colorbox{gray}{\hspace{0.3798554652cm}} \\
\href{https://github.com/awslabs/sockeye}{Sockeye} &	\citet{nmt-tool-sockeye} & MXNet	& \setlength{\fboxsep}{1pt}\colorbox{gray}{\hspace{0.1093044264cm}} \\
\href{https://github.com/NVIDIA/OpenSeq2Seq}{OpenSeq2Seq} &	\citet{nmt-tool-openseq2seq} & TensorFlow	& \setlength{\fboxsep}{1pt}\colorbox{gray}{\hspace{0.09153869316cm}} \\
\href{https://github.com/EdinburghNLP/nematus}{Nematus} &	\citet{nmt-tool-nematus} & TensorFlow, Theano	& \setlength{\fboxsep}{1pt}\colorbox{gray}{\hspace{0.07964468534cm}} \\
\href{https://github.com/pytorch/translate}{PyTorch/Translate} &	- & PyTorch	& \setlength{\fboxsep}{1pt}\colorbox{gray}{\hspace{0.06202950918cm}} \\
\href{https://github.com/marian-nmt/marian}{Marian} &	\citet{nmt-tool-marian} & C++	& \setlength{\fboxsep}{1pt}\colorbox{gray}{\hspace{0.05871725384cm}} \\
\href{https://github.com/lvapeab/nmt-keras}{NMT-Keras} &	\citet{nmt-tool-nmtkeras} & TensorFlow, Theano	& \setlength{\fboxsep}{1pt}\colorbox{gray}{\hspace{0.05164107197cm}} \\
\href{https://github.com/ufal/neuralmonkey}{Neural Monkey} &	\citet{nmt-tool-neuralmonkey} & TensorFlow	& \setlength{\fboxsep}{1pt}\colorbox{gray}{\hspace{0.05088828666cm}} \\
\href{https://github.com/THUNLP-MT/THUMT}{THUMT} &	\citet{nmt-tool-thumt} & TensorFlow, Theano	& \setlength{\fboxsep}{1pt}\colorbox{gray}{\hspace{0.04170430593cm}} \\
\href{https://github.com/eske/seq2seq}{Eske/Seq2Seq} &	- & TensorFlow	& \setlength{\fboxsep}{1pt}\colorbox{gray}{\hspace{0.04125263475cm}} \\
\href{https://github.com/neulab/xnmt}{XNMT} &	\citet{nmt-tool-xnmt} & DyNet	& \setlength{\fboxsep}{1pt}\colorbox{gray}{\hspace{0.01776573321cm}} \\
\href{https://github.com/whr94621/NJUNMT-pytorch}{NJUNMT} &	- & PyTorch, TensorFlow	& \setlength{\fboxsep}{1pt}\colorbox{gray}{\hspace{0.01641071966cm}} \\
\href{https://github.com/duyvuleo/Transformer-DyNet}{Transformer-DyNet} &	- & DyNet	& \setlength{\fboxsep}{1pt}\colorbox{gray}{\hspace{0.005420054201cm}} \\
\href{https://github.com/ucam-smt/sgnmt}{SGNMT} &	\citet{sgnmt1,sgnmt2} & TensorFlow, Theano	& \setlength{\fboxsep}{1pt}\colorbox{gray}{\hspace{0.004366154773cm}} \\
\href{https://github.com/arthurxlw/cytonMt}{CythonMT} &	\citet{nmt-tool-cytonmt} & C++	& \setlength{\fboxsep}{1pt}\colorbox{gray}{\hspace{0.001957241795cm}} \\
\href{https://github.com/anoidgit/transformer}{Neutron} &	\citet{nmt-tool-neutron} & PyTorch	& \setlength{\fboxsep}{1pt}\colorbox{gray}{\hspace{0.001957241795cm}} \\
\bottomrule
\end{tabular}
\caption{\label{tab:nmt-toolkits} NMT tools that have been updated in the past year (as of 2019). GitHub stars indicate the popularity of tools on GitHub.}
\end{table}

\section{Nomenclature}
\label{sec:notations}

We will denote the source sentence of length $I$ as $\mathbf{x}$. We use the subscript $i$ to index tokens in the source sentence. We refer to the source language vocabulary as $\Sigma_{src}$.
\begin{equation}
\mathbf{x}=x_1^I=(x_1,\dots, x_I)\in \Sigma_{src}^I
\end{equation}
The translation of source sentence $\mathbf{x}$ into the target language is denoted as $\mathbf{y}$. We use an analogous nomenclature on the target side.
\begin{equation}
\mathbf{y}=y_1^J=(y_1,\dots, y_J)\in \Sigma_{trg}^J
\end{equation}
In case we deal with only one language we drop the subscript $src$/$trg$. For convenience we represent tokens as indices in a list of subwords or word surface forms. Therefore, $\Sigma_{src}$ and $\Sigma_{trg}$ are the first $n$ natural numbers (i.e.\ $\Sigma= \{n'\in\mathbb{N}|n'\leq n\}$ where $n = |\Sigma|$ is the vocabulary size).
Additionally, we use the projection function\index{Projection function|textbf} $\pi_k$ which maps a tuple or vector to its $k$-th entry:
\begin{equation}
\label{eq:project}
\pi_k(z_1,\dots,z_k,\dots,z_n) = z_k.
\end{equation}
For a matrix $A\in\mathbb{R}^{m \times n}$ we denote the element in the $p$-th row and the $q$-th column as $A_{p,q}$, the $p$-th row vector as $A_{p,:}\in\mathbb{R}^n$ and the $q$-th column vector as $A_{:,q}\in\mathbb{R}^m$. For a series of $m$ $n$-dimensional vectors $a_p\in\mathbb{R}^n$ ($p\in [1,m]$) we denote the $m \times n$ matrix which results from stacking the vectors horizontally as $(a_p)_{p=1:m}$ as illustrated with the following tautology:
\begin{equation}
A = (A_{p,:})_{p=1:m} = {((A_{:,q})_{q=1:n})}^T.
\end{equation}

\section{Word Embeddings}
\label{sec:word-embeddings}

Representing words or phrases as continuous vectors is arguably one of the keys in connectionist models for NLP.\index{Natural language processing} To the best of our knowledge, continuous space word representations were first successfully used for language modelling~\citep{embed-first,nlm}. The key idea is to represent a word $x\in \Sigma$ as a $d$-dimensional vector of real numbers. The size $d$ of the embedding layer is normally chosen to be much smaller than the vocabulary size ($d\ll |\Sigma|$) in order to obtain interesting representations. The mapping from the word to its distributed representation can be represented by an embedding matrix\index{Embedding matrix} $E\in\mathbb{R}^{d\times|\Sigma|}$~\citep{embed-nlp}. The $x^{th}$ column of $E$ (denoted as $E_x$) holds the $d$-dimensional representation for the word $x$.

Learned continuous word representations have the potential of capturing morphological,\index{Morphology} syntactic and semantic similarity across words~\citep{embed-nlp}. In neural machine translation, embedding matrices are usually trained jointly with the rest of the network using backpropagation~\citep{nn-backprop} and a gradient based optimizer such as stochastic gradient descent.\index{Gradient-based optimization} In other areas of NLP, pre-trained\index{Pre-training} word embeddings\index{Word embeddings|textbf} trained on unlabelled text have become ubiquitous~\citep{embed-nlp2}. Methods for training word embeddings on raw text often take the context\index{Context} into account in which the word occurs frequently~\citep{embed-glove,embed-word2vec}, or use cross-lingual information to improve embeddings~\citep{embed-bilingual-word-mapping,embed-bilingual-word}.\index{Word embeddings!Cross-lingual word embeddings}

A newly emerging type of {\em contextualized} word embeddings~\citep{embed-context-lstm,embed-context-cv}\index{Word embeddings!Contextualized word embeddings} is gaining popularity in various fields of NLP.\index{Natural language processing}  Contextualized representations do not only depend on the word itself but on the entire input sentence. Thus, they cannot be described by a single embedding matrix but are usually generated by neural sequence models which have been trained under a language model objective.\index{Language models} Most approaches either use LSTM~\citep{embed-context-lstm,embed-context-elmo}\index{LSTM} or Transformer\index{Transformer} architectures~\citep{embed-context-openai,embed-context-bert} but differ in the way these architectures are used to compute the word representations. Contextualized word embeddings have advanced the state-of-the-art in several NLP benchmarks~\citep{embed-context-elmo,embed-context-elmo-friends,embed-context-bert}. \citet{embed-context-bert-goldberg} showed that contextualized embeddings are remarkably sensitive to syntax. \citet{embed-context-bag-nmt} reported gains from contextualizing word embeddings in NMT using a bag of words.\index{Bag of words}

\section{Phrase Embeddings}



\begin{figure}[t!]
  \begin{subfigure}[b]{0.45\textwidth}
    \centering
    \includegraphics[scale=0.43]{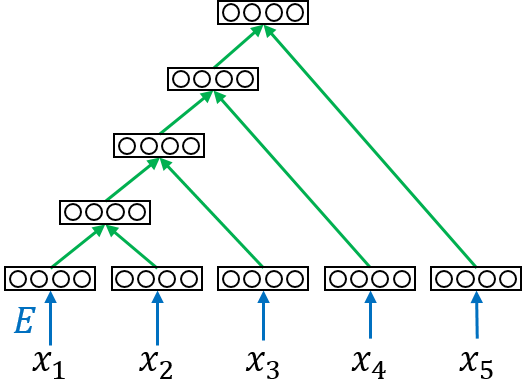}
    \caption{Recursive autoencoder following \citet{embed-recursive-auto}.}
    \label{fig:recursive-autoencoder}
  \end{subfigure}             
  \begin{subfigure}[b]{0.45\textwidth}
    \centering
    \includegraphics[scale=0.43]{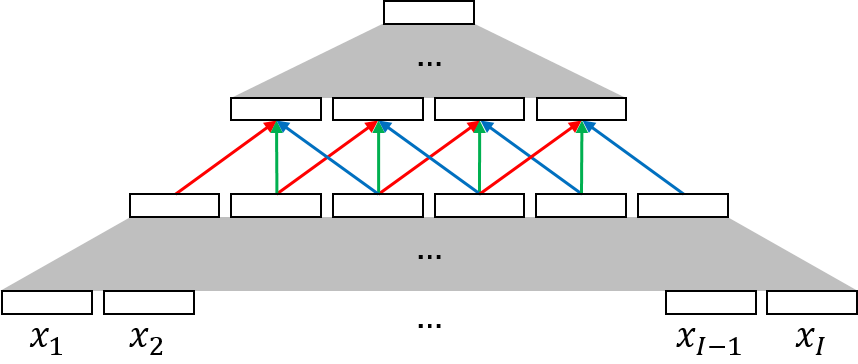}
    \caption{The convolutional sentence model~\citep{nnjm-recurrent}.}
    \label{fig:csm}
  \end{subfigure}             
  \caption{Phrase and sentence embedding architectures. The color coding indicates weight sharing.}
  \label{fig:phrase-and-sentence-embeddings}
\end{figure}

For various NLP\index{Natural language processing} tasks such as sentiment analysis\index{Sentiment analysis} or MT it is desirable to embed whole phrases or sentences instead of single words.\index{Phrase embeddings} For example, a distributed representation of the source sentence $\mathbf{x}$ could be used as conditional for the distribution over the target sentences $P(\mathbf{y}|\mathbf{x})$. Early approaches to phrase embedding were based on recurrent autoencoders~\citep{embed-raam,embed-recursive-auto}.\index{Recurrent autoencoders}\index{Autoencoders!Recurrent autoencoders} To represent a phrase $\mathbf{x}\in\Sigma^I$ as $d$-dimensional vector, \citet{embed-recursive-auto} first trained a word embedding matrix $E\in\mathbb{R}^{d\times|\Sigma|}$.\index{Word embeddings} Then, they recursively applied an autoencoder network which finds $d$-dimensional representations for $2d$-dimensional inputs, where the input is the concatenation of two parent representations. The parent representations are either word embeddings or representations calculated by the same autoencoder\index{Autoencoders} from two different parents. The order in which representations are merged is determined by a binary tree\index{Trees}\index{Trees!Binary trees} over $\mathbf{x}$ which can be constructed greedily~\citep{embed-recursive-auto} or derived from an Inversion Transduction Grammar~\citep[ITG]{fst-itg}~\citep{embed-itg}. Fig.~\ref{fig:recursive-autoencoder}\index{Grammars!Inversion transduction grammars} shows an example of a recurrent autoencoder embedding a phrase with five words into a four dimensional space. One of the disadvantages of recurrent autoencoders is that the word and sentence embeddings need to have the same dimensionality. This restriction is not very critical in sentiment analysis\index{Sentiment analysis} because the sentence representation is only used to extract the sentiment of the writer~\citep{embed-recursive-auto}. In MT, however, the sentence representations need to convey enough information to condition the target sentence distribution on it, and thus should be higher dimensional than the word embeddings.

\section{Sentence Embeddings}
\label{sec:sentence-embeddings}

\citet{nnjm-recurrent} used convolution\index{Convolutional neural networks} to find vector representations of phrases or sentences and thus avoided the dimensionality issue of recurrent autoencoders.\index{Autoencoders} As shown in Fig.~\ref{fig:csm}, their model yields $n$-gram representations at each convolution level, with $n$ increasing with depth. The top level can be used as representation for the whole sentence. Other notable examples of using convolution for sentence representations include \citep{embed-cnn-sentence,embed-cnn-sentence-class,embed-cnn-tree,embed-cnn-short-texts,embed-cnn-att}. However, the convolution operations in these models loose information about the exact word order. and are thus more suitable for sentiment analysis\index{Sentiment analysis} than for tasks like machine translation.\footnote{This is not to be confused with convolutional {\em translation} models which will be reviewed in Sec.~\ref{sec:nmt-cnn}} A recent line of work uses self-attention rather than convolution to find sentence representations~\citep{embed-att-disan,embed-att-phraselevel,embed-att-variational}. Another interesting idea explored by \citet{embed-rn} is to resort to (recursive) relation networks~\citep{nn-rn,nn-recurrent-rn}\index{Relation networks} which repeatedly aggregate pairwise relations between words in the sentence. Recurrent architectures are also commonly used for sentence representation. It has been noted that even random RNNs without any training can work surprisingly well for several NLP\index{Natural language processing} tasks~\citep{embed-sentence-random1,embed-sentence-random2,embed-sentence-random3}.

\section[Encoder-Decoder Networks with Fixed Length Encodings]{Encoder-Decoder Networks with Fixed Length Sentence Encodings}
\label{sec:fixed-length}

\begin{figure}[tbp!]
\centering    
\includegraphics[scale=0.43]{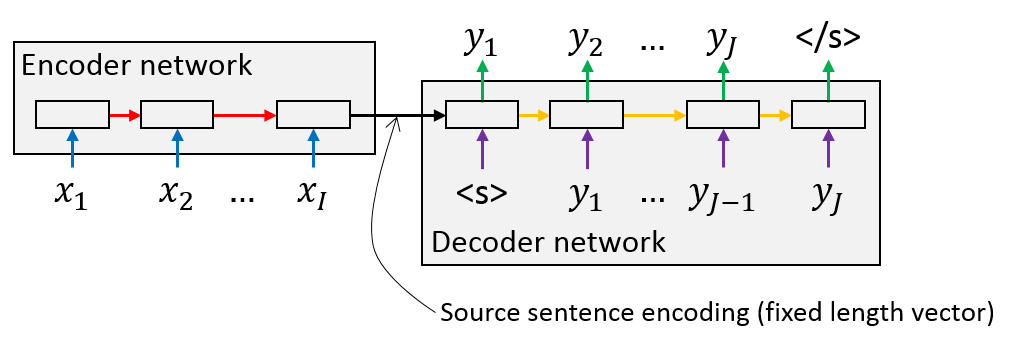}
\caption{The encoder-decoder architecture of~\citet{nmt-sutskever}. The color coding indicates weight sharing.}
\label{fig:encoder-decoder-sutskever}
\end{figure}

\citet{nnjm-recurrent} were the first who conditioned the target sentence distribution on a distributed fixed-length representation\index{Fixed-length representation|textbf} of the source sentence. Their recurrent continuous translation models\index{Recurrent continuous translation models} (RCTM) I and II gave rise to a new family of so-called encoder-decoder networks\index{Encoder-decoder networks|textbf} which is the current prevailing architecture for NMT. Encoder-decoder networks are subdivided into an encoder network which computes a representation of the source sentence, and a decoder\index{Decoder} network which generates the target sentence from that representation. As introduced in Sec.~\ref{sec:notations} we denote the source sentence as $\mathbf{x}=x_1^I$ and the target sentence as $\mathbf{y}=y_1^J$. All existing NMT models define a probability distribution over the target sentences $P(\mathbf{y}|\mathbf{x})$ by factorizing it into conditionals:\index{Left-to-right factorization|textbf}
\begin{equation}
\label{eq:nmt-factorization}
P(\mathbf{y}|\mathbf{x})\overset{\text{Chain rule}}{=}\prod_{j=1}^J P(y_j|y_1^{j-1},\mathbf{x}).
\end{equation}
Different encoder-decoder architectures differ vastly in how they model the distribution $P(y_j|y_1^{j-1},\mathbf{x})$. We will first discuss encoder-decoder networks in which the encoder represents the source sentence as a fixed-length vector $c(\mathbf{x})$ like the methods in Sec.~\ref{sec:sentence-embeddings}. The conditionals $P(y_j|y_1^{j-1},\mathbf{x})$ are modelled as:
\begin{figure}[t!]
  \centering
  \begin{subfigure}[b]{0.45\textwidth}
    \centering
    \includegraphics[scale=0.43]{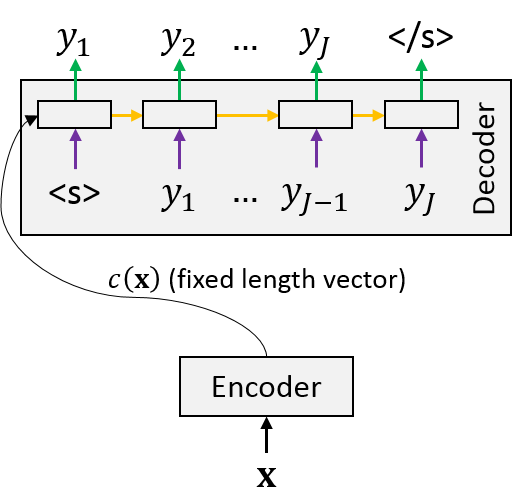}
    \caption{Source sentence is used to initialize the decoder state.\index{Decoder state}}
    \label{fig:fixed-length-init}
  \end{subfigure}             
  \begin{subfigure}[b]{0.45\textwidth}
    \centering
    \includegraphics[scale=0.43]{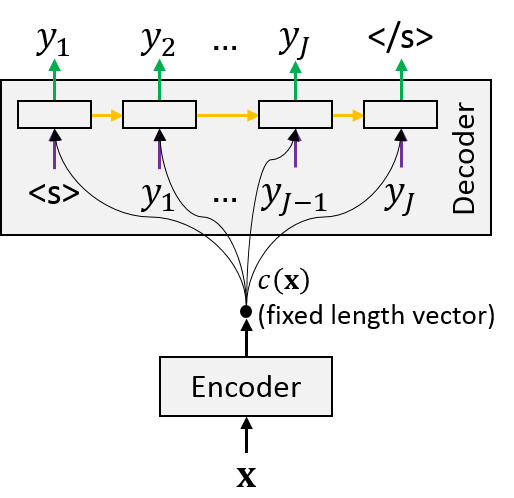}
    \caption{Source sentence is fed to the decoder at each time step.}
    \label{fig:fixed-length-all}
  \end{subfigure}             
  \caption{Encoder-decoder architectures with fixed-length sentence encodings. The color coding indicates weight sharing.}
  \label{fig:fixed-length-enc-dec}
\end{figure}
\begin{equation}
\label{eq:nmt-fixed-length}
P(y_j|y_1^{j-1},\mathbf{x}) = g(y_j|s_j,y_{j-1}, c(\mathbf{x}))
\end{equation}
where $s_j$ is the hidden state of a recurrent neural (decoder)\index{Decoder} network (RNN).\index{RNN} We will formally introduce $s_j$ in Sec.~\ref{sec:nmt-att-recurrent}. Gated activation functions\index{Activation function}\index{Gated activation} such as the long short-term memory~\citep[LSTM]{nn-lstm}\index{LSTM} or the gated recurrent unit \citep[GRU]{nnjm-enc-dec}\index{GRU} are commonly used to alleviate the vanishing gradient problem~\citep{nn-vanishing-gradient}\index{Vanishing gradient problem} which makes it difficult to train RNNs to capture long-range dependencies. Deep architectures with stacked LSTM cells were used by~\citet{nmt-sutskever}. The encoder can be a convolutional network as in the RCTM I~\citep{nnjm-recurrent}, an LSTM network~\citep{nmt-sutskever}, or a GRU network~\citep{nnjm-enc-dec}. $g(\cdot)$ is a feedforward network\index{Feedforward neural network} with a softmax layer\index{Softmax} at the end which takes as input the decoder state $s_j$\index{Decoder state} and an embedding of the previous target token $y_{j-1}$. In addition, $g(\cdot)$ may also take the source sentence encoding $c(\mathbf{x})$ as input to condition on the source sentence~\citep{nnjm-recurrent,nnjm-enc-dec}. Alternatively, $c(\mathbf{x})$ is just used to initialize the decoder state\index{Decoder state} $s_1$~\citep{nmt-sutskever,nmt-bahdanau}. Fig.~\ref{fig:fixed-length-enc-dec} contrasts both methods. Intuitively, once the source sentence has been encoded, the decoder starts generating the first target sentence symbol $y_1$ which is then fed back\index{Feedback loop} to the decoder\index{Decoder} network for producing the second symbol $y_2$. The algorithm terminates when the network produces the end-of-sentence symbol\index{End-of-sentence symbol} $\text{</s>}$. Sec.~\ref{sec:nmt-decoding} explains more formally what we mean by the network ``generating'' a symbol $y_j$ and sheds more light on the aspect of decoding in NMT. Fig.~\ref{fig:encoder-decoder-sutskever} shows the complete architecture of~\citet{nmt-sutskever} who presented one of the first working standalone NMT systems that did not rely on any SMT baseline. One of the reasons why this paper was groundbreaking is the simplicity of the architecture, which stands in stark contrast to traditional SMT systems that used a very large number of highly engineered features.

\begin{figure}[t!]
  \centering
  \begin{subfigure}[t]{1.0\textwidth}
    \centering
    \includegraphics[scale=0.43]{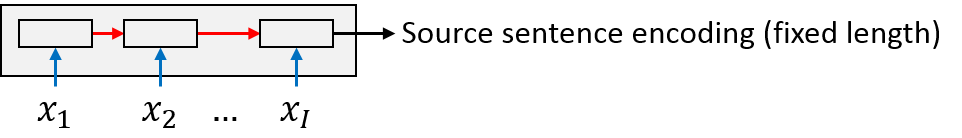}
    \caption{Unidirectional encoder used by~\citet{nnjm-enc-dec}.}
    \label{fig:unidirectional-encoder}
  \end{subfigure}             
  \begin{subfigure}[t]{1.0\textwidth}
    \centering
    \includegraphics[scale=0.43]{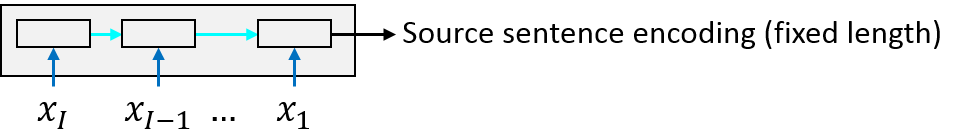}
    \caption{Reversed unidirectional encoder from~\citet{nmt-sutskever}.}
    \label{fig:unidirectional-encoder-reversed}
  \end{subfigure}             
  \begin{subfigure}[b]{1.0\textwidth}
    \centering
    \includegraphics[scale=0.43]{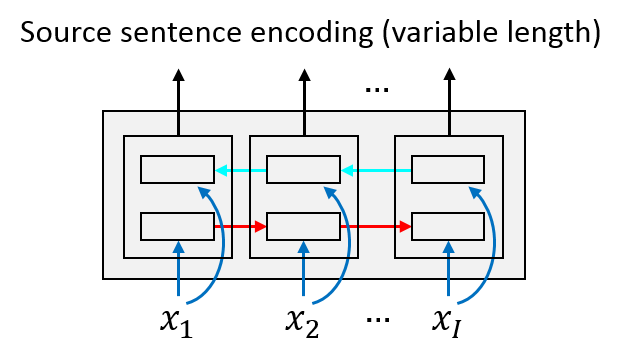}
    \caption{Bidirectional encoder used by~\citet{nmt-bahdanau}.}
    \label{fig:bidirectional-encoder}
  \end{subfigure}
  \caption{Encoder\index{Encoder} architectures. The color coding indicates weight sharing.}
  \label{fig:encoders}
\end{figure}

Different ways of providing the source sentence to the encoder\index{Encoder} network have been explored in the past. \citet{nnjm-enc-dec} fed the tokens to the encoder in the natural order they appear in the source sentence (cf.\ Fig.~\ref{fig:unidirectional-encoder}).\index{RNN!Unidirectional RNNs} \citet{nmt-sutskever} reported gains from simply feeding the sequence in reversed order (cf.\ Fig.~\ref{fig:unidirectional-encoder-reversed}). They argue that these improvements might be ``caused by the introduction of many short term dependencies to the dataset''~\citep{nmt-sutskever}. Bidirectional RNNs~\citep[BiRNN]{nn-birnn}\index{RNN!Bidirectional RNNs} are able to capture both directions (cf.\ Fig.~\ref{fig:bidirectional-encoder}) and are often used in attentional NMT~\citep{nmt-bahdanau}.

\section{Attentional Encoder-Decoder Networks}

\subsection{Attention}
\label{sec:attention}

\index{Encoder-decoder networks}One problem of early NMT models which is still not fully solved yet (see Sec.~\ref{sec:sentence-length}) is that they often produced poor translations for long sentences~\citep{nmt-lengthbias}.\index{Sentence length} \citet{nmt-cho} suggested that this weakness is due to the fixed-length source sentence encoding.\index{Fixed-length representation} Sentences with varying length convey different amounts of information. Therefore, despite being appropriate for short sentences, a fixed-length vector ``does not have enough capacity to encode a long sentence with complicated structure and meaning''~\citep{nmt-cho}.  \citet{nmt-chopping} tried to mitigate this problem by chopping the source sentence into short clauses. They composed the target sentence by concatenating the separately translated clauses. However, this approach does not cope well with long-distance reorderings as word reorderings are only possible within a clause. \citet{nmt-bahdanau} introduced the concept of {\em attention}\index{Attention|textbf} to avoid having a fixed-length source sentence representation. Their model does not use a constant context vector\index{Source context} $c(\mathbf{x})$ any more which encodes the whole source sentence. By contrast, the attentional decoder\index{Decoder} can place its attention only on parts of the source sentence which are useful for producing the next token. The constant context vector $c(\mathbf{x})$ is thus replaced by a series of context vectors $c_j(\mathbf{x})$; one for each time step~$j$.\footnote{We refer to $j$ as `time step' due to the sequential structure of autoregressive models and the left-to-right order of NMT decoding. We note, however, that $j$ does not specify a point in time in the usual sense but rather the position in the target sentence.}

We will first introduce attention as a general concept before describing the architecture of \citet{nmt-bahdanau} in detail in Sec.~\ref{sec:nmt-att-recurrent}. We follow the terminology of \citet{nmt-transformer} and describe attention as mapping $n$ query vectors to $n$ output vectors via a mapping table (or a {\em memory})\index{Memory} of $m$ key-value pairs. This view is related to memory-augmented neural networks which we will discuss in greater detail in Sec.~\ref{sec:memory-networks}. We make the simplifying assumption that all vectors have the same dimension $d$ so that we can stack the vectors into matrices $Q\in \mathbb{R}^{n\times d}$, $K\in \mathbb{R}^{m\times d}$, and $V\in \mathbb{R}^{m\times d}$. Intuitively, for each query vector we compute an output vector as a weighted sum of the value vectors. The weights are determined by a similarity score between the query vector and the keys (cf.\ \citep[Eq.\ 1]{nmt-transformer}):
\begin{equation}
\underbrace{\text{Attention}(K, V, Q)}_{n\times d} = \text{Softmax}(\underbrace{\text{score}(Q,K)}_{n\times m})\underbrace{V}_{m\times d}.
\end{equation}
The output of $\text{score}(Q,K)$ is an $n \times m$ matrix of similarity scores.\index{Attention!Scoring function} The softmax\index{Softmax} function normalizes over the columns of that matrix so that the weights for each query vector sum up to one. A straight-forward choice for $\text{score}(\cdot)$ proposed by \citet{nmt-dot-product-att} is the dot product (i.e.\ $\text{score}(Q,K)=QK^\intercal$).\index{Attention!Dot-product} The most common scoring functions are summarized in Tab.~\ref{tab:attention-functions}.

\begin{table}
\centering
\footnotesize
\begin{tabular}{l l l}
\toprule
Name & Scoring function & Citation \\ 
\midrule
Additive\index{Attention!Additive} & $\text{score}(Q,K)_{p,q}=v^\intercal \tanh(WQ_{p,:} + UK_{q,:})$ & \citet{nmt-bahdanau}  \\
Dot-product\index{Attention!Dot-product} & $\text{score}(Q,K)=QK^\intercal$ & \citet{nmt-dot-product-att}  \\
Scaled dot-product\index{Attention!Scaled dot-product} & $\text{score}(Q,K)=QK^\intercal d^{-0.5}$ & \citet{nmt-transformer}  \\
\bottomrule
\end{tabular}
\caption{\label{tab:attention-functions} Common attention scoring functions. $v\in \mathbb{R}^{d_\text{att}}$, $W\in \mathbb{R}^{d_\text{att}\times d}$, and $U\in \mathbb{R}^{d_\text{att}\times d}$ in additive attention are trainable parameters with $d_\text{att}$ being the dimensionality of the attention layer.}
\end{table}

A common way to use attention in NMT is at the interface between encoder\index{Encoder} and decoder.\index{Decoder} \citet{nmt-bahdanau,nmt-dot-product-att} used the hidden decoder states\index{Decoder state} $s_j$ as query vectors. Both the key and value vectors are derived from the hidden states $h_i$ of a recursive encoder.\footnote{$s_j$ and $h_i$ are defined in Sec.~\ref{sec:fixed-length} and Sec.~\ref{sec:nmt-att-recurrent}.} Formally, this means that  $Q=s_j$ are the query vectors , $n=J$ is the target sentence length, $K=V=h_i$ are the key and value vectors, and $m=I$ is the source sentence length.\footnote{An exception is the model of \citet{nmt-att-key-value} that splits $h_i$ into two parts and uses the first part as key and the second as value.} The outputs of the attention layer are used as time-dependent context vectors $c_j(\mathbf{x})$. In other words, rather than using a fixed-length sentence encoding $c(\mathbf{x})$ as in Sec.~\ref{sec:fixed-length}, at each time step $j$ we query a memory\index{Memory} in which entries store (context-sensitive)\index{Context} representations of the source words. In this setup it is possible to derive an attention matrix $A\in \mathbb{R}^{J\times I}$\index{Attention!Attention weight matrix} to visualize the learned relations between words in the source sentence and words in the target sentence:
\begin{equation}
A:= \text{Softmax}(\text{score}((s_j)_{j=1:J},(h_i)_{i=1:I})).
\end{equation}
Fig.~\ref{fig:attention-example} shows an example of $A$ from an English-German NMT system with additive attention. The attention matrix captures cross-lingual word relationships such as $\text{``is''}\rightarrow\text{``ist''}$ or $\text{``great''}\rightarrow\text{``großer''}$. The system has learned that the English source word ``is'' is relevant for generating the German target word ``ist'' and thus emits a high attention weight for this pair. Consequently, the context vector $c_j(\mathbf{x})$ at time step $j=3$ mainly represents the source word ``is'' ($c_3(\mathbf{x}) \approx h_2$). This is particularly significant as the system was not explicitly trained to align words  but to optimize translation performance. However, as we will argue in Sec.~\ref{sec:hard-attention}, it would be wrong to think of $A$ as a soft version of a traditional SMT word alignment\index{Alignments!Soft alignment}.

\begin{figure}[tbp!]
\centering
\small
\begin{tabular}{| l | x{10mm} x{10mm} x{10mm} x{10mm} x{11mm} x{10mm} x{10mm} |} \hline
 & history & is & a & great & teacher & . & $\text{</s>}$ \\ \hline
die & \cellcolor{blue!75.386834} & \cellcolor{blue!0.384868} & \cellcolor{blue!0.984385} & \cellcolor{blue!0.735846} & \cellcolor{blue!1.713178} & \cellcolor{blue!0.512879} & \cellcolor{blue!20.282011} \\
Geschichte & \cellcolor{blue!89.520002} & \cellcolor{blue!0.482775} & \cellcolor{blue!1.258595} & \cellcolor{blue!1.520045} & \cellcolor{blue!2.247319} & \cellcolor{blue!0.280355} & \cellcolor{blue!4.690909} \\
ist & \cellcolor{blue!0.766737} & \cellcolor{blue!88.627211} & \cellcolor{blue!3.569861} & \cellcolor{blue!0.482497} & \cellcolor{blue!0.535995} & \cellcolor{blue!1.752184} & \cellcolor{blue!4.265515} \\
ein & \cellcolor{blue!0.454577} & \cellcolor{blue!5.041393} & \cellcolor{blue!77.570467} & \cellcolor{blue!13.786621} & \cellcolor{blue!0.340361} & \cellcolor{blue!1.776967} & \cellcolor{blue!1.029613} \\
großer & \cellcolor{blue!0.00} & \cellcolor{blue!0.7} & \cellcolor{blue!0.406551143} & \cellcolor{blue!92.347939} & \cellcolor{blue!0.992106} & \cellcolor{blue!0.255438915} & \cellcolor{blue!1.5516974} \\
Lehrer & \cellcolor{blue!0.072973978} & \cellcolor{blue!0.5419294} & \cellcolor{blue!1.21056767} & \cellcolor{blue!13.2688512} & \cellcolor{blue!84.2249539} & \cellcolor{blue!0.0539773520} & \cellcolor{blue!0.62674640} \\
. & \cellcolor{blue!3.234488} & \cellcolor{blue!29.144532} & \cellcolor{blue!7.628806} & \cellcolor{blue!2.561437} & \cellcolor{blue!3.927034} & \cellcolor{blue!46.445488} & \cellcolor{blue!7.058215} \\
$\text{</s>}$ & \cellcolor{blue!30.408343} & \cellcolor{blue!0.746102} & \cellcolor{blue!0.27925} & \cellcolor{blue!0.323479} & \cellcolor{blue!6.243327} & \cellcolor{blue!9.513191} & \cellcolor{blue!52.486308} \\ \hline
\end{tabular}
\caption{Attention weight matrix $A$ for the translation from the English sentence ``history is a great teacher .'' to the German sentence ``die Geschichte ist ein großer Lehrer .''. Dark shades of blue indicate high attention weights.}
\label{fig:attention-example}
\end{figure}


\begin{figure}[t!] 
\centering    
\includegraphics[scale=0.43]{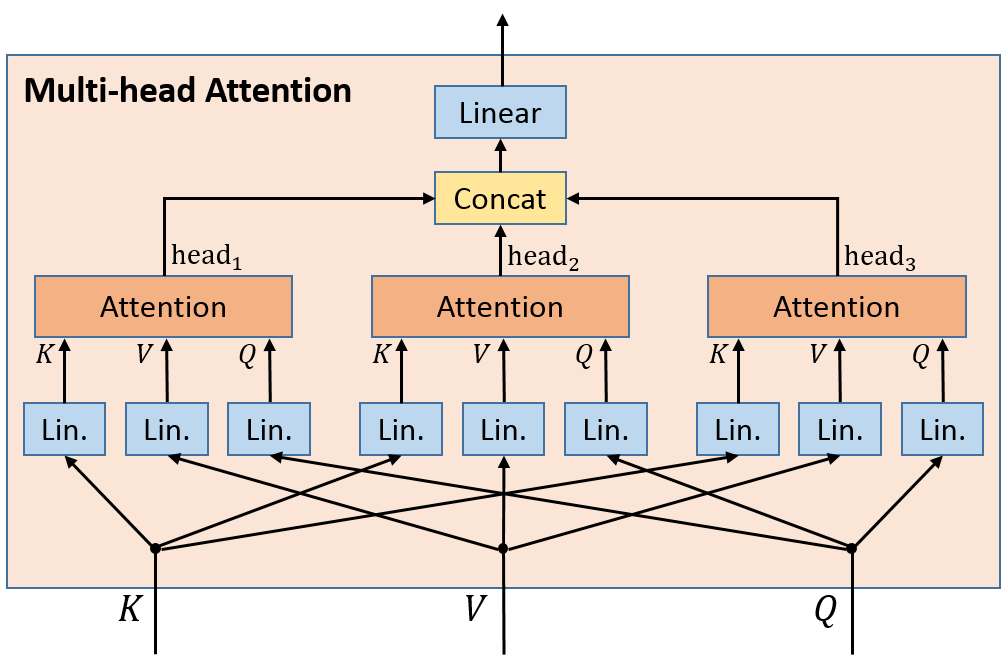}
\caption{Multi-head attention with three attention heads.}
\label{fig:multi-head-att}
\end{figure}

An important generalization of attention is {\em multi-head} attention\index{Attention!Multi-head attention|textbf} proposed by~\citet{nmt-transformer}. The idea is to perform $H$ attention operations instead of a single one where $H$ is the number of attention heads\index{Attention heads} (usually $H=8$). The query, key, and value vectors for the attention heads are linear transforms of $Q$, $K$, and $V$. The output of multi-head attention is the concatenation of the outputs of each attention head. The dimensionality of the attention heads is usually divided by $H$ to avoid increasing the number of parameters. Formally, it can be described as follows~\citep{nmt-transformer}:
\begin{equation}
\text{MultiHeadAttention}(K,V,Q) = \text{Concat}(\text{head}_1,\dots,\text{head}_H)W^O
\end{equation}
with weight matrix $W^O\in\mathbb{R}^{d\times d}$ where
\begin{equation}
\text{head}_h = \text{Attention}(KW^K_h,VW^V_h,QW^Q_h)
\end{equation}
with weight matrices $W^K_h,W^V_h,W^Q_h\in\mathbb{R}^{d\times\frac{d}{H}}$ for $h\in[1,H]$. Fig.~\ref{fig:multi-head-att} shows a multi-head attention module with three heads. Note that with multi-head attention it is not obvious anymore how to derive a single attention weight matrix $A$\index{Attention!Attention weight matrix} like shown in Fig.~\ref{fig:attention-example}. Therefore, models using multi-head attention tend to be more difficult to interpret.

 The concept of attention is no longer just a technique to improve sentence lengths in NMT. Since its introduction by \citet{nmt-bahdanau} it has become a vital part of various NMT architectures, culminating in the Transformer\index{Transformer} architecture (Sec.~\ref{sec:transformer}) which is entirely attention-based. Attention has also been proven effective for, inter alia, object recognition~\citep{att-hinton,att-object-recognition1,att-object-recognition2}, image caption generation~\citep{att-image-caption}, video description~\citep{att-video-description}, speech recognition~\citep{att-speech,att-spell},\index{Speech recognition} cross-lingual word-to-phone alignment~\citep{att-word2phone}, bioinformatics~\citep{att-bioinformatics}, text summarization~\citep{att-summarization}, text normalization~\citep{neural-text-norm},\index{Text normalization} grammatical error correction~\citep{gec-nmt}, question answering~\citep{att-qa,att-qa-image,att-qa-mem}, natural language understanding and inference~\citep{att-nlu,embed-att-disan,att-nli-self,att-nli}, uncertainty detection~\citep{att-heike-uncertainty}, photo optical character recognition~\citep{att-ocr}, and natural language conversation~\citep{att-conversation}.

\subsection{Attention Masks and Padding}
\label{sec:mask-attention}

\begin{figure}[tbp!]
\centering
\small
\begin{tabular}{|x{15mm} x{15mm} x{15mm} x{15mm} x{15mm} x{15mm} |} \hline
\cellcolor{green!40.0} the & \cellcolor{green!40.0} first & \cellcolor{green!40.0} cold & \cellcolor{green!40.0} shower & \cellcolor{red!40.0} $\text{<pad>}$ & \cellcolor{red!40.0} $\text{<pad>}$ \\
\cellcolor{green!40.0} even & \cellcolor{green!40.0} the & \cellcolor{green!40.0} monkey & \cellcolor{green!40.0} seems & \cellcolor{green!40.0} to & \cellcolor{green!40.0} want \\
\cellcolor{green!40.0} a & \cellcolor{green!40.0} little & \cellcolor{green!40.0} coat & \cellcolor{green!40.0} of & \cellcolor{green!40.0} straw & \cellcolor{red!40.0} $\text{<pad>}$ \\
\hline
\end{tabular}
\caption{A tensor containing a batch of three source sentences of different lengths (``the first cold shower'', ``even the monkey seems to want'', ``a little coat of straw'' -- a haiku by Basho~\citep{misc-basho}). Short sentences are padded with $\text{<pad>}$. The training loss and attention masks are visualized with green (enabled) and red (disabled) background.}
\label{fig:mask-attention}
\end{figure}

NMT usually groups sentences into batches to make more efficient use of the available hardware and to reduce noise in gradient estimation (cf.\ Sec.~\ref{sec:xent}). However, the central data structure for many machine learning frameworks~\citep{nn-theano,nn-tensorflow} are {\em tensors} -- multi-dimensional arrays with fixed dimensionality. Re-arranging source sentences as tensor often results in some unused space as the sentences may vary in length. In practice, shorter sentences are filled up with a special padding\index{Padding|textbf} symbol $\text{<pad>}$ to match the length of the longest sentence in the batch (Fig.~\ref{fig:mask-attention}). Most implementations work with masks to avoid taking padded positions into account when computing the training loss. Attention layers also have to be restricted to non-padding symbols which is also usually realized by multiplying the attention weights by a mask that sets the attention weights for padding symbols to zero.\index{Attention!Masked attention|textbf}

\subsection{Recurrent Neural Machine Translation}
\label{sec:nmt-att-recurrent}

\begin{figure}[t!] 
\centering    
\includegraphics[scale=0.43]{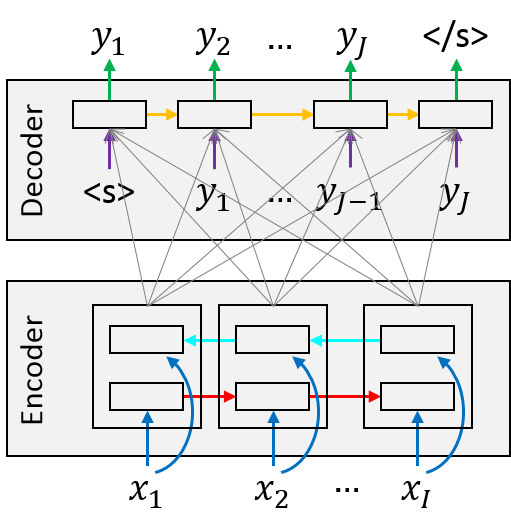}
\caption{The RNNsearch model following~\citet{nmt-bahdanau}. The color coding indicates weight sharing. Gray arrows represent attention.}
\label{fig:bahdanau-encoder-decoder}
\end{figure}

This section contains a complete formal description of the RNNsearch\index{RNNsearch} architecture of \citet{nmt-bahdanau} which was the first NMT model using attention. Recall that NMT uses the chain rule\index{Chain rule} to decompose the probability $P(\mathbf{y}|\mathbf{x})$ of a target sentence $\mathbf{y}=y_1^J$ given a source sentence $\mathbf{x}=x_1^I$ into left-to-right conditionals\index{Left-to-right factorization} (Eq.~\ref{eq:nmt-factorization}). RNNsearch models the conditionals as follows~\citep[Eq.~2,4]{nmt-bahdanau}:
\begin{equation}
P(\mathbf{y}|\mathbf{x})\overset{\text{Eq.~\ref{eq:nmt-factorization}}}{=}\prod_{j=1}^J P(y_j|y_1^{j-1},\mathbf{x})=\prod_{j=1}^J g(y_j|y_{j-1},s_j,c_j(\mathbf{x})).
\label{eq:nmt-bahdanau}
\end{equation}
Similarly to Eq.~\ref{eq:nmt-fixed-length}, the function $g(\cdot)$ encapsulates the decoder\index{Decoder} network which computes the distribution for the next target token $y_j$ given the last produced token $y_{j-1}$, the RNN decoder state\index{Decoder state} $s_j\in\mathbb{R}^n$, and the context vector $c_j(\mathbf{x})\in\mathbb{R}^m$. The sizes of the encoder\index{Encoder} and decoder hidden layers are denoted with $m$ and $n$. The context vector $c_j(\mathbf{x})$ is a distributed representation of the relevant parts of the source sentence. In NMT without attention~\citep{nmt-sutskever,nnjm-enc-dec} (Sec.~\ref{sec:fixed-length}), the context vector is constant and thus needs to encode the whole source sentence. Adding an attention\index{Attention} mechanism results in different context vectors for each target sentence position $j$. This effectively addresses issues in NMT due to the limited capacity of a fixed context vector\index{Fixed-length representation} as illustrated in Fig.~\ref{fig:bahdanau-encoder-decoder}.

As outlined in Sec.~\ref{sec:attention}, the context vectors $c_j(\mathbf{x})$ are weighted sums of source sentence {\em annotations} $\mathbf{h}=(h_1,\dots,h_I)$. The annotations are produced by the encoder\index{Encoder} network. In other words, the encoder converts the input sequence $\mathbf{x}$ to a sequence of annotations $\mathbf{h}$ of the same length. Each annotation $h_i\in\mathbb{R}^m$ encodes information about the entire source sentence $\mathbf{x}$ ``with a strong focus on the parts surrounding the $i$-th word of the input sequence''~\citep[Sec.~3.1]{nmt-bahdanau}. RNNsearch uses a bidirectional RNN~\citep[BiRNN]{nn-birnn} to generate the annotations.\index{RNN!Bidirectional RNNs|textbf} A BiRNN consists of two independent RNNs. The forward RNN $\overset{\rightarrow}{f}$ reads $\mathbf{x}$ in the original order (from $x_1$ to $x_I$). The backward RNN $\overset{\leftarrow}{f}$ consumes $\mathbf{x}$ in reversed order (from $x_I$ to $x_1$):
\begin{equation}
\overset{\rightarrow}{h}_i = \overset{\rightarrow}{f}(x_i, \overset{\rightarrow}{h}_{i-1})
\end{equation} 
\begin{equation}
\overset{\leftarrow}{h}_i = \overset{\leftarrow}{f}(x_{i}, \overset{\leftarrow}{h}_{i+1}).
\end{equation}
The RNNs $\overset{\rightarrow}{f}(\cdot)$ and $\overset{\leftarrow}{f}(\cdot)$ are usually LSTM~\citep{nn-lstm}\index{LSTM} or GRU~\citep{nnjm-enc-dec}\index{GRU} cells. The annotation\index{Annotations} $h_i$ is the concatenation of the hidden states $\overset{\rightarrow}{h}_i$ and $\overset{\leftarrow}{h}_i$~\citep[Sec.~3.2]{nmt-bahdanau}:
\begin{equation}
h_i = [\overset{\rightarrow}{h}_i^\intercal; \overset{\leftarrow}{h}_i^\intercal]^\intercal.
\end{equation}
The context vectors $c_j(\mathbf{x})\in \mathbb{R}^m$ are computed from the annotations as weighted sum with weights $\mathbf{\alpha}_j\in [0,1]^I$~\citep[Eq.~5]{nmt-bahdanau}:
\begin{equation}
\label{eq:nmt-c}
c_j(\mathbf{x}) = \sum_{i=1}^I \alpha_{j,i}  h_i.
\end{equation}
The weights are determined by the alignment model $a(\cdot)$:
\begin{equation}
\label{eq:nmt-alpha}
\alpha_{j,i} = \frac{1}{Z} \exp(a(s_{j-1},h_i))\text{ with }Z=\sum_{k=1}^I \exp(a(s_{j-1},h_k))
\end{equation}
where $a(s_{j-1},h_i)$ is a feedforward neural network\index{Feedforward neural network} which estimates the importance of annotation $h_i$ for producing the $j$-th target token given the current decoder state\index{Decoder state} $s_{j-1}\in \mathbb{R}^n$. In the terminology of Sec.~\ref{sec:attention}, $h_i$ represent the keys and values, $s_j$ are the queries, and $a(\cdot)$ is the attention scoring function.

\begin{figure}[t!] 
\centering    
\includegraphics[scale=0.4]{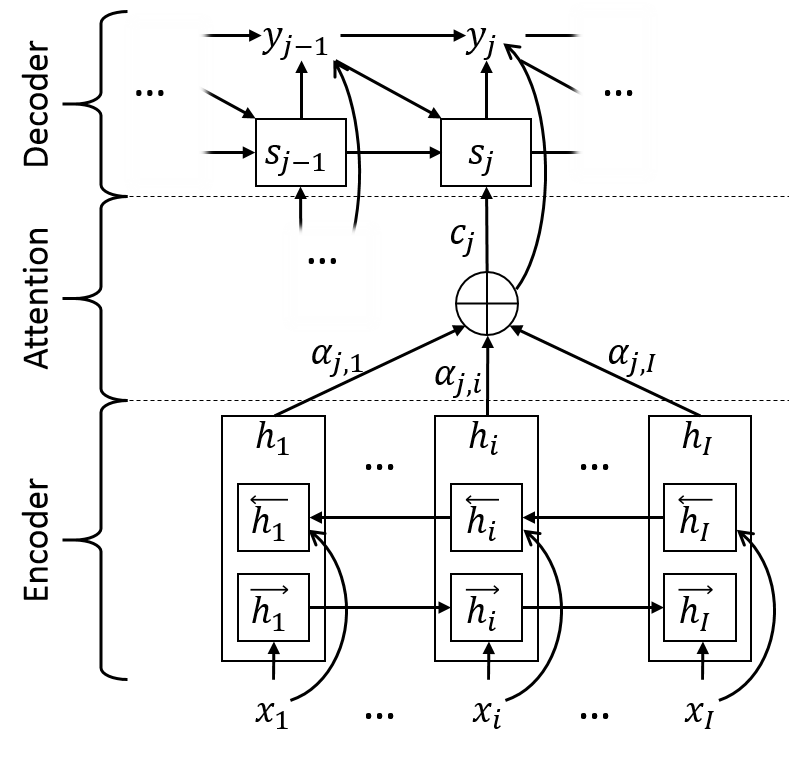}
\caption{Illustration of the attention mechanism in RNNsearch~\citep{nmt-bahdanau}.}
\label{fig:rnnsearch}
\end{figure}

The function $g(\cdot)$ in Eq.~\ref{eq:nmt-bahdanau} does not only take the previous target token $y_{j-1}$ and the context vector $c_j$ but also the decoder hidden state $s_j$.
\begin{equation}
\label{eq:nmt-s}
s_j = f(s_{j-1}, y_{j-1}, c_j)
\end{equation}
where $f(\cdot)$ is modelled by a GRU\index{GRU} or LSTM\index{LSTM} cell. The function $g(\cdot)$ is defined as follows.

\begin{equation}
g(y_j|y_{j-1},s_j,c_j)\propto \exp(W_o  \max(t_j,u_j))
\end{equation}
with
\begin{equation}
t_j = T_s s_{j}+T_y E y_{j-1} + T_c c_j
\end{equation}
\begin{equation}
u_j = U_s s_{j}+U_y E y_{j-1} + U_c c_j
\end{equation}
where $\max(\cdot)$ is the {\em element-wise} maximum, and $W_o\in\mathbb{R}^{|\Sigma_{trg}|\times l}$, $T_s,U_s\in\mathbb{R}^{l\times n}$, $T_y,U_y\in\mathbb{R}^{l\times k}$, $E\in\mathbb{R}^{k\times |\Sigma_{trg}|}$, $T_c,U_c\in\mathbb{R}^{l\times m}$ are weight matrices. The definition of $g(\cdot)$ can be seen as connecting the output of the recurrent layer, an $k$-dimensional embedding of the previous target token, and the context vector with a single maxout layer~\citep{nn-maxout}\index{Maxout layer} of size $l$ and using a softmax\index{Softmax} over the target language vocabulary~\citep{nmt-bahdanau}. Fig.~\ref{fig:rnnsearch} illustrates the complete RNNsearch\index{RNNsearch} model.

\subsection{Convolutional Neural Machine Translation}
\label{sec:nmt-cnn}

Although convolutional neural networks (CNNs)\index{Convolutional neural networks|textbf} have first been proposed by \citet{nn-tdnn} for phoneme recognition, their traditional use case is computer vision~\citep{nn-cnn-zipcode,nn-cnn-mnist,nn-cnn-document}.\index{Computer vision} CNNs are especially useful for processing images because of two reasons. First, they use a high degree of weight tying and thus reduce the number of parameters dramatically compared to fully connected networks. This is crucial for high dimensional input like visual imagery. Second, they automatically learn space invariant features. Spatial invariance\index{Spatial invariance}\index{Space invariance} is desirable in vision since we often aim to recognize objects or features regardless of their exact position in the image. In NLP,\index{Natural language processing} convolutions are usually one dimensional since we are dealing with sequences rather than two dimensional images as in computer vision. We will therefore limit our discussions to the one dimensional case. We will also exclude concepts like pooling or strides as they are uncommon for sequence models in NLP.

\begin{figure}[t!]
  \centering
  \begin{subfigure}[b]{0.31\textwidth}
    \centering
    \includegraphics[scale=0.43]{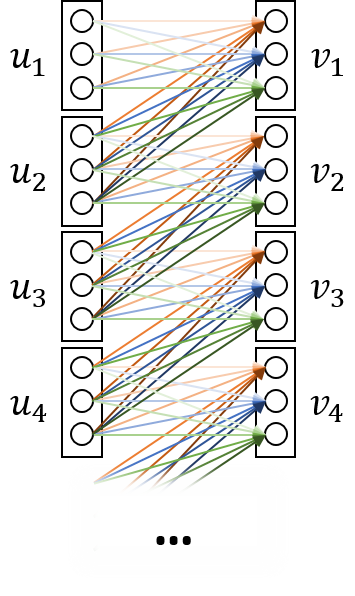}
    \caption{Standard convolution.}
    \label{fig:conv-std}
  \end{subfigure}             
  \begin{subfigure}[b]{0.31\textwidth}
    \centering
    \includegraphics[scale=0.43]{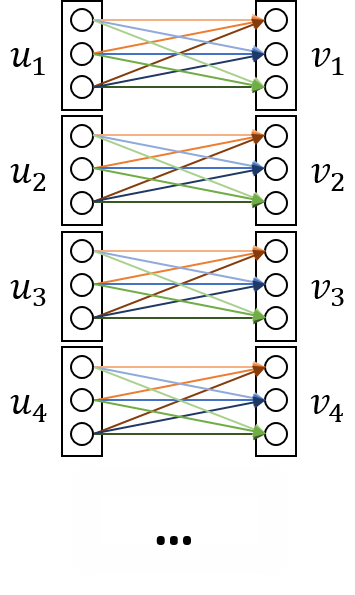}
    \caption{Pointwise convolution.}
    \label{fig:conv-pointwise}
  \end{subfigure}             
  \begin{subfigure}[b]{0.31\textwidth}
    \centering
    \includegraphics[scale=0.43]{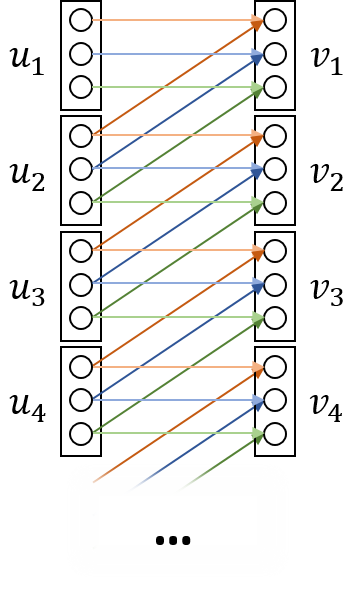}
    \caption{Depthwise convolution.}
    \label{fig:conv-depthwise}
  \end{subfigure}
  \caption{Types of 1D-convolution used in NMT. The color coding indicates weight sharing.}
  \label{fig:conv}
\end{figure}

The input to an 1D convolutional layer is a sequence of $M$-dimensional vectors $u_1,\dots,u_I$. The literature about CNNs usually refers to the $M$ dimensions in each $u_i\in\mathbb{R}^M$ ($i\in [1,I]$) as {\em channels}\index{Channels}, and to the $i$-axis as {\em spatial dimension}.\index{Spatial dimension} The convolution transforms the input sequence $u_1,\dots,u_I$ to an output sequence of $N$-dimensional $v_1,\dots,v_I$ of the same length by moving a {\em kernel}\index{Kernel} of width $K$ over the input sequence. The kernel is a linear transform which maps the $K$-gram $u_i,\dots,u_{i+K-1}$ to the output $v_i$ for $i\in [1,I]$ (we append $K-1$ padding symbols to the input).\index{Padding} Standard convolution parameterizes this linear transform with a full weight matrix $W^\text{std}\in\mathbb{R}^{KM\times N}$:
\begin{equation}
\label{eq:std-conv}
\text{StdConv:} (v_i)_n= \sum_{m=1}^M\sum_{k=0}^{K-1} W^\text{std}_{kM+m,n} (u_{i+k})_m
\end{equation}
with $i\in [1,I]$ and $n\in [1,N]$. Standard convolution represents two kinds of dependencies: Spatial dependency (inner sum in Eq.~\ref{eq:std-conv}) and cross-channel dependency (outer sum in Eq.~\ref{eq:std-conv}). Pointwise and depthwise convolution\index{Convolution!Pointwise convolution}\index{Convolution!Depthwise convolution} factor out these dependencies into two separate operations:
\begin{equation}
\label{eq:pointwise-conv}
\text{PointwiseConv:} (v_i)_n= \sum_{m=1}^M W^\text{pw}_{m,n} (u_{i})_m = u_{i}W^\text{pw}
\end{equation}
\begin{equation}
\label{eq:depthwise-conv}
\text{DepthwiseConv:} (v_i)_n= \sum_{k=0}^{K-1} W^\text{dw}_{k,n} (u_{i+k})_n
\end{equation}
where $W^\text{pw}\in\mathbb{R}^{M\times N}$ and $W^\text{dw}\in\mathbb{R}^{K\times N}$ are weight matrices. Fig.~\ref{fig:conv} illustrates the differences between these types of convolution. The idea behind {\em depthwise separable}\index{Convolution!Depthwise separable convolution} convolution is to replace standard convolutional with depthwise convolution followed by pointwise convolution. As shown in Tab.~\ref{tab:conv-params}, the decomposition into two simpler steps reduces the number of parameters and has been shown to make more efficient use of the parameters than regular convolution in vision~\citep{nn-xception,nn-mobile}.

\begin{table}[t!]
\centering
\footnotesize
\begin{tabular}{l l}
\toprule
Name & Number of parameters \\ 
\midrule
Standard convolution & KMN \\
Pointwise convolution & MN \\
Depthwise convolution & KN \\
Depthwise  separable convolution & N(M+K) \\
\bottomrule
\end{tabular}
\caption{\label{tab:conv-params} Types of convolution and their number of parameters.}
\end{table}



\begin{figure}[t!]
  \centering
  \begin{subfigure}[b]{0.48\textwidth}
    \centering
    \includegraphics[scale=0.43]{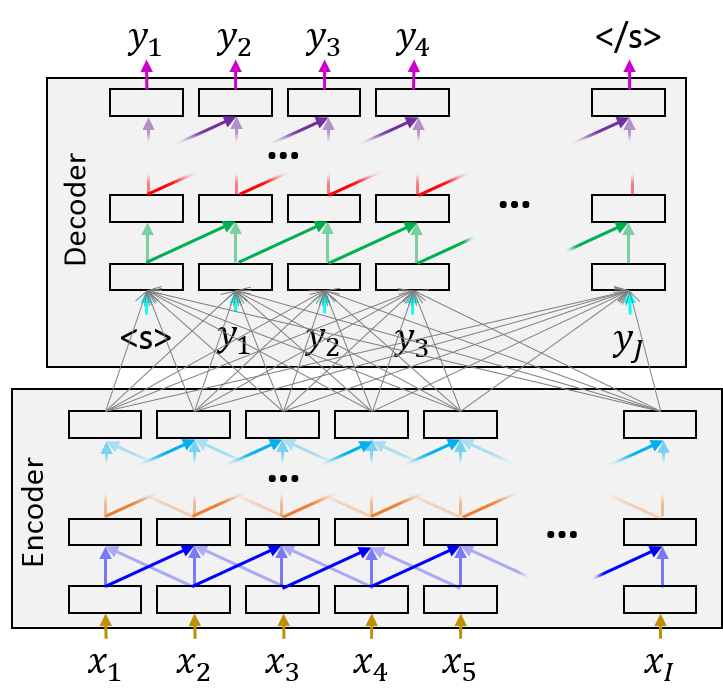}
    \caption{NMT with a convolutional encoder and a convolutional decoder like in the ConvS2S architecture~\citep{nmt-conv-convs2s}.}
    \label{fig:convs2s}
  \end{subfigure}             
  \begin{subfigure}[b]{0.48\textwidth}
    \centering
    \includegraphics[scale=0.43]{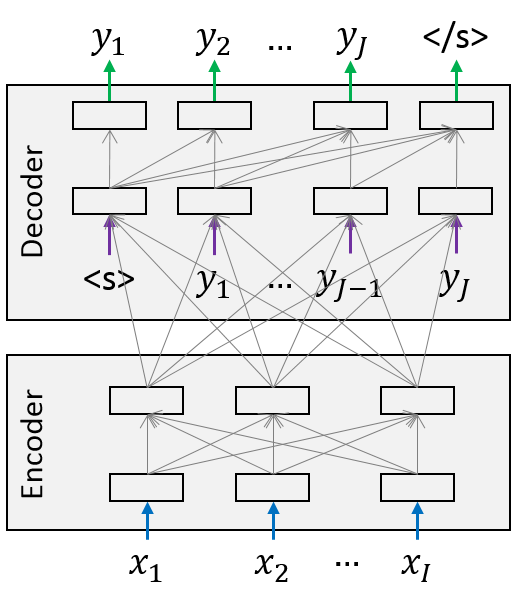}
    \caption{Purely attention-based NMT as proposed by~\citet{nmt-transformer} with two layers.}
    \label{fig:attention-based-nmt}
  \end{subfigure}             
  \caption{Convolutional and purely attention-based architectures. The color coding indicates weight sharing. Gray arrows represent attention.}
  \label{fig:conv-and-transformer}
\end{figure}

Using convolution rather than recurrence in NMT models has several potential advantages. First, they reduce sequential computation and are therefore easier parallelizable on GPU hardware.\index{GPU}\index{Parallelization} Second, their hierarchical structure connects distant words via a shorter path than sequential topologies~\citep{nmt-conv-convs2s} which eases learning~\citep{nn-vanishing-gradient}. Both regular~\citep{nmt-conv-bytenet,nmt-conv-convs2s,nmt-conv-enc} and depthwise separable~\citep{nmt-conv-slicenet,nmt-conv-less-att} convolution have been used for NMT in the past. Fig.\ \ref{fig:convs2s} shows the general architecture for a fully convolutional NMT model such as ConvS2S~\citep{nmt-conv-convs2s} or SliceNet~\citep{nmt-conv-slicenet}\index{ConvS2S}\index{SliceNet|textbf} in which both encoder and decoder are convolutional. Stacking multiple convolutional layers increases the effective context size.\index{Context} In the decoder, we need to mask the receptive field of the convolution operations to make sure that the network has no access to future information~\citep{nn-cnn-mask}. Encoder and decoder are connected via attention. \citet{nmt-conv-convs2s} used attention into the encoder representations after each convolutional layer in the decoder.

\subsection{Self-attention-based Neural Machine Translation}
\label{sec:transformer}

Recall that Eq.~\ref{eq:nmt-factorization} states that NMT factorizes $P(\mathbf{y}|\mathbf{x})$ into conditionals $P(y_j|y_1^{j-1},\mathbf{x})$. We have reviewed two ways to model the dependency on the source sentence $\mathbf{x}$ in NMT: via a fixed-length sentence encoding $c(\mathbf{x})$ (Sec.~\ref{sec:fixed-length}) or via time-dependent context vectors $c_j(\mathbf{x})$ which are computed using attention (Sec.~\ref{sec:attention}). We have also presented two ways to implement the dependency on the target sentence prefix $y_1^{j-1}$:\index{Translation prefix} via a recurrent connection which passes through the decoder state to the next time step (Sec.~\ref{sec:nmt-att-recurrent}) or via convolution (Sec.~\ref{sec:nmt-cnn}). A third option to model target side dependency is using {\em self-attention}.\index{Attention!Self-attention} Using the terminology introduced in Sec.~\ref{sec:attention}, decoder self-attention derives all three components (queries, keys, and values) from the decoder state.\index{Decoder state} The decoder conditions on the translation prefix $y_1^{j-1}$\index{Translation prefix} by attending to its own states from previous time steps. Besides machine translation, self-attention has been applied to various NLP tasks\index{Natural language processing} such as sentiment analysis~\citep{att-self},\index{Sentiment analysis} natural language inference~\citep{embed-att-disan,att-self-nli,att-nli,att-self-hybrid}, text summarization~\citep{att-self-summ-rl}, headline generation~\citep{att-self-headline}, sentence embedding~\citep{att-self-structured-sentence-embedding,embed-att-phraselevel,embed-att-variational}, and reading comprehension~\citep{att-self-reading-comprehension}.  Similarly to convolution, self-attention introduces short paths between distant words and reduces the amount of sequential computation. Studies indicate that these short paths are especially useful for learning strong semantic feature extractors, but (perhaps somewhat counter-intuitively) less so for modelling long-range subject-verb agreement~\citep{nmt-why-self-att}.\index{Long-range dependencies} Like in convolutional models we also need to mask future decoder states to prevent conditioning on future tokens (cf.\ Sec.~\ref{sec:mask-attention}).\index{Attention!Masked attention} The general layout for self-attention-based NMT models is shown in Fig.~\ref{fig:attention-based-nmt}. The first example of this new class of NMT models was the Transformer~\citep{nmt-transformer}.\index{Transformer} The Transformer uses attention for three purposes: 1) within the encoder to enable context-sensitive word representations\index{Context} which depend on the whole source sentence, 2) between the encoder and the decoder as in previous models, and 3) within the decoder to condition on the current translation history. The Transformer uses multi-head attention\index{Attention!Multi-head attention} (Sec.~\ref{sec:attention}) rather than regular attention. Using multi-head attention has been shown to be essential for the Transformer architecture~\citep{nmt-why-self-att,nmt-rnmt}.

A challenge in self-attention-based models (and to some extent in convolutional models) is that vanilla attention as introduced in Sec.~\ref{sec:attention} by itself has no notion of order.\index{Word order} The key-value pairs in the memory\index{Memory} are accessed purely based on the correspondence between key and query ({\em content-based} addressing)\index{Content-based addressing} and not based on a location of the key in the memory ({\em location-based}).\index{Location-based addressing}\footnote{We will discuss cases in which both content and location are taken into account in Secs.~\ref{sec:advanced-attention} and~\ref{sec:memory-networks}} This is less of a problem in recurrent NMT (Sec.~\ref{sec:nmt-att-recurrent}) as queries, keys, and values are derived from RNN states and already carry a strong sequential signal due to the RNN topology.\index{RNN} In the Transformer architecture, however, recurrent connections are removed in favor of attention. \citet{nmt-transformer} tackled this problem using {\em positional encodings}. Positional encodings\index{Positional encodings|textbf} are (potentially partial) functions $\text{PE}:\mathbb{N}\nrightarrow \mathbb{R}^D$ where $D$ is the word embedding size, i.e.\ they are $D$-dimensional representations of natural numbers. They are added to the (input and output) word embeddings\index{Word embeddings} to make them (and consequently the queries, keys, and values) position-sensitive. \citet{nmt-transformer} stacked sine and cosine functions of different frequencies to implement $\text{PE}(\cdot)$:
\begin{equation}
\text{PE}_\text{sin}(n)_d = \left\{
  \begin{array}{lr}
    \sin(10000^{-\frac{d}{D}}n) & : d\text{ is even} \\
    \cos(10000^{-\frac{d}{D}}n) & : d\text{ is odd}
  \end{array}
\right.
\end{equation}
for $n\in\mathbb{N}$ and $d\in[1,D]$. Alternatively, positional encodings can be learned in an embedding matrix~\citep{nmt-conv-convs2s}:
\begin{equation}
\text{PE}_\text{learned}(n) = W_{:,n}
\end{equation}
with weight matrix $W\in\mathbb{R}^{d\times N}$ for some sufficiently large $N$. The input to $\text{PE}(\cdot)$ is usually the absolute position of the word in the sentence~\citep{nmt-transformer,nmt-conv-convs2s}, but relative positioning is also possible~\citep{nmt-transformer-relative}. We will give an overview of extensions to the Transformer architecture in Sec.~\ref{sec:transformer-ext}.

\subsection{Comparison of the Fundamental Architectures}
\label{sec:fundamental-archs}

\begin{figure}[t!]
  \centering
  \begin{subfigure}[b]{0.47\textwidth}
    \centering
    \includegraphics[scale=0.38]{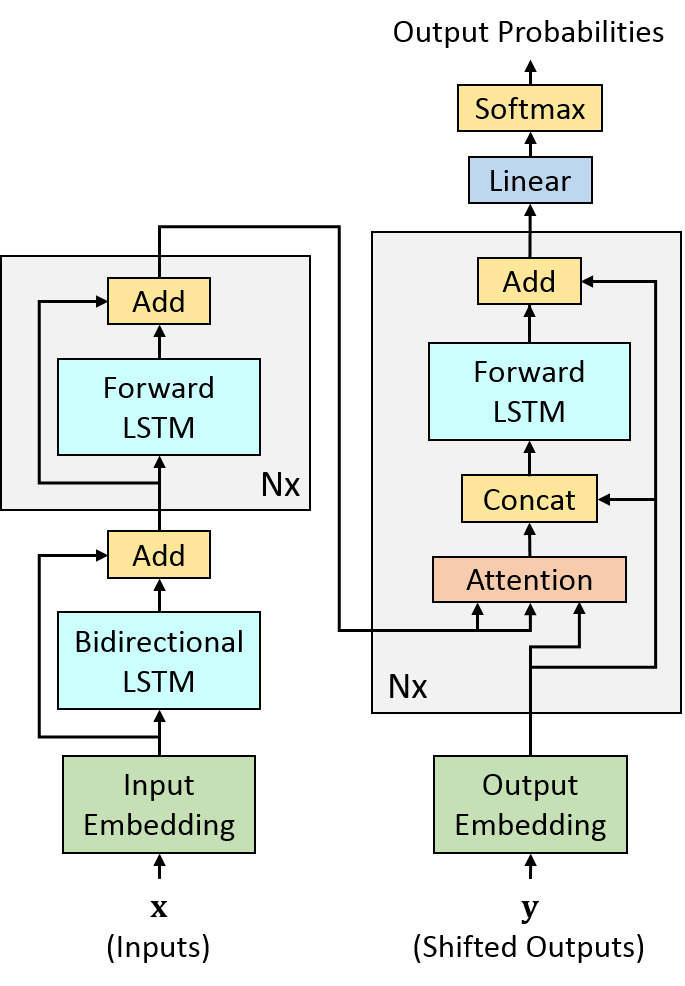}
    \caption{GNMT~\citep{production-gnmt}.}
    \label{fig:plate-gnmt}
  \end{subfigure}             
  \begin{subfigure}[b]{0.47\textwidth}
    \centering
    \includegraphics[scale=0.38]{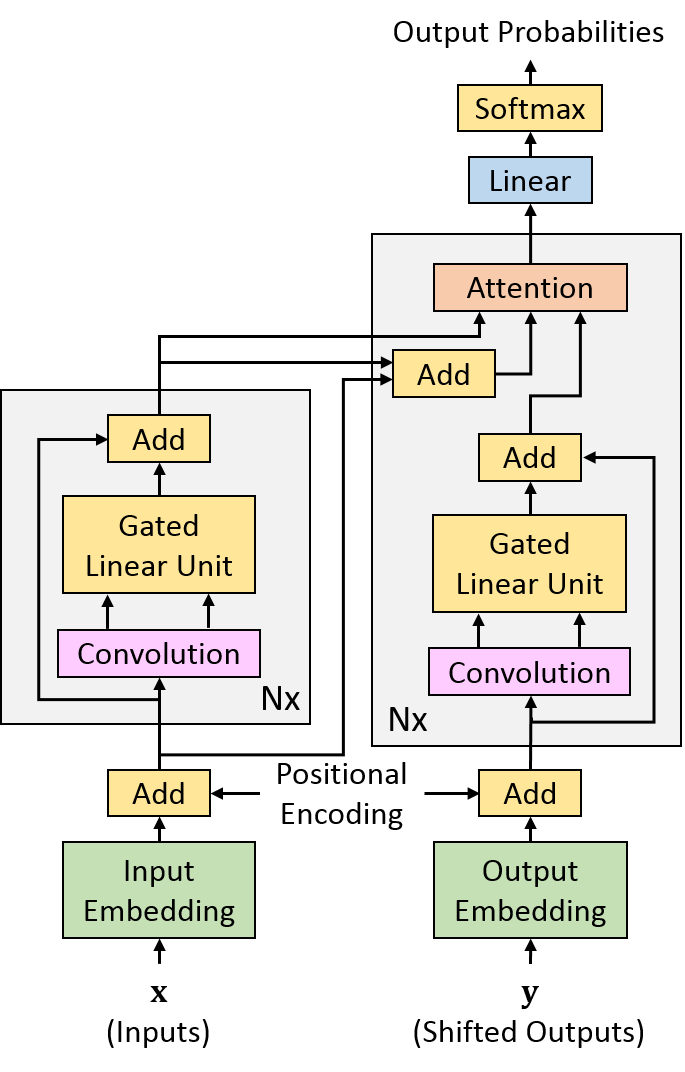}
    \caption{ConvS2S~\citep{nmt-conv-convs2s}.}
    \label{fig:plate-convs2s}
  \end{subfigure}
  \begin{subfigure}[b]{0.47\textwidth}
    \centering
    \includegraphics[scale=0.38]{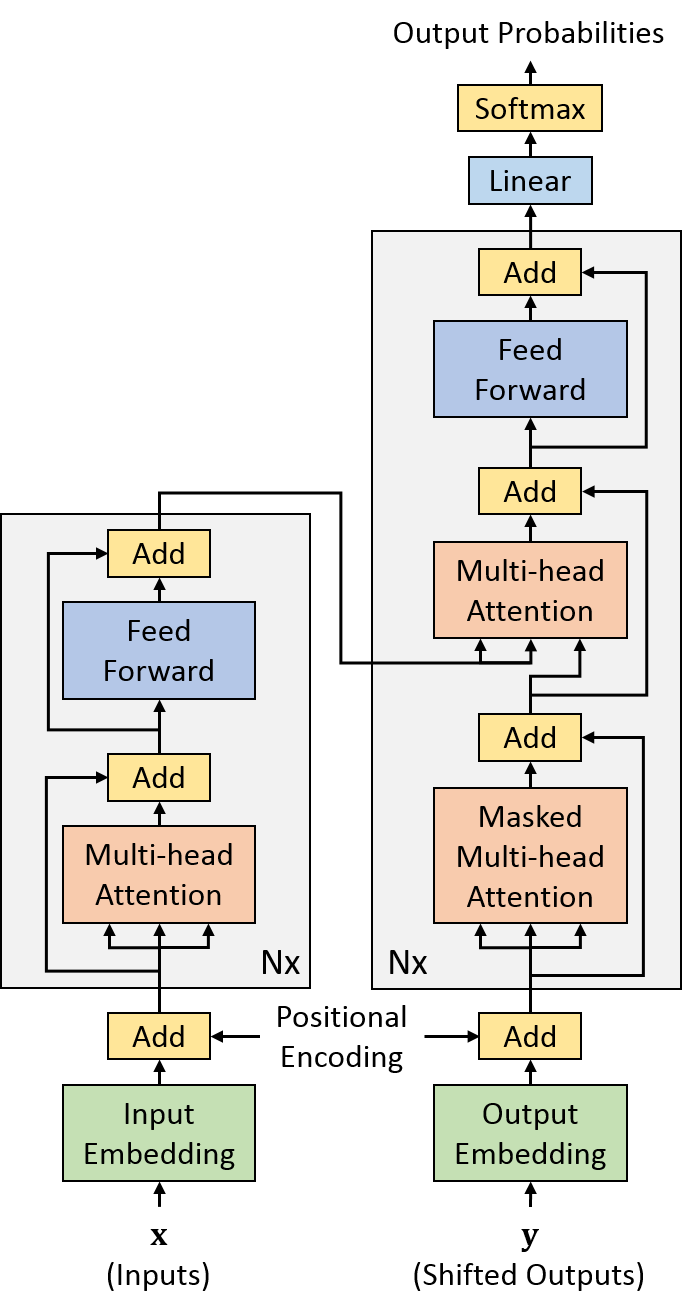}
    \caption{Transformer~\citep{nmt-transformer}.}
    \label{fig:plate-transformer}
  \end{subfigure}
  \begin{subfigure}[b]{0.47\textwidth}
    \centering
    \includegraphics[scale=0.38]{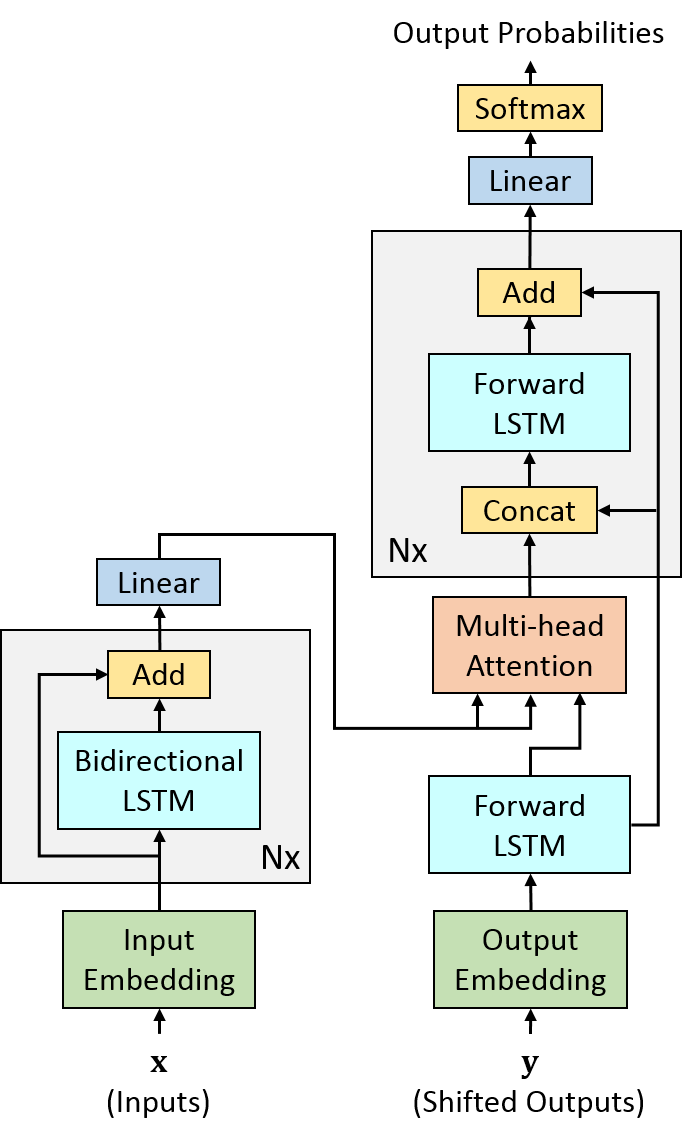}
    \caption{RNMT+~\citep{nmt-rnmt}.}
    \label{fig:plate-rnmt}
  \end{subfigure}
  \caption{Comparison of NMT architectures. The three inputs to attention modules are (from left to right): keys ($K$), values ($V$), and queries ($Q$) as in Fig.~\ref{fig:multi-head-att}.}
  \label{fig:plate-nmt-architectures}
\end{figure}

As outlined in the previous sections, NMT can come in one of three flavors: recurrent, convolutional, or self-attention-based. In this section, we will discuss three concrete architectures in greater detail -- one of each flavor. For an empirical comparison see \citep{ucam-wmt18}. Fig.~\ref{fig:plate-nmt-architectures} visualizes the data streams in Google's Neural Machine Translation system~\citep[GNMT]{production-gnmt}\index{GNMT|textbf} as example of a recurrent network, the convolutional ConvS2S\index{ConvS2S|textbf} model \citep{nmt-conv-convs2s}, and the self-attention-based Transformer\index{Transformer|textbf} model~\citep{nmt-transformer} in plate notation. We excluded components like dropout~\citep{nn-dropout}, batch normalization~\citep{nn-batch-norm}, and layer normalization~\citep{nn-layer-norm} to simplify the diagrams. All models fall in the general category of encoder-decoder networks\index{Encoder-decoder networks}, with the encoder in the left column and the decoder in the right column. Output probabilities are generated by a linear projection layer\index{Projection layer} followed by a softmax\index{Softmax} activation at the end. They all use attention\index{Attention} at each decoder layer to connect the encoder with the decoder, although the specifics differ. GNMT (Fig.~\ref{fig:plate-gnmt}) uses regular attention, ConvS2S (Fig.~\ref{fig:plate-convs2s}) adds the source word encodings to the values, and the Transformer (Fig.~\ref{fig:plate-transformer}) uses multi-head attention (Sec.~\ref{sec:attention}).\index{Attention!Multi-head attention} Residual connections~\citep{nn-residual-connections}\index{Residual connections} are used in all three architectures to encourage gradient flow in multi-layer networks.\index{Gradient-based optimization} Positional encodings\index{Positional encodings} are used in ConvS2S and the Transformer, but not in GNMT. An interesting fusion is the RNMT+\index{RNMT+} model~\citep{nmt-rnmt} shown in Fig.~\ref{fig:plate-rnmt} which reintroduces ideas from the Transformer like multi-head attention into recurrent NMT. Other notable mixed architectures include \citet{nmt-conv-enc} who used a convolutional encoder with a recurrent decoder, \citet{nmt-arch-recurrent-plus-self-att1,nlm-transformer-aws,nmt-arch-recurrent-plus-self-att2} who added self-attention connections to a recurrent decoder, \citet{nmt-arch-att-rec-enc} who used a Transformer encoder and a recurrent encoder in parallel, and \citet{nmt-conv-deconv-decoder} who equipped a recurrent decoder with a convolutional decoder to provide global target-side context.

\section{Neural Machine Translation Decoding}
\label{sec:nmt-decoding}

\subsection{The Search Problem in NMT}

So far we have described how NMT defines the translation probability $P(\mathbf{y}|\mathbf{x})$. However, in order to apply these definitions directly, both the source sentence $\mathbf{x}$ and the target sentence $\mathbf{y}$ have to be given. They do not directly provide a method for generating a target sentence $\mathbf{y}$ from a given source sentence $\mathbf{x}$ which is the ultimate goal in machine translation. The task of finding the most likely translation $\hat{\mathbf{y}}$ for a given source sentence $\mathbf{x}$ is known as the {\em decoding} or {\em inference} problem:
\begin{equation}
\hat{\mathbf{y}}=\argmax_{\mathbf{y}\in \Sigma_{trg}^*} P(\mathbf{y}|\mathbf{x}).
\end{equation}
NMT decoding is non-trivial for mainly two reasons. First, the search space is vast as it grows exponentially with the sequence length. For example, if we assume a common vocabulary size of $|\Sigma_{trg}|=32,000$, there are already more possible translations with 20 words or less than atoms in the observable universe ($32,000^{20}\gg 10^{82}$). Thus, complete enumeration of the search space is impossible. Second, as we will see in Sec.~\ref{sec:nmt-model-errors}, certain types of model errors are very common in NMT. The mismatch between the most likely and the ``best'' translation has deep implications on search as more exhaustive search often leads to worse translations~\citep{cat-tongue}. We will discuss possible solutions to both problems in the remainder of Sec.~\ref{sec:nmt-decoding}.

\subsection{Greedy and Beam Search}

\begin{figure}[tbp!] 
\centering    
\includegraphics[scale=0.52]{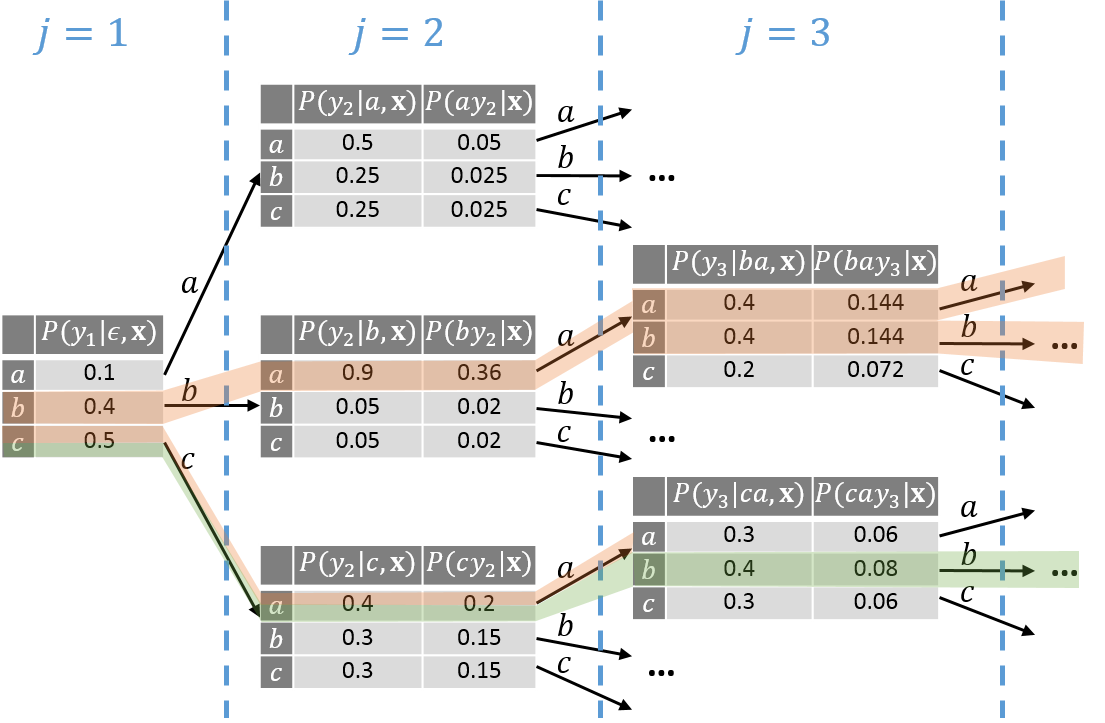}
\caption{Comparison between greedy (highlighted in green) and beam search (highlighted in orange) with beam size 2.}
\label{fig:nmt-decoding}
\end{figure}

The most popular decoding algorithms for NMT are greedy search\index{Greedy decoding|textbf} and beam search.\index{Beam decoding|textbf} Both search procedures are based on the left-to-right factorization\index{Left-to-right factorization} of NMT in Eq.~\ref{eq:nmt-factorization}. Translations are built up from left to right while partial translation prefixes\index{Translation prefix} are scored using the conditionals $P(y_j|y_1^{j-1},\mathbf{x})$. This means that both algorithms work in a time-synchronous manner: in each iteration $j$, partial hypotheses\index{Partial hypothesis} of (up to) length $j$ are compared to each other, and a subset of them is selected for expansion in the next time step. The algorithms terminate if either all or the best of the selected hypotheses end with the end-of-sentence symbol $\text{</s>}$\index{End-of-sentence symbol} or if some maximum number of iterations is reached. Fig.~\ref{fig:nmt-decoding} illustrates the difference between greedy search and beam search. Greedy search (highlighted in green) selects the single best expansion\index{Hypothesis expansion} at each time step: `c' at $j=1$, `a' at $j=2$, and `b' at $j=3$. However, greedy search is vulnerable to the so-called {\em garden-path problem}~\citep{nmt-overview-smt-book}.\index{Garden-path problem} The algorithm selects `c' in the first time step which turns out to be a mistake later on as subsequent distributions are very smooth and scores are comparably low. However, greedy decoding cannot correct this mistake later as it is already committed to this path. Beam search (highlighted in orange in Fig.~\ref{fig:nmt-decoding}) tries to mitigate the risk of the garden-path problem by passing not one but $n$ possible translation prefixes to the next time step ($n=2$ in Fig.~\ref{fig:nmt-decoding}). The $n$ hypotheses which survive a time step are called {\em active hypotheses}.\index{Active hypothesis} At each time step, the accumulated path scores for all possible continuations of active hypotheses are compared, and the $n$ best ones are selected. Thus, beam search does not only expand `c' but also `b' in time step 1, and thereby finds the high scoring translation prefix `ba'. Note that although beam search seems to be the more accurate search procedure, it is not guaranteed to always find a translation with higher or equal score as greedy decoding.\footnote{For example, imagine a series of high entropy conditionals after `baa' and low entropy conditionals after `cab' in Fig.~\ref{fig:nmt-decoding}} It is therefore still prone to the garden-path problem,\index{Garden-path problem} although less so than greedy search. \citet{cat-tongue} demonstrated that even beam search suffers from a high number of search errors.

\begin{algorithm}[t!]
\small
\caption{OneStepRNNsearch$(s_{prev},y_{prev},\mathbf{h})$}
\label{alg:nmt-one-step}
\begin{algorithmic}[1]
\STATE{$\alpha\overset{Eq.~\ref{eq:nmt-alpha}}{\gets} \frac{1}{Z}[\exp(a(s_{prev},h_i))]_{i\in[1,I]}$}
\COMMENT{Attention weights ($\alpha\in \mathbb{R}^{I}$, $Z$ as in Eq.~\ref{eq:nmt-alpha})}
\STATE{$c \overset{Eq.~\ref{eq:nmt-c}}{\gets} \sum_{i=1}^I \alpha_i\cdot h_i$}
\COMMENT{Context vector update ($c\in \mathbb{R}^{m}$)}
\STATE{$s \overset{Eq.~\ref{eq:nmt-s}}{\gets} f(s_{prev}, y_{prev}, c)$}
\COMMENT{RNN state update ($s\in \mathbb{R}^{n}$)}
\STATE{$p \overset{Eq.~\ref{eq:nmt-factorization}}{\gets} g(y_{prev}, s, c)$}
\COMMENT{$p\in \mathbb{R}^{|\Sigma_{trg}|}$ is the distribution over the next target token $P(y_j|\cdot)$}
\RETURN{$s,p$}
\end{algorithmic}
\end{algorithm}

\begin{algorithm}[t!]
\small
\caption{GreedyRNNsearch$(s_{init},\mathbf{h})$}
\label{alg:nmt-search-greedy}
\begin{algorithmic}[1]
\STATE{$\mathbf{y} \gets \langle \rangle$}
\STATE{$s \gets s_{init}$}
\STATE{$y \gets \text{<s>}$}
\WHILE{$y \neq \text{</s>}$}
\STATE{$s,p \gets \text{OneStepRNNsearch}(s,y,\mathbf{h})$}
\STATE{$y \gets \argmax_{\displaystyle w\in \Sigma_{trg}} \pi_w(p)$}
\STATE{$\mathbf{y}.\text{append}(y)$}
\ENDWHILE
\RETURN{$\mathbf{y}$}
\end{algorithmic}
\end{algorithm}

\begin{algorithm}[t!]
\small
\caption{BeamRNNsearch$(s_{init},\mathbf{h},n\in \mathbb{N}_+)$}
\label{alg:nmt-search-beam}
\begin{algorithmic}[1]
\STATE{$\mathcal{H}_{cur}\gets \{(\epsilon, 0.0, s_{init})\}$} \COMMENT{Initialize with empty translation prefix and zero score}
\REPEAT
  \STATE{$\mathcal{H}_{next}\gets \emptyset$}
  \FORALL{$(\mathbf{y},p_{acc},s)\in \mathcal{H}_{cur}$}
    \IF{$y_{|\mathbf{y}|}=\text{</s>}$} 
      \STATE{$\mathcal{H}_{next}\gets \mathcal{H}_{next} \cup \{(\mathbf{y},p_{acc},s)\}$}
      \COMMENT{Hypotheses ending with $\text{</s>}$ are not extended}
    \ELSE
      \STATE{$s,p \gets \text{OneStepRNNsearch}(s,y_{|\mathbf{y}|},\mathbf{h})$}
      \STATE{$\mathcal{H}_{next}\gets \mathcal{H}_{next} \cup \bigcup_{\displaystyle  w\in \Sigma_{trg}} (\mathbf{y}\cdot w, p_{acc} \pi_w(p), s)$}
      \COMMENT{Add all possible continuations}
    \ENDIF
  \ENDFOR
  \STATE{$\mathcal{H}_{cur}\gets \{(\mathbf{y},p_{acc},s)\in \mathcal{H}_{next} : |\{(\mathbf{y}',p_{acc}',s')\in \mathcal{H}_{next} : p_{acc}' > p_{acc}\}| < n\}$}
  \COMMENT{Select $n$-best}
  \STATE{$(\hat{\mathbf{y}},\hat{p}_{acc},\hat{s}) \gets \argmax_{\displaystyle (\mathbf{y},p_{acc},s)\in \mathcal{H}_{cur}} p_{acc}$}
\UNTIL{$\hat{y}_{|\hat{\mathbf{y}}|} = \text{</s>}$}
\RETURN{$\hat{\mathbf{y}}$}
\end{algorithmic}
\end{algorithm}

\subsection{Formal Description of Decoding for the RNNsearch Model}
\label{sec:formal-rnnsearch-search}


In this section, we will formally define decoding for the RNNsearch model~\citep{nmt-bahdanau}\index{RNNsearch|textbf}. We will resort to the mathematical symbols used in Sec.~\ref{sec:nmt-att-recurrent} to describe the algorithms. First, the source annotations\index{Annotations} $\mathbf{h}$ are computed and stored as this does not require any search. Then, we compute the distribution for the first target token $y_1$ using $\text{OneStepRNNsearch}(s_{init},\text{<s>},\mathbf{h})$ (Alg.~\ref{alg:nmt-one-step}). The initial decoder state\index{Decoder state} $s_{init}$ is often a linear transform of the last encoder hidden state $h_I$: $s_{init}=W h_I$ for some weight matrix $W\in \mathbb{R}^{n\times m}$.

Greedy decoding\index{Greedy decoding} selects the most likely target token according the returned distribution and iteratively calls $\text{OneStepRNNsearch}(\cdot)$ until the end-of-sentence symbol $\text{</s>}$ is emitted (Alg.~\ref{alg:nmt-search-greedy}).\index{End-of-sentence symbol} We use the projection function\index{Projection function} $\pi_w(p)$ (Eq.~\ref{eq:project}) which maps the posterior vector $p\in \mathbb{R}^{|\Sigma_{trg}|}$ to the $w$-th component.

The beam search strategy\index{Beam decoding} (Alg.~\ref{alg:nmt-search-beam}) does not only keep the single best partial hypothesis\index{Partial hypothesis} but a set of $n$ promising hypotheses where $n$ is the size of the beam. A partial hypothesis is represented by a 3-tuple $(\mathbf{y},p_{acc},s)$ with the translation prefix $\mathbf{y}\in \Sigma_{trg}^*$\index{Translation prefix}, the accumulated score $p_{acc}\in \mathbb{R}$, and the last decoder state $s\in \mathbb{R}^n$.

\subsection{Ensembling}
\label{sec:ensembling}

\begin{figure}[tbp!] 
\centering    
\includegraphics[scale=0.17]{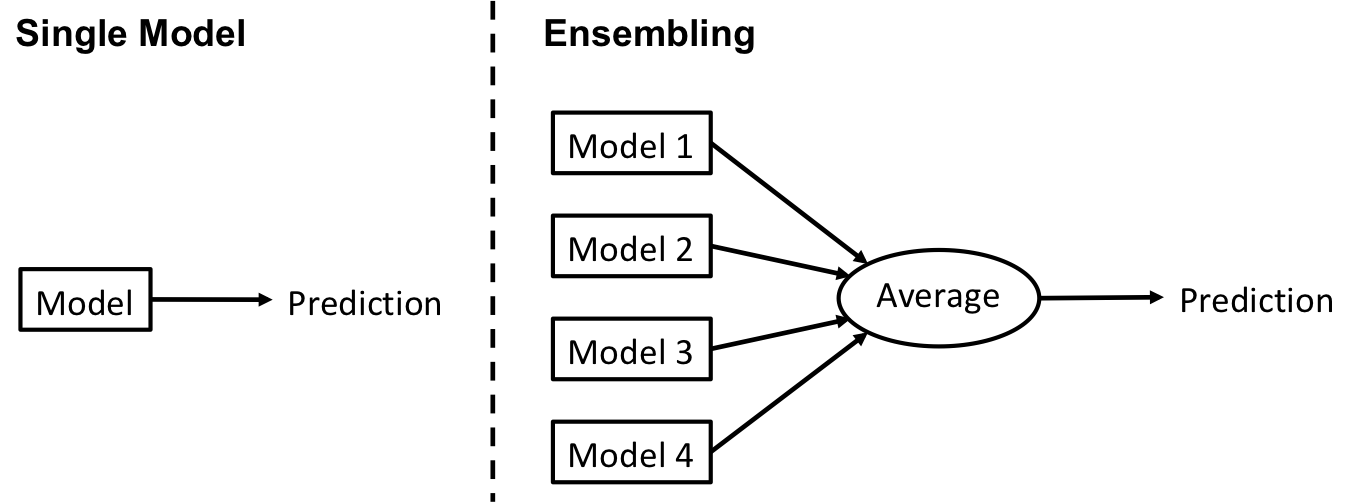}
\caption{Ensembling four NMT models.}
\label{fig:nmt-ensembling}
\end{figure}

Ensembling~\citep{ml-ensembles,nn-ensembles}\index{Ensembling|textbf} is a simple yet very effective technique to improve the accuracy of NMT. The basic idea is illustrated in Fig.~\ref{fig:nmt-ensembling}. The decoder makes use of $K$ NMT networks rather than only one which are either trained independently \citep{nmt-sutskever,sys-neubig-wat16,production-gnmt} or share some amount of training iterations~\citep{sys-uedin-wmt16,sys-kyoto-wat16,sys-qcri}. The ensemble decoder computes predictions for each of the individual models which are then combined using the arithmetic~\citep{nmt-sutskever} or geometric~\citep{sys-kyoto-wat16} average:
\begin{equation}
S_\text{arith}(y_j|y_1^{j-1},\mathbf{x})=\frac{1}{K}\sum_{k=1}^K P_k(y_j|y_1^{j-1},\mathbf{x})
\end{equation}
\begin{equation}
S_\text{geo}(y_j|y_1^{j-1},\mathbf{x})=\sum_{k=1}^K \log P_k(y_j|y_1^{j-1},\mathbf{x}).
\end{equation}
Both $S_\text{arith}(\cdot)$ and $S_\text{geo}(\cdot)$ can be used as drop-in replacement for the conditionals $P(y_j|y_1^{j-1},\mathbf{x})$ in Eq.~\ref{eq:nmt-factorization}. The arithmetic average is more sound as $S_\text{arith}(\cdot)$ still forms a valid probability distribution which sums up to one. However, the geometric average $S_\text{arith}(\cdot)$ is numerically\index{Numerical stability} more stable as log-probabilities can be directly combined without converting them to probabilities. Note that the core idea of ensembling is similar to language model interpolation used in statistical machine translation or speech recognition.

Ensembling consistently outperforms single NMT by a large margin. All top systems in recent machine translation evaluation campaigns ensemble a number of NMT systems~\citep{sys-wmt16,sys-wmt17,sys-wmt18,sys-wmt19,sys-uedin-wmt16,sys-uedin-wmt17,sys-neubig-wat16,sys-kyoto-wat16,sys-qcri,ucam-wmt18,sys-sogou-wmt17,sys-microsoft-wmt18,sys-tencent-wmt18}, perhaps most famously taken to the extreme by the WMT18\index{WMT} submission of Tencent that ensembled up to 72 translation models~\citep{sys-tencent-wmt18}. However, the decoding speed\index{Decoding speed} is significantly worse since the decoder needs to apply $K$ NMT models rather than only one. This means that the decoder has to perform $K$ more forward passes through the networks, and has to apply the expensive softmax\index{Softmax} function $K$ more times in each time step. Ensembling also often increases the number of CPU/GPU switches and the communication overhead between CPU and GPU when averaging is implemented on the CPU.\index{CPU}\index{GPU} Ensembling is also often more difficult to implement than single system NMT. Knowledge distillation\index{Knowledge distillation} which we will discuss in Sec.~\ref{sec:nmt-model-size} is one method to deal with the shortcomings of ensembling. \citet{unfolding} proposed to unfold the ensemble into a single network and shrink the unfolded network afterwards for efficient ensembling.

In NMT, all models in an ensemble usually have the same size and topology and are trained on the same data. They differ only due to the random weight initialization and the randomized order of the training samples. Notable exceptions include \citet{nmt-adaptation-ensembling} who use ensembling to prevent overfitting in domain adaptation,\index{Adaptation} \citet{nmt-ensembling-learning} who combined models that selected their training data based on marginal likelihood,  and the UCAM submission to WMT18~\citep{ucam-wmt18}\index{WMT} that ensembled different NMT architectures with each other.\footnote{Multi-source ensembling~\citep{multiling-multisrc-ensembling,multiling-multisrc}\index{Multi-source NMT} will be discussed in Sec.~\ref{sec:nmt-multilingual} in the context of multilingual NMT.} 

When all models are equally powerful and are trained with the same data, it is surprising that ensembling is so effective. One common narrative is that different models make different mistakes, but the mistake of one model can be outvoted by the others in the ensemble~\citep{ensemble-classifiers}. This explanation is plausible for NMT since translation quality can vary widely between training runs~\citep{nmt-bpe}.\index{Training} The variance in translation performance may also indicate that the NMT error surface is highly non-convex such that the optimizer often ends up in local optima. Ensembling might mitigate this problem. Ensembling may also have a regularization effect on the final translation scores~\citep{nn-deep-book}.

Checkpoint averaging~\citep{sys-amu-wmt16,nmt-tool-marian}\index{Checkpoint averaging|textbf} is a technique which is often discussed in conjunction with ensembling~\citep{nmt-ensembling-compare}. Checkpoint averaging keeps track of the few most recent checkpoints during training, and averages their weight matrices to create the final model. This results in a single model and thus does not increase the decoding time. Therefore, it has become a very common technique in NMT~\citep{nmt-transformer,nmt-transformer-tips,ucam-wmt18}. Checkpoint averaging addresses a quite different problem than ensembling as it mainly smooths out minor fluctuations in the training curve which are due to the optimizer's update rule or noise in the gradient estimation due to mini-batch training.\index{Gradient-based optimization} In contrast, the weights of independently trained models are very different from each other, and there is no obvious direct correspondence between neuron activities\index{Neuron activity} across the models. Therefore, checkpoint averaging cannot be applied to independently trained models.

\subsection{Decoding Direction}
\label{sec:decoding-direction}

Standard NMT factorizes the probability $P(\mathbf{y}|\mathbf{x})$ from left to right (L2R) according Eq.~\ref{eq:nmt-factorization}. Mathematically, the left-to-right order is rather arbitrary, and other arrangements such as a right-to-left (R2L)\index{Right-to-left decoding} factorization are equally correct:
\begin{equation}
P(\mathbf{y}|\mathbf{x})=\underbrace{\prod_{j=1}^J P(y_j|y_1^{j-1},\mathbf{x})}_{=P(y_1|\mathbf{x})\cdot P(y_2|y_1,\mathbf{x})\cdot P(y_3|y_1,y_2,\mathbf{x}) \cdots}=\underbrace{\prod_{j=1}^J P(y_j|y_{j+1}^J,\mathbf{x})}_{=P(y_J|\mathbf{x})\cdot P(y_{J-1}|y_J,\mathbf{x})\cdot P(y_{J-2}|y_{J-1},y_J,\mathbf{x}) \cdots}.
\end{equation}
NMT models which produce the target sentence in reverse order have led to some gains in evaluation systems when combined with left-to-right models~\citep{sys-uedin-wmt16,sys-sogou-wmt17,ucam-wmt18,sys-tencent-wmt18}. A common combination scheme is based on rescoring: A strong L2R ensemble first creates an $n$-best list which is then rescored with an R2L model~\citep{nmt-direction-r2l,sys-uedin-wmt16}. \citet{ucam-wmt18} used R2L models via a minimum Bayes risk framework.\index{Minimum Bayes-risk decoding} The L2R and R2L systems are normally trained independently, although some recent work proposes joint training schemes in which each direction is used as a regularizer\index{Regularization} for the other direction~\citep{nmt-direction-regularization-kl,nmt-direction-regularization-hidden}. Other orderings besides L2R and R2L have also been proposed such as middle-out~\citep{nmt-direction-middleout}, top-down in a binary tree~\citep{nmt-direction-bintree},\index{Trees!Binary trees}\index{Trees} insertion-based~\citep{nmt-direction-ins,nmt-direction-ins-trans,nmt-direction-ins-lowres,nmt-direction-ins-lev}, or in source sentence order~\citep{osnmt}.\index{Insertion-based translation}\index{Insertions}

Another way to give the decoder access to the full target-side context\index{Context} is the two-stage approach of~\citet{nmt-direction-draft} who first drafted a translation, and then employed a multi-source NMT\index{Multi-source NMT} system to generate the final translation from both the source sentence and the draft. \citet{nmt-direction-async} proposed a similar scheme but generated the draft translations in reverse order. A similar two-pass approach was used by \citet{nmt-direction-arabic-two-phase} to make Arabic MT more robust against domain\index{Domain} shifts. \citet{nmt-direction-multi-pass} used reinforcement learning\index{Reinforcement learning} to choose the best number of decoding passes.\index{Multi-pass decoding}

Besides explicit combination with an R2L model and multi-pass strategies, we are aware of following efforts to make the decoder more sensitive to the right-side target context: \citet{nmt-decode-value-networks} used reinforcement learning to estimate the long-term value of a candidate. \citet{nmt-conv-deconv-decoder} provided global target sentence information to a recurrent decoder via a convolutional model. \citet{nmt-decode-cont-relax} proposed a very appealing theoretical framework to relax the discrete NMT optimization problem into a continuous optimization\index{Continuous optimization} problem which allows to include both decoding directions.

\subsection{Efficiency}

NMT decoding is very fast on GPU hardware and can reach up to 5000 words per second.\footnote{\url{https://marian-nmt.github.io/features/}}\index{Decoding speed} However, GPUs\index{GPU} are very expensive, and speeding up CPU\index{CPU} decoding to the level of SMT remains more challenging.\index{Statistical machine translation} Therefore, how to improve the efficiency of neural sequence decoding algorithms is still an active research question. One bottleneck is the sequential left-to-right order\index{Left-to-right factorization} of beam search\index{Beam decoding} which makes parallelization\index{Parallelization} difficult. \citet{nmt-decode-eff-blockwise} suggested to compute multiple time steps in parallel and validate translation prefixes\index{Translation prefix} afterwards. \citet{nmt-decode-eff-latent} reduced the amount of sequential computation by learning a sequence of latent discrete variables which is shorter than the actual target sentence, and generating the final sentence from this latent representation in parallel. \citet{nmt-arch-sru} sped up recurrent NMT by using a simplified architecture for recurrent units. Another line of research tries to reintroduce the idea of {\em hypothesis recombination}\index{Hypothesis recombination} to neural models. This technique is used extensively in traditional SMT~\citep{pb-koehn}. The idea is to keep only the better of two partial hypotheses if it is guaranteed that both will be scored equally in the future. For example, this is the case for $n$-gram language models if both hypotheses end with the same $n$-gram. The problem in neural sequence models is that they condition on the full translation history. Therefore, hypothesis recombination for neural sequence models does not insist on exact equivalence but cluster\index{Clustering} hypotheses based on the similarity between RNN states or the $n$-gram history~\citep{nmt-decode-eff-hypo-recombination,nmt-decode-eff-cluster-gales}. A similar idea was used by \citet{nn-rnn-to-fst}\index{RNN} to approximate RNNs with WFSTs  which also requires mapping histories into equivalence classes.\index{Weighted finite state transducers}

It is also possible to speed up\index{Decoding speed} beam search by reducing the beam size.\index{Beam size} \citet{production-gnmt,nmt-decode-eff-vari-beamsize} suggested to use a variable beam size, using various heuristics to decide the beam size at each time step. Alternatively, the NMT model training can be tailored towards the decoding algorithm~\citep{nmt-train-beam-cont,nmt-train-beam,nmt-train-diff-beam,nmt-train-greedy}. \citet{nmt-train-beam} proposed a loss function for NMT training which penalizes when the reference falls off the beam during training. \citet{kd-nmt}  reported that knowledge distillation\index{Knowledge distillation} (discussed in Sec.~\ref{sec:nmt-model-size}) reduces the gap between greedy decoding\index{Greedy decoding} and beam decoding significantly. Greedy decoding can also be improved by using a small actor network\index{Actor network} which modifies the hidden states in an already trained model~\citep{nmt-train-greedy,nmt-train-greedy2}.

\subsection{Generating Diverse Translations}
\label{sec:diverse-decoding}

An issue with using beam search is that the hypotheses found by the decoder are very similar to each other and often differ only by one or two words~\citep{nmt-decode-diverse,nmt-decode-diverse-mmi,nmt-decode-not-diverse}. The lack of diversity\index{Diversity} is problematic for several reasons. First, natural language in general and translation in particular often come with a high level of ambiguity\index{Ambiguity} that is not represented well by non-diverse $n$-best lists. Second, it impedes user interaction as NMT is not able to provide the user with alternative translations if needed. Third, collecting statistics about the search space such as estimating the probabilities of $n$-grams for minimum Bayes-risk decoding~\citep{mbr-asr,mbr-smt,lmbr-tromble,mbr-nmt-sdl,ucam-wmt18,mbr-nmt} or risk-based training (Sec.~\ref{sec:risk-training}) is much less effective.

\citet{nmt-decode-noisy} added noise\index{Noise} to the activations\index{Neuron activity} in the hidden layer of the decoder network to produce alternative high scoring hypotheses. This is justified by the observation that small variations of a hidden configuration encode semantically similar context~\citep{nn-similar-repr-similar-semantics}. \citet{nmt-decode-diverse,nmt-decode-diverse-mmi} proposed a diversity promoting modification of the beam search objective function. They added an explicit penalization term to the NMT score based on a maximum mutual information\index{Maximum mutual information} criterion which penalizes hypotheses from the same parent node. Note that both extensions can be used together~\citep{nmt-decode-noisy}. \citet{nmt-decode-diverse-groups} suggested to partition the active hypotheses in groups, and use a dissimilarity term to ensure diversity between groups. \citet{nmt-decode-diverse-lookup} found alternative translations by $k$-nearest neighbor search from the greedy translation in a translation memory.\index{Translation memory}

\subsection{Simultaneous Translation}
\label{sec:sim-trans}

Most of the research in MT assumes an offline scenario: a complete source sentence is to be translated to a complete target sentence. However, this basic assumption does not hold up for many real-life applications. For example, useful machine translation for parliamentary speeches and lectures~\citep{nmt-decode-sim-lecture,nmt-decode-sim-kit}\index{Lecture translation} or voice call services such as Skype~\citep{nmt-decode-sim-skype} does not only have to produce good translations but also have to do so with very low latency~\citep{nmt-decode-sim-speed-vs-acc}.\index{Latency} To reduce the latency in such real-time speech-to-speech translation\index{Real-time translation}\index{Simultaneous translation|textbf}\index{Speech-to-speech translation} scenarios it is desirable to start translating before the full source sentence has been vocalized by the speaker. Most approaches frame simultaneous machine translation as source sentence segmentation problem\index{Segmentation}. The source sentence is revealed one word at a time. After a certain number of words, the segmentation policy decides to translate the current partial source sentence prefix and commit to a translation prefix which may not be a complete translation of the partial source.\index{Translation prefix} This process is repeated until the full source sentence is available. The segmentation policy can be heuristic~\citep{nmt-decode-sim-heu} or learned with reinforcement learning~\citep{nmt-decode-sim-rl-smt,nmt-decode-sim-rl-nmt}.\index{Reinforcement learning} The translation itself is usually carried out by a standard MT system which was trained on full sentences. This is sub-optimal for two reasons. First, using a system which was trained on full sentences to translate partial sentences is brittle due to the significant mismatch between training and testing time. \citet{nmt-decode-sim-stacl} tried to tackle this problem by training NMT to generate the target sentence with a fixed maximum latency to the source sentence. Second, human simultaneous interpreters use sophisticated strategies to reduce the latency by changing the grammatical structure~\citep{nmt-decode-sim-paulik1,nmt-decode-sim-paulik2,nmt-decode-sim-interpretese}. These strategies are neglected by a vanilla translation system. Unfortunately, training data from human simultaneous translators is rare~\citep{nmt-decode-sim-paulik2} which makes it difficult to adapt MT to it.

\section{Open Vocabulary Neural Machine Translation}
\label{sec:open-vocabulary}

\subsection{Using Large Output Vocabularies}
\label{sec:large-vocab}

\begin{table}[tb!]
\centering
\footnotesize
\begin{tabular}{l r r r}
\toprule
Vocabulary size & \multicolumn{3}{c}{Number of parameters} \\
 & Embeddings  & Rest & Total \\
\midrule
30K & 55.8M & 27.9M & 83.7M \\
50K & 93.1M & 27.9M & 121.0M \\
150K & 279.2M & 27.9M & 307.1M \\
\bottomrule
\end{tabular}
\caption{\label{tab:rnnsearch-params} Number of parameters in the original RNNsearch model~\citep{nmt-bahdanau} as presented in Sec.~\ref{sec:nmt-att-recurrent} (1000 hidden units, 620-dimensional embeddings). The model size highly depends on the vocabulary size.}
\end{table}

As discussed in Sec.~\ref{sec:word-embeddings}, NMT and other neural NLP models use embedding matrices\index{Embedding matrix}\index{Natural language processing} to represent words as real-valued vectors. Embedding matrices need to have a fixed shape to make joint training with the translation model possible, and thus can only be used with a fixed and pre-defined vocabulary.\index{Closed vocabulary|textbf} This has several major implications for NMT.

\begin{figure}[tbp!] 
\centering    
\includegraphics[scale=0.68]{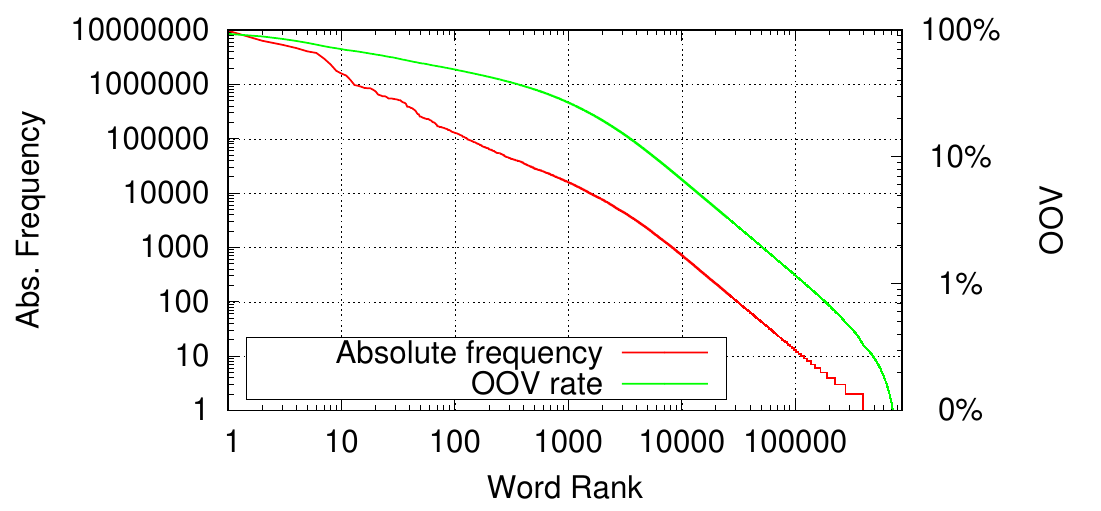}
\caption{Distribution of words in the English portion of the English-German WMT18 training set (5.9M sentences, 140M words).}
\label{fig:zipf}
\end{figure}

First, the size of the embedding matrices grows with the vocabulary size.\index{Vocabulary size} As shown in Tab.~\ref{tab:rnnsearch-params}, the embedding matrices make up most of the model parameters of a standard RNNsearch\index{RNNsearch} model. Increasing the vocabulary size inflates the model drastically. Large models require a small batch size because they take more space in the (GPU)\index{GPU} memory, but reducing the batch size\index{Batch size} often leads to noisier gradients, slower training, and eventually worse model performance~\citep{nmt-transformer-tips}.\index{Gradient-based optimization} Furthermore, a large softmax\index{Softmax} output layer is computationally very expensive. In contrast, traditional (symbolic) MT systems can easily use very large vocabularies~\citep{lm-heafield,misc-mapreduce,hiero-hiero,pb-koehn}.
Besides these practical issues, training embedding matrices for large vocabularies is also complicated by the long-tail distribution of words in a language. Zipf's law~\citep{misc-zipf}\index{Zipf's law} states that the frequency of any word and its rank in the frequency table are inversely proportional to each other. Fig.~\ref{fig:zipf} shows that 843K of the 875K distinct words  (96.5\%) occur less than 100 times in an English text with 140M running words -- that is less than 0.00007\% of the entire text. It is difficult to train robust word embeddings\index{Word embeddings}\index{Robustness} for such rare words.\index{Rare words}
Word-based NMT models\index{Word-based NMT} address this issue by restricting the vocabulary to the $n$ most frequent words, and replacing all other words by a special token UNK.\index{UNK|textbf} A problem with that approach is that the UNK token may appear in the generated translation. In fact, limiting the vocabulary to the 30K most frequent words results in an out-of-vocabulary rate (OOV)\index{OOV rate} of 2.9\% on the training set (Fig.~\ref{fig:zipf}). That means an UNK token can be expected to occur every 35 words. In practice, the number of UNKs is usually even higher. One simple reason is that the test set OOV rate is often higher than on the training set because the distribution of words and phrases naturally varies across genre, corpora, and time. Another observation is that word-based NMT often prefers emitting UNK\index{UNK} even if a more appropriate word is in the NMT vocabulary. This is possibly due to the misbalance between the UNK token and other words: replacing all rare words with the same UNK token leads to an over-representation of UNK in the training set, and therefore a strong bias towards UNK during decoding. 

\subsubsection{Translation-specific Approaches}
\label{sec:nmt-large-vocab-trans-specific}

\citet{nmt-largevoc} distinguished between {\em translation-specific} and {\em model-specific} approaches. Translation-specific approaches keep the shortlist vocabulary in the original form, but correct UNK tokens afterwards. For example, the UNK replace\index{UNK replace} technique~\citep{nmt-unkreplace,nmt-unkreplace-patent} keeps track of the positions of source sentence words which correspond to the UNK tokens. In a post-processing step, they replaced the UNK tokens with the most likely translation of the aligned source word according a bilingual word-level dictionary which was extracted from a word-aligned training corpus. \citet{nmt-pointing-unk} followed a similar idea but used a special pointer network for referring to source sentence words. These approaches are rather ad-hoc because simple dictionary lookup without context is not a very strong model of translation. \citet{nmt-unk2similar} replaced each OOV word with a similar in-vocabulary word based on the cosine similarity between their distributed representations in a pre-processing step. However, this technique cannot tackle all OOVs\index{OOV} as it is based on vector representations of words which are normally only available for a closed vocabulary.\index{Closed vocabulary} Moreover, the replacements might differ from the original meaning significantly. Further UNK replacement strategies were presented by~\citet{nmt-unk-cwmt1,nmt-unk-cwmt2,nmt-unk-cwmt3}, but all share the inevitable limitation of all translation-specific approaches, namely that the translation model itself is indiscriminative between a large number of OOVs.

\subsubsection{Model-specific Approaches}

Model-specific approaches change the NMT model to make training with large vocabularies feasible. For example, \citet{nmt-lex-choice} improved the translation of rare words in NMT by adding a lexical translation model which directly connects corresponding source and target words. Another very popular idea is to train networks to output probability distributions without using the full softmax~\citep{nn-selfnorm}. Noise-contrastive estimation~\citep[NCE]{nn-nce,nn-nce-notes-dyer}\index{Noise-contrastive estimation} trains a logistic regression model which discriminates between real training examples and noise. For example, to train an embedding for a word $w$, \citet{embed-nce} treat $w$ as positive example, and sample from the global unigram word distribution\index{Unigram} in the training data to generate negative examples. The logistic regression model is a binary classifier and thus does not need to sum over the full vocabulary. NCE has been used to train large vocabulary neural sequence models such as language models~\citep{nlm-nce}.\index{Language models} The technique falls into the category of self-normalizing training~\citep{nn-selfnorm}\index{Self-normalizing training} because the model is trained to emit normalized distributions without explicitly summing over the output vocabulary. Self-normalization can also be achieved by adding the value of the partition function to the training loss~\citep{nnjm-devlin}, encouraging the network to learn parameters which generate normalized output. 

Another approach (sometimes referred to as {\em vocabulary selection}) is to approximate the partition function of the full softmax\index{Softmax} by using only a subset of the vocabulary. This subset can be selected in different ways. For example, \citet{nmt-largevoc}\index{RNNsearch-LV|textbf} applied importance sampling\index{Importance sampling} to select a small set of words for approximating the partition function. Both softmax sampling and UNK replace have been used in one of the winning systems at the WMT'15 evaluation on English-German~\citep{sys-montreal-wmt15}.\index{WMT} Various methods have been proposed to select the vocabulary to normalize over during decoding,\index{Vocabulary selection} such as fetching all possible translations in a conventional phrase table~\citep{nmt-sentence-vocab},\index{Phrase table} using the vocabulary of the translation lattices\index{Lattices} from a traditional MT system~\citep[local softmax]{sgnmt},\index{Statistical machine translation} and attention-based~\citep{nmt-vocab-select-att}\index{Attention} and embedding-based~\citep{nmt-vocab-select-embed} methods.

\subsection{Character-based NMT}
\label{sec:char-based}

Arguably, both translation-specific and model-specific approaches to word-based NMT are fundamentally flawed. Translation-specific techniques like UNK\index{UNK} replace\index{UNK replace} are indiscriminative between translations that differ only by OOV words.\index{OOV} A translation model which assigns exactly the same score to a large number of hypotheses is of limited use by its own. Model-specific approaches suffer from the difficulty of training embeddings for rare words\index{Rare words} (Sec.~\ref{sec:large-vocab}). Compound or morpheme splitting~\citep{nmt-morph-split1,nmt-morph-split2}\index{Compound-splitting}\index{Morphology} can mitigate this issue only to a certain extent. More importantly, a fully-trained NMT system even with a very large vocabulary\index{Vocabulary size} cannot be extended with new words. However, customizing systems to new domains\index{Domain} (and thus new vocabularies) is a crucial requirement for commercial MT.\index{Commercial MT} Moreover, many OOV words are proper names which can be passed through untranslated. Hiero~\citep{hiero-hiero}\index{Hiero} and other symbolic systems can easily be extended with new words and phrases.

More recent attempts try to alleviate the vocabulary issue in NMT by departing from words as modelling units. These approaches decompose the word sequences into finer-grained units and model the translation between those instead of words. To the best of our knowledge, \citet{nmt-char} were the first who proposed an NMT architecture which translates between sequences of characters.\index{Character-based NMT|textbf} The core of their NMT network is still on the word-level, but the input and output embedding layers are replaced with subnetworks that compute word representations from the characters of the word. Such a subnetwork can be recurrent~\citep{nmt-char,nmt-char-char3}\index{LSTM}\index{RNN} or convolutional~\citep{nmt-char2,nmt-char-embed-cnn}.\index{Convolutional neural networks} This idea was extended to a hybrid model by~\citet{nmt-word-char-hybrid} who used the standard lookup table embeddings for in-vocabulary words and the LSTM-based embeddings only for OOVs.\index{OOV}

Having a word-level model at the core of a character-based system does circumvent the closed vocabulary\index{Closed vocabulary} restriction of purely word-based models, but it is still segmentation-dependent: The input text has to be preprocessed\index{Preprocessing} with a tokenizer\index{Tokenization} that separates words by blank symbols in languages without word boundary markers, optionally applies compound or morpheme splitting in morphologically rich languages, and isolates punctuation symbols. Since tokenization is by itself error-prone and can degrade the translation performance~\citep{nmt-tokenizers}, it is desirable to design character-level systems that do not require any prior segmentation. \citet{nmt-char-noseg} used a bi-scale recurrent neural network\index{RNN!Bi-scale RNN} that is similar to dynamically segmenting the input using jointly learned gates\index{Gated activation} between a slow and a fast recurrent layer. \citet{nmt-char-noseg2,nmt-char-conv} used convolution\index{Convolutional neural networks} to achieve segmentation-free character-level NMT. \citet{nmt-byte} took character-level NMT one step further and used bytes rather than characters to help multilingual systems.\index{Multilingual NMT} \citet{nmt-char-plan} added a planning mechanism to improve the attention weights between character-based encoders and decoders.\index{Attention}

\subsection{Subword-unit-based NMT}
\label{sec:subword-nmt}

As compromise between characters and full words, compression methods\index{Compression} like Huffman codes~\citep{nmt-huffman},\index{Huffman codes} word piece models\index{Wordpiece model}~\citep{nmt-wordpiece,production-gnmt}, or byte pair encoding~\citep[BPE]{nmt-bpe,nmt-bpe-org}\index{Byte pair encoding|textbf}\index{Subword units|textbf} can be used to transform the words to sequences of subword units. Subwords have been used rarely for traditional SMT~\citep{smt-bpe-length,smt-bpe-fast,smt-bpe-oov}, but are currently the most common translation units for NMT. Byte pair encoding (BPE) initializes the set of available subword units with the character set of the language. This set is extended iteratively in subsequent merge operations. Each merge combines the two units with the highest number of co-occurrences in the text.\footnote{\citet{nmt-bpe-goodness} proposed alternatives to the co-occurrence counts. The wordpiece model~\citep{nmt-wordpiece,production-gnmt} can also be seen as replacing the co-occurrence counts with a language model objective.} This process terminates when the desired vocabulary size is reached. This vocabulary size\index{Vocabulary size} is often set empirically, but can also be tuned on data~\citep{nmt-iterative-bpe}.




Given a fixed BPE vocabulary, there are often multiple ways to segment an unseen text.\footnote{This is not true for other subword compression algorithms. For example, Huffman codes~\citep{nmt-huffman} are prefix codes and thus unique.} The ambiguity\index{Ambiguity} stems from the fact that symbols are still part of the vocabulary even after they are merged. Most BPE implementations select a segmentation greedily by preferring longer subword units. Interestingly, the ambiguity can also be used as source of noise\index{Noise} for regularization.\index{Subword regularization} \citet{nmt-bpe-regularization} reported surprisingly large gains by augmenting the training data with alternative subword segmentations and by decoding from multiple segmentations of the same source sentence.

Segmentation approaches differ in the level of constraints they impose on the subwords. A common constraint is that subwords cannot span over multiple words~\citep{nmt-bpe}. However, enforcing this constraint again requires a tokenizer\index{Tokenization} which is a potential source of errors (see Sec.~\ref{sec:char-based}). The SentencePiece model~\citep{nmt-sentencepiece}\index{SentencePiece model} is a tokenization-free subword model that is estimated on raw text. On the other side of the spectrum, it has been observed that automatically learned subwords generally do not correspond to linguistic entities such as morphemes, suffixes, affixes etc. However, linguistically-motivated subword units\index{Linguistically-motivated subword units}~\citep{nmt-bpe-morph1,nmt-bpe-morph2,nmt-bpe-morph3,nmt-bpe-morph4} that also take morpheme boundaries\index{Morphology} into account do not always improve over completely data-driven ones.

\begin{table}[tb!]
\centering
\footnotesize
\begin{tabular}{p{0.4\textwidth}@{\hspace{0.1\textwidth}}p{0.4\textwidth}}
\toprule
Character-based NMT & Subword-based NMT \\
\midrule
\begin{itemize}[leftmargin=*]
    \item[$+$] Better at transliteration~\citep{nmt-char-vs-bpe-grammatical}.
    \item[$+$] Dynamic segmentation favors characters~\citep{nmt-char-vs-bpe-dynamic}.
    \item[$+$] More robust against noise\index{Noise}\index{Robustness}~\citep{nmt-char-vs-bpe-beyond-trans,nmt-char-noise}.
    \item[$+$] Better modelling of morphology~\citep{nmt-char-vs-bpe-beyond-trans}.
    \item[$+$] Character-level decoders better than subword-based ones in some studies~\citep{nmt-char-noseg,nmt-char-google}.
    \item[$-$] Character-based NMT computationally more expensive than subword-based NMT~\citep{nmt-char-google}.
    \item[$-$] More prone to vanishing gradients~\citep{nmt-char-noseg}.
    \item[$-$] Long-range dependencies have to be modelled over longer time-spans~\citep{nmt-char-noseg2}.
\end{itemize} & \begin{itemize}[leftmargin=*]
    \item[$+$] More grammatical~\citep{nmt-char-vs-bpe-grammatical}.
    \item[$+$] Iterative BPE\index{Byte pair encoding!Iterative BPE} segmentation favors larger vocabulary sizes~\citep{nmt-iterative-bpe}.
    \item[$+$] Better at syntax~\citep{nmt-char-vs-bpe-beyond-trans}.
    \item[$+$] Tends to outperform character-based models in recent MT evaluations~\citep{sys-wmt16,sys-wmt17,sys-wmt18}.
\end{itemize} \\
\bottomrule
\end{tabular}
\caption{\label{tab:char-vs-bpe} Summary of studies comparing characters and subword-units for neural machine translation.}
\end{table}

\subsection{Words, Subwords, or Characters?}

There is no conclusive agreement in the literature whether characters or subwords are the better translation units for NMT.\index{Character-based NMT}\index{Subword units} Tab.~\ref{tab:char-vs-bpe} summarizes some of the arguments. The tendency seems to be that character-based systems have the potential of outperforming subword-based NMT, but they are technically difficult to deploy. Therefore, most systems in the WMT18 evaluation are based on subwords~\citep{sys-wmt18}.
On a more profound level, we do see the shift towards small modelling units not without some concern. \citet{nmt-char-noseg} noted that ``we often have a priori belief that a word, or its segmented-out lexeme, is a basic unit of meaning, making it natural to approach translation as mapping from a sequence of source-language words to a sequence of target-language words.'' Translation is the task of transferring {\em meaning} from one language to another, and it makes intuitive sense to model this process with meaningful units. The decades of research in traditional SMT were characterized by a constant movement towards larger translation units -- starting from the word-based IBM models~\citep{word-ibm}\index{IBM models} to phrase-based\index{Phrase-based translation} MT~\citep{pb-koehn} and hierarchical SMT~\citep{hiero-hiero}\index{Hiero} that models syntactic structures.\index{Syntax} Expressions consisting of multiple words are even more appropriate units than words for translation since there is rarely a 1:1 correspondence between source and target words. In contrast, the starting point for character- and subword-based models is the language's writing system.\index{Writing system} Most writing systems are not logographic but alphabetic or syllabaric and thus use symbols without any relation to meaning. The introduction of symbolic word-level and phrase-level information to NMT is one of the main motivations for NMT-SMT hybrid systems (Sec.~\ref{sec:nmt-smt-hybrids}).

\section{Using Monolingual Training Data}
\label{sec:monolingual-data}

In practice, parallel training data for MT is hard to acquire and expensive, whereas untranslated monolingual data\index{Monolingual data|textbf} is usually abundant. This is one of the reasons why language models (LMs) are central to traditional SMT. For example, in Hiero~\citep{hiero-hiero},\index{Hiero}\index{Statistical machine translation} the translation grammar\index{Grammars} spans a vast space of possible translations but is weak in assigning scores to them. The LM is mainly responsible for selecting a coherent and fluent\index{Fluency} translation from that space. However, the vanilla NMT formalism does not allow the integration of an LM or monolingual data in general.

There are several lines of research which investigate the use of monolingual training data in NMT. \citet{nmt-mono-rnnlm,nmt-mono-rnnlm-csl} suggested to integrate a separately trained RNN-LM into the NMT decoder. Similarly to traditional SMT~\citep{pb-koehn}\index{Statistical machine translation} they started out with combining RNN-LM and NMT scores via a log-linear model\index{Log-linear models} (`shallow fusion').\index{Shallow fusion} They reported even better performance with `deep fusion'\index{Deep fusion} which uses a controller network that dynamically adjusts the weights between RNN-LM and NMT. Both deep fusion and $n$-best reranking with count-based language models\index{Language models} have led to some gains in WMT evaluation systems~\citep{sys-montreal-wmt15,sys-sogou-wmt17}.\index{WMT} The `simple fusion' technique~\citep{simple-fusion} trains the translation model to predict the residual probability of the training data added to the prediction of a pre-trained and fixed LM.

The second line of research makes use of monolingual text via data augmentation. The idea is to add monolingual data in the target language to the natural parallel training corpus. Different strategies for filling in the source side for these sentences have been proposed such as using a single dummy token~\citep{nmt-mono-backtrans} or copying the target sentence over to the source side~\citep{nmt-mono-copytarget}. The most successful strategy is called back-translation~\citep{nmt-mono-backtrans-pre,nmt-mono-backtrans}\index{Back-translation|textbf} which employs a separate translation system in the reverse direction to generate the source sentences for the monolingual target language sentences. The back-translating system is usually smaller and computationally cheaper than the final system for practical reasons, although with enough computational resources improving the quality of the reverse system can affect the final translation performance significantly~\citep{nmt-mono-backtrans-quality}. Iterative approaches that back-translate with systems that were by themselves trained with back-translation can yield improvements~\citep{nmt-mono-backtrans-iter,nmt-mono-backtrans-iter-bi,nmt-mono-backtrans-iter-joint} although they are not widely used due to their computational costs. Back-translation has become a very common technique and has been used in nearly all neural submissions to recent evaluation campaigns~\citep{sys-uedin-wmt16,sys-wmt17,sys-wmt18}.\index{WMT}

A major limitation of back-translation is that the amount of synthetic data has to be balanced with the amount of real parallel data~\citep{nmt-mono-backtrans,sys-uedin-wmt16,nmt-mono-backtrans-ana}. Therefore, the back-translation technique can only make use of a small fraction of the available monolingual data. A misbalance between synthetic and real data can be partially corrected by over-sampling -- duplicating real training samples a number of times to match the synthetic data size.\index{Over-sampling} However, very high over-sampling rates often do not work well in practice. Recently, \citet{nmt-mono-backtrans-scale} proposed to add noise\index{Noise} to the back-translated sentences to provide a stronger training signal from the synthetic sentence pairs. They showed that adding noise does not only improve the translation quality but also makes the training more robust against a high ratio of synthetic against real sentences. The effectiveness of using noise for data augmentation in NMT has also been confirmed by \citet{nmt-data-switchout}. These methods increase the variety of the training data and thus make it harder for the model to fit which ultimately leads to stronger training signals. The variety of synthetic sentences in back-translation can also be increased by sampling multiple sentences from the reverse translation model~\citep{nmt-mono-backtrans-sampling}.

A third class of approaches changes the NMT training loss function to incorporate monolingual data. For example, \citet{nmt-train-autoencoder,nmt-mono-reconstruction,nmt-train-autoencoder-uni}\index{Autoencoders} proposed to add autoencoder terms to the training objective which capture how well a sentence can be reconstructed from its translated representation. Using the reconstruction error is also central to (unsupervised)\index{Unsupervised learning} dual learning approaches~\citep{nmt-mono-dual-unsupervised,sys-ms-parity,nmt-mono-dual-unsupervised-transfer}.\index{Dual learning!Dual unsupervised learning} However, training with respect to the new loss is often computationally intensive and requires approximations. Alternatively, multi-task learning\index{Multi-task learning} has been used to incorporate source-side~\citep{nmt-mono-source-multi-task} and target-side~\citep{nmt-mono-target-multi-task} monolingual data. Another way of utilizing monolingual data in both source and target language is to warm start Seq2Seq training from pre-trained\index{Pre-training} encoder and decoder networks~\citep{nmt-mono-pretrain,nmt-mono-pretrain-and-shallow-fusion}.
An extreme form of leveraging monolingual training data is unsupervised NMT\index{Unsupervised NMT} which removes the need for parallel training data entirely. We will discuss unsupervised NMT in Sec.~\ref{sec:nmt-unsupervised}.

\section{NMT Model Errors}
\label{sec:nmt-model-errors}
\label{sec:nmt-model-search-errors}

NMT is highly effective in assigning scores (or probabilities) to translations because, in stark contrast to SMT, it does not make any conditional independence assumptions in Eq.~\ref{eq:nmt-factorization} to model sentence-level translation.\footnote{It does, however, assume that each sentence can be translated in isolation. We will take a closer look at this assumption in Sec.~\ref{sec:document-level}.} A potential drawback of such a powerful model is that it prohibits the use of sophisticated search procedures. Compared to hierarchical SMT systems like Hiero\index{Hiero}~\citep{hiero-hiero} that explore very large search spaces, NMT beam search appears to be overly simplistic. This observation suggests that translation errors in NMT are more likely due to {\em search errors}\index{Search errors} (the decoder does not find the highest scoring translation) than {\em model errors}\index{Model errors} (the model assigns a higher probability to a worse translation). Interestingly, this is not necessarily the case. Search errors in NMT have been studied by~\citet{nmt-search-vs-model-errors,sgnmt2,cat-tongue}. In particular, \citet{cat-tongue} demonstrated the high number of search errors in NMT decoding. However, as we will show in this section, NMT also suffers from various kinds of model errors in practice despite its theoretical advantage.

\subsection{Sentence Length}
\label{sec:sentence-length}

\begin{figure}[tbp!] 
\centering    
\includegraphics[scale=0.68]{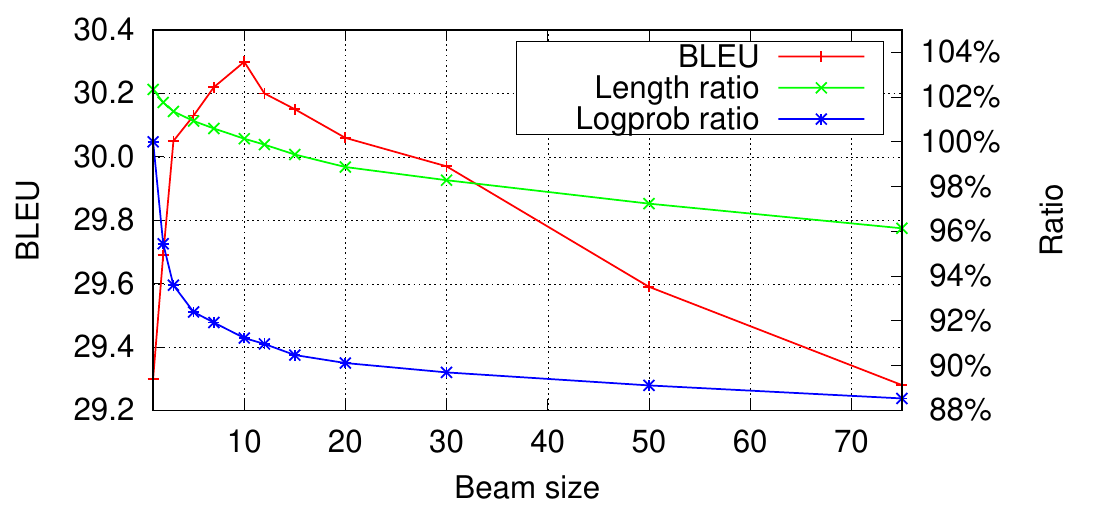}
\caption{Performance of a Transformer model on English-German (WMT15) under varying beam sizes. The BLEU score peaks at beam size 10, but then suffers from a length ratio (hypothesis length / reference length) below 1. The log-probabilities are shown as a ratio with respect to greedy decoding.}
\label{fig:beam-size}
\end{figure}

\begin{figure}[tbp!] 
\centering    
\includegraphics[scale=0.68]{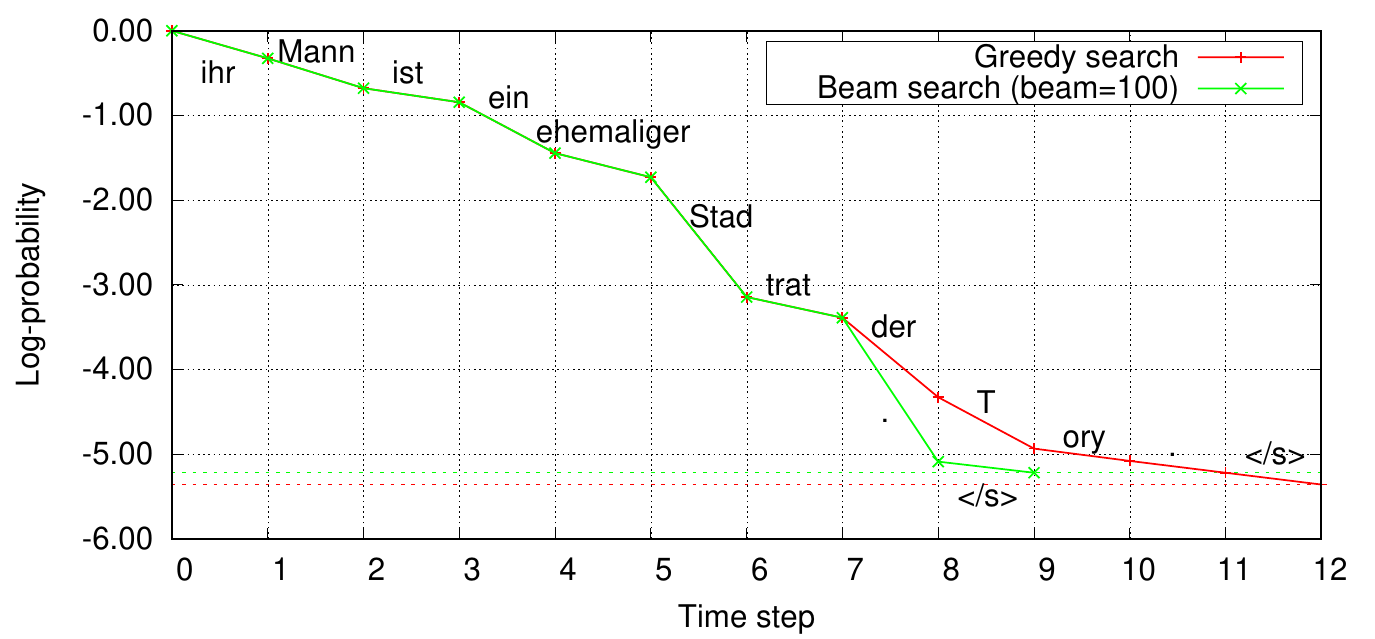}
\caption{The length deficiency in NMT translating the English source sentence ``Her husband is a former Tory councillor.'' into German following \citet{nmt-correcting-lengthbias}. The NMT model assigns a better score to the short translation ``Ihr Mann ist ein ehemaliger Stadtrat.'' than to the greedy translation ``Ihr Mann ist ein ehemaliger Stadtrat der Tory.'' even though it misses the former affiliation of the husband with the Tory Party.}
\label{fig:length-bias}
\end{figure}

Increasing the beam size exposes one of the most noticeable model errors in NMT. The red curve in Fig.~\ref{fig:beam-size} plots the BLEU score~\citep{bleu} of a recent Transformer\index{Transformer} NMT model against the beam size. A beam size\index{Beam size} of 10 is optimal on this test set. Wider beams lead to a steady drop in translation performance because the generated translations are becoming too short (green curve).\index{Sentence length} However, as expected, the log-probabilities of the found translations (blue curve) are decreasing as we increase the beam size. NMT seems to assign too much probability mass to short hypotheses which are only found with more exhaustive search. \citet{nmt-lengthbias} argue that this model error is due to the locally normalized maximum likelihood training\index{Maximum likelihood training} objective in NMT that underestimates the margin between the correct translation and shorter ones if trained with regularization\index{Regularization} and finite data. A similar argument was made by \citet{nmt-correcting-lengthbias} who pointed out the difficulty for a locally normalized model to estimate the ``budget'' for all remaining (longer) translations in each time step. \citet{nmt-length-calibration} demonstrated that NMT models are often poorly calibrated, and that calibration\index{Calibration} issues can cause the length deficiency in NMT. A similar case is illustrated in Fig.~\ref{fig:length-bias}. The NMT model underestimates the combined probability mass of translations continuing after ``Stadtrat'' in time step 7 and overestimates the probability of the period symbol. Greedy decoding\index{Greedy decoding} does not follow the green translation since ``der'' is more likely in time step 7. However, beam search with a large beam keeps the green path and thus finds the shorter (incomplete) translation with better score. In fact, \citet{cat-tongue} linked the bias of large beam sizes towards short translations with the reduction of search errors.

At first glance this seems to be good news: fast beam search with a small beam size is already able to find good translations. However, fixing the model error\index{Model errors} of short translations by introducing search errors\index{Search errors} with a narrow beam seems like fighting fire with fire. In practice, this means that the beam size is yet another hyper-parameter which needs to be tuned for each new NMT training technique (eg.\ label smoothing~\citep{nn-label-smoothing}\index{Label smoothing} usually requires a larger beam), NMT architecture (the Transformer\index{Transformer} model is usually decoded with a smaller beam than typical recurrent models), and language pair~\citep{nmt-overview-six-challenges}. More importantly, it is not clear whether there are gains to be had from reducing the number of search errors with wider beams which are simply obliterated by the NMT length deficiency.

\subsubsection{Model-agnostic Length Models}

The first class of approaches to alleviate the length problem is model-agnostic. Methods in this class treat the NMT model as black box but add a correction term to the NMT score to bias beam search towards longer translations. A simple method is called {\em length normalization}\index{Length normalization|textbf} which divides the NMT probability by the sentence length~\citep{sys-montreal-wmt15,nn-length-norm}:
\begin{equation}
S_\text{LN}(\mathbf{y}|\mathbf{x}) = \frac{\log P(\mathbf{y}|\mathbf{x})}{|\mathbf{y}|}
\end{equation}
\citet{production-gnmt} proposed an extension of this idea by introducing a tunable parameter $\alpha$:
\begin{equation}
S_\text{LN-GNMT}(\mathbf{y}|\mathbf{x}) = \log P(\mathbf{y}|\mathbf{x}) \frac{(1+5)^\alpha}{(1+|\mathbf{y}|)^\alpha}
\end{equation}
Alternatively, like in SMT we can use a word penalty $\gamma(j,\mathbf{x})$\index{Word penalty} which rewards\index{Reward} each word in the sentence:
\begin{equation}
S_\text{WP}(\mathbf{y}|\mathbf{x}) = \sum_{j=1}^J \gamma(j,\mathbf{x})+\log P(y_j|y_1^{j-1},\mathbf{x})
\end{equation}
A constant reward which is independent of $\mathbf{x}$ and $j$ can be found with the standard minimum-error-rate-training~\citep[MERT]{pb-mert} algorithm~\citep{hybrid-nmt-with-smt-features}\index{Minimum error rate training} or with a gradient-based learning scheme~\citep{nmt-correcting-lengthbias}.\index{Gradient-based optimization} Alternative policies which reward words with respect to some estimated sentence length were suggested by~\citet{nmt-optimal-beam,nmt-stopping-criteria}.

\subsubsection{Source-side Coverage Models}

\citet{nmt-coverage-tu} connected the sentence length issue in NMT  with the lack of an explicit mechanism to check the source-side coverage of a translation.\index{Coverage model} Traditional SMT keeps track of a coverage vector\index{Coverage vector} $\mathcal{C}_\text{SMT}\in \{0,1\}^I$ which contains 1 for source words which are already translated and 0 otherwise. $\mathcal{C}_\text{SMT}$ is used to guard against {\em under-translation} (missing translations of some words)\index{Under-translation|textbf} and {\em over-translation} (some words are unnecessarily translated multiple times).\index{Over-translation|textbf} Since vanilla NMT does not use an explicit coverage vector it can be prone to both under- and over-translation~\citep{nmt-coverage-tu,nmt-coverage-otem-utem} and tends to prefer fluency over adequacy~\citep{nmt-train-adequacy}\index{Adequacy}\index{Fluency}. There are two popular ways to model coverage in NMT, both make use of the encoder-decoder attention weight matrix $A$ introduced in Sec.~\ref{sec:attention}. The simpler methods combine the scores of an already trained NMT system with a coverage penalty $cp(\mathbf{x},\mathbf{y})$ without retraining.\index{Coverage penalty} This penalty represents how much of the source sentence is already translated. \citet{production-gnmt} proposed the following term:
\begin{equation}
cp(\mathbf{x},\mathbf{y}) = \beta \sum_{i=1}^I \log \big(\min(\sum_{j=1}^J A_{i,j}, 1.0)\big).
\end{equation}
A very similar penalty was suggested by~\citet{nmt-coverage-almostgnmt}:
\begin{equation}
cp(\mathbf{x},\mathbf{y}) = \alpha \sum_{i=1}^I \log \big(\max(\sum_{j=1}^J A_{i,j}, \beta)\big)
\end{equation}
where $\alpha$ and $\beta$ are hyper-parameters that are tuned on the development set.

An even tighter integration can be achieved by changing the NMT architecture itself and jointly training it with a coverage model~\citep{nmt-coverage-tu,nmt-coverage-embed}. \citet{nmt-coverage-tu} reintroduced an explicit coverage matrix $\mathcal{C}\in [0,1]^{I\times J}$ to NMT. Intuitively, the $j$-th column $\mathcal{C}_{:,j}$ stores to what extend each source word has been translated in time step $j$.  $\mathcal{C}$ can be filled with an RNN-based controller network (the ``neural network based'' coverage model\index{Coverage model!Neural network based coverage model} of \citet{nmt-coverage-tu}).  Alternatively, we can directly use $A$ to compute the coverage (the ``linguistic''\index{Coverage model!Linguistic coverage model} coverage model of \citet{nmt-coverage-tu}):
\begin{equation}
\mathcal{C}_{i,j}=\frac{1}{\Phi_i}\sum_{k=1}^j A_{i,k}
\end{equation}
where $\Phi_i$ is the estimated number of target words the $i$-th source word generates which is similar to fertility\index{Fertility} in SMT. $\Phi_i$ is predicted by a feedforward network\index{Feedforward neural network} that conditions on the $i$-th encoder state. In both the neural network based and the linguistic coverage model, the decoder is modified to additionally condition on $\mathcal{C}$. The idea of using fertilities to prevent over- and under-translation has also been explored by~\citet{nmt-coverage-constrained-fertility}. A coverage model for character-based NMT\index{Character-based NMT} was suggested by \citet{nmt-coverage-char}. 

All approaches discussed in this section operate on the attention weight matrix $A$ and are thus only readily applicable to models with single encoder-decoder attention like GNMT,\index{GNMT} but not to models with multiple encoder-decoder attention modules such as ConvS2S\index{ConvS2S} or the Transformer\index{Transformer} (see Sec.~\ref{sec:fundamental-archs} for detailed descriptions of GNMT, ConvS2S, and the Transformer).

\subsubsection{Controlling Mechanisms for Output Length}

In some sequence prediction tasks such as headline generation or text summarization, the approximate desired output length is known in advance. In such cases, it is possible to control the length of the output sequence by explicitly feeding in the desired length to the neural model. The length information can be provided as additional input to the decoder network~\citep{nn-length-embed-abs1,nn-length-embed-abs2}, at each time step as the number of remaining tokens~\citep{nn-length-embed-remaining}, or by modifying Transformer positional embeddings~\citep{nn-length-pos-embed}.\index{Positional encodings} However, these approaches are not directly applicable to machine translation as the translation length is difficult to predict with sufficient accuracy.

\section{NMT Training}
\label{sec:nmt-training}

NMT models are normally trained using backpropagation~\citep{nn-backprop}\index{Backpropagation} and a gradient-based optimizer\index{Gradient-based optimization} like Adadelta~\citep{nn-adadelta}\index{Adadelta} with cross-entropy loss (Sec.~\ref{sec:xent}). Modern NMT architectures like the Transformer, ConvS2S, or recurrent networks with LSTM \citep{nn-lstm} or GRU~\citep{nnjm-enc-dec} cells help to address known training problems like vanishing gradients~\citep{nn-vanishing-gradient}.\index{Vanishing gradient problem} However, there is evidence that the optimizer still fails to exploit the full potential of NMT models and often gets stuck in suboptima:
\begin{enumerate}
\item NMT models vary greatly in performance, even if they use exactly the same architecture, training data, and are trained for the same number of iterations. \citet{nmt-bpe} observed up to 1 BLEU difference between different models. 
\item NMT ensembling (Sec.~\ref{fig:nmt-ensembling})\index{Ensembling} combines the scores of multiple separately trained NMT models of the same kind. NMT ensembles consistently outperform single NMT by a large margin. The achieved gains through ensembling might indicate difficulties in training of the single models.\footnote{I thank Adrià de Gispert for making that point in our discussions.}
\end{enumerate}
Training is therefore still a very active and diverse research topic. We will outline the different efforts in the literature on NMT training in this section.

\subsection{Cross-entropy Training}
\label{sec:xent}

The most common objective function for NMT training is cross-entropy loss.\index{Cross-entropy|textbf} The optimization problem over model parameters $\Theta$ for a single sentence pair $(\mathbf{x},\mathbf{y})$ under this loss is defined as follows:
\begin{equation}
\label{eq:mle}
\argmin_\Theta \mathcal{L}_\text{CE}(\mathbf{x},\mathbf{y},\Theta) = \argmin_\Theta - \sum_{j=1}^{|\mathbf{y}|} \log P_\Theta (y_j | y_1^{j-1}, \mathbf{x}).
\end{equation}
In practice, NMT training groups several instances from the training corpus into batches, and optimizes $\Theta$ by following the gradient of the average $\mathcal{L}_\text{CE}(\mathbf{x},\mathbf{y},\Theta)$ in the batch. There are various ways to interpret this loss function.

\paragraph{Cross-entropy loss maximizes the log-likelihood of the training data}

A direct interpretation of Eq.~\ref{eq:mle} is that it yields a maximum likelihood estimate of $\Theta$ as it directly maximizes the probability $P_\Theta(\mathbf{y}|\mathbf{x})$:\index{Maximum likelihood training}
\begin{equation}
\label{eq:cross-ent-is-mle}
-\log P_\Theta(\mathbf{y}|\mathbf{x}) \overset{\text{Eq.~\ref{eq:nmt-factorization}}}{=} - \sum_{j=1}^{|\mathbf{y}|} \log P_\Theta (y_j | y_1^{j-1}, \mathbf{x}) = \mathcal{L}_\text{CE}(\mathbf{x},\mathbf{y},\Theta).
\end{equation}

\paragraph{Cross-entropy loss optimizes a Monte Carlo approximation of the cross-entropy to the real sequence-level distribution}

Another intuition behind the cross-entropy loss is that we want to find model parameters $\Theta$ that make the model distribution $P_{\Theta}(\cdot|\mathbf{x})$ similar to the {\em real} distribution $P(\cdot|\mathbf{x})$ over translations for a source sentence $\mathbf{x}$. The similarity is measured with the cross-entropy $H_\mathbf{x}(P,P_\Theta)$. In practice, the real distribution $P(\cdot|\mathbf{x})$ is not known, but we have access to a training corpus of pairs $(\mathbf{x},\mathbf{y})$. For each such pair we consider the target sentence $\mathbf{y}$ as a {\em sample} from the real distribution $P(\cdot|\mathbf{x})$. We now approximate the cross-entropy $H_\mathbf{x}(P,P_\theta)$ using Monte Carlo estimation\index{Monte Carlo estimation} with only one sample ($N=1$):
 \begin{eqnarray*}
 H_\mathbf{x}(P,P_\Theta) &=& \mathbb{E}_\mathbf{y}[-\log P_\Theta(\mathbf{y}|\mathbf{x})] \\
 &=& -\sum_{\mathbf{y}'} P(\mathbf{y}'|\mathbf{x}) \log P_\Theta(\mathbf{y}'|\mathbf{x}) \\
 &\overset{\text{MC}}{\approx}& - \frac{1}{N} \sum_{\mathbf{y}'} \log P_\Theta(\mathbf{y}'|\mathbf{x}) \\
 &\overset{N=1}{=}& - \log P_\Theta(\mathbf{y}|\mathbf{x}) \\
 &=& - \sum_{j=1}^{|\mathbf{y}|} \log P_\Theta (y_j | y_1^{j-1}, \mathbf{x}) \\
 &=& \mathcal{L}_\text{CE}(\mathbf{x},\mathbf{y},\Theta).
 \end{eqnarray*}

\paragraph{Cross-entropy loss optimizes a Monte Carlo approximation of the cross-entropy to the real token-level distribution}

We arrive at the same result if we consider the cross-entropy between the {\em conditionals} of $P(\cdot|y_1^{j-1},\mathbf{x})$ and $P_\Theta(\cdot|y_1^{j-1},\mathbf{x})$ for given $\mathbf{x}$ and prefix $y_1^{j-1}$:
 \begin{eqnarray*}
 \mathbb{E}_{y_j}[-\log P_\Theta(y_j|y_1^{j-1},\mathbf{x})] &=& -\sum_{y_j'} P(y_j'|y_1^{j-1},\mathbf{x}) \log P_\Theta(y_j'|y_1^{j-1}, \mathbf{x}) \\
  &\overset{\text{MC with } N=1}{\approx}& - \sum_{j=1}^{|\mathbf{y}|} \log P_\Theta (y_j | y_1^{j-1}, \mathbf{x}) \\
 &=& \mathcal{L}_\text{CE}(\mathbf{x},\mathbf{y},\Theta).
 \end{eqnarray*}

\paragraph{Cross-entropy loss optimizes the cross-entropy to the Dirac distribution}

Alternatively, we can define a (Dirac) distribution\index{Dirac distribution} which assigns the probability of one to $\mathbf{y}$ and zero to all other target sentences:
\begin{equation}
P_\delta (\mathbf{y}'|\mathbf{x}) = \begin{cases}
     1       & \quad \text{if } \mathbf{y}'=\mathbf{y}\\
    0  & \quad \text{if } \mathbf{y}'\neq \mathbf{y}
  \end{cases}
\end{equation}
The cross-entropy between the Dirac distribution (in this context taking the role of the empirical distribution) and our model distribution $P_\Theta(\cdot|\mathbf{x})$ is: 
\begin{equation}
H_\mathbf{x}(P_\delta,P_\Theta) = -\sum_{\mathbf{y}'}  P_\delta(\mathbf{y}'|\mathbf{x}) \log P_\Theta(\mathbf{y}'|\mathbf{x}) = - \log P_\Theta(\mathbf{y}|\mathbf{x}) \overset{\text{Eq.~\ref{eq:cross-ent-is-mle}}}{=} \mathcal{L}_\text{CE}(\mathbf{x},\mathbf{y},\Theta).
\label{eq:xent-dirac}
\end{equation}



To recap, we have found that the following are equivalent:
\begin{itemize}
    \item Training under cross-entropy loss (Eq.~\ref{eq:mle}).
    \item Maximizing the likelihood of the training data.
    \item Minimizing an estimate of the cross-entropy to the real sequence-level distribution.
    \item Minimizing an estimate of the cross-entropy to the real token-level distribution.
    \item Minimizing the cross-entropy to the Dirac distribution.
\end{itemize}
In particular, we emphasize the equivalence between the sequence-level and the token-level estimation since cross-entropy loss is often characterized as token-level objective in the literature whereas the term {\em sequence-level training} somewhat misleadingly usually refers to risk-based training under BLEU~\citep{nmt-train-reinforce,nmt-train-fb-seqlevel} which is discussed in Sec.~\ref{sec:risk-training}.

\subsection{Training Deep Architectures}

Deep encoders and decoders\index{Deep learning} consisting of multiple layers have now superseded earlier shallow architectures. However, since the gradients have to be back-propagated\index{Backpropagation} through more layers, deep architectures -- especially recurrent ones -- are prone to vanishing gradients~\citep{nn-train-rnn}\index{Vanishing gradient problem} and are thus harder to train. A number of tricks have been proposed recently that make it possible to train deep NMT models reliably. Residual connections~\citep{nn-residual-connections}\index{Residual connections} are direct connections that bypass more complex sub-networks in the layer stack. For example, all the architectures presented in Sec.~\ref{sec:fundamental-archs} (GNMT, ConvS2S, Transformer, RNMT+) add residual connections around attentional, recurrent, or convolutional cells to ease learning (Fig.~\ref{fig:plate-nmt-architectures}). Another technique to counter vanishing gradients is called {\em batch normalization}~\citep{nn-batch-norm}\index{Batch normalization} which normalizes the hidden activations\index{Neuron activity} in each layer in a mini-batch to a mean of zero and a variance of 1. An extension of batch normalization which is independent of the batch size and is especially suitable for recurrent networks is called {\em layer normalization}~\citep{nn-layer-norm}. Layer normalization\index{Layer normalization} is popular for training deep NLP models like the Transformer~\citep{nmt-transformer}.\index{Transformer}

\subsection{Regularization}
\label{sec:regularization}

Modern NMT architectures are vastly over-parameterized~\citep{unfolding}\index{Over-parameterization} to help training~\citep{nn-over-param}. For example, a subword-unit-level Transformer in a standard ``big'' configuration can easily have 200-300 million parameters~\citep{ucam-wmt18}. 
The large number of parameters potentially makes the model prone to {\em over-fitting}:\index{Over-fitting|textbf} The model fits the training data perfectly, but the performance on held-out data suffers as the large number of parameters allows the optimizer to marginally improve training loss at the cost of generalization\index{Generalization} as training proceeds. Techniques that aim to prevent over-fitting in  over-parameterized neural networks are called {\em regularizers}.\index{Regularization|textbf} Perhaps the two simplest regularization techniques are L1 and L2 regularization.\index{L2 regularization|textbf} The idea is to add terms to the loss function that penalize the magnitude of weights in the network. Intuitively, such penalties draw many parameters towards zero and limit their significance. Thus, L1 and L2 effectively serve as soft constraint on the model capacity.

The three most popular regularization techniques for NMT are {\em early stopping},\index{Early stopping} {\em dropout},\index{Dropout} and {\em label smoothing}. Early stopping can be seen as regularization in time as it stops training as soon as the performance on the development set does not improve anymore. Dropout~\citep{nn-dropout} is arguably one of the key techniques that have made deep learning practical. Dropout randomly sets the activities\index{Neuron activity} of hidden and visible units to zero during training. Thus, it can be seen as a strong regularizer for simultaneously training a large collection of networks with extensive weight sharing.

Label smoothing\index{Label smoothing|textbf} has been derived for expectation–maximization training by~\citet{nn-label-smoothing-bill}, and has been applied to large-scale computer vision\index{Computer vision} by~\citet{nn-label-smoothing}. Label smoothing changes the training objective such that the model produces smoother distributions. We have already established in Sec.~\ref{sec:xent} that standard cross-entropy\index{Cross-entropy} training measures the distance of the output distribution to the Dirac\index{Dirac distribution} distribution around the training sample. Label smoothing discounts the likelihood of the training sample and distributes some of the free probability mass among other hypotheses. In NMT, label smoothing is applied as cross-entropy loss to a smoothed distribution $Q(\cdot)$ on the token level:
\begin{equation}
\mathcal{L}_Q(\mathbf{x},\mathbf{y},\Theta) =  -\sum_{j=1}^{|\mathbf{y}|} \sum_{y'\in \Sigma_{trg}} Q(y'|j,\mathbf{y}) \log P_\Theta(y'|y_1^{j-1},\mathbf{x}).
\end{equation}
The distribution $Q(\cdot)$ can take language modelling scores into account~\citep{nn-label-smoothing-ngram}, but usually it is just a smoothed version of the Dirac distribution for the reference label:
\begin{equation}
Q_\alpha(y'|j, \mathbf{y}) =   \begin{cases} 
   \alpha & \text{if } y' = \mathbf{y}_j \\
   \frac{1-\alpha}{|\Sigma_{trg}|-1}       & \text{if } y' \neq \mathbf{y}_j
  \end{cases}
\end{equation}
for some smoothing factor $\alpha\in(0,1]$. Setting $\alpha=1$ recovers the normal cross-entropy loss from Sec.~\ref{sec:xent}.

While label smoothing makes intuitive sense for computer vision, applying it to neural sequence prediction in this way has objectionable side effects on the sequence level.  Considering the probabilities $Q(\cdot)$ assigns to {\em full sequences}, we first note that $Q(\cdot)$ does {\em not} uniformly distribute the remaining probability mass among all other sequences. In fact, distributing it uniformly would result in infinitely small probabilities as there are infinitely many possible sequences. Interestingly, $Q(\cdot)$ does also not assign  a fixed probability of $\alpha$ to the correct sequence $\mathbf{y}$:
\begin{equation}
Q_\alpha(\mathbf{y})=\prod_{j=1}^{|\mathbf{y}|} Q_\alpha(y_j|y_1^{j-1}) = \prod_{j=1}^{|\mathbf{y}|} \alpha = \alpha^{|\mathbf{y}|}.
\end{equation}
Since $\alpha$ is less than one, $Q_\alpha(\cdot)$ is sharper if the correct sequence $\mathbf{y}$ is short, and smoother if it is long. 
Alternative loss functions that encourage smooth output distributions include explicit entropy penalization~\citep{nmt-train-entropy-smoothing} and knowledge distillation\index{Knowledge distillation} (Sec.~\ref{sec:nmt-model-size}).
A regularization effect can also be achieved by making the training data harder to fit by adding noise,\index{Noise} e.g.\ via subword regularization~\citep{nmt-bpe-regularization}, SwitchOut~\citep{nmt-data-switchout}, or noisy back-translation~\citep{nmt-mono-backtrans-scale}\index{Back-translation} (see Secs.~\ref{sec:subword-nmt} and~\ref{sec:monolingual-data}).

\subsection{Large Batch Training}

Another practical trick which is becoming increasingly feasible with the availability of multi-GPU training and large GPU\index{GPU} memories is to use very large batch sizes.\index{Batch size} Large batch training can yield almost linear speed-ups~\citep{train-largebatch-model} as the computation can be distributed across multiple GPUs.\index{Parallelization} Even more importantly, gradients estimated on large batches are naturally less noisy than gradients from small batches, and can yield better overall convergence ~\citep{nmt-transformer-tips,nmt-transformer,ucam-wmt18}. For example, distributing Transformer training across 16 (effective) GPUs can improve over single GPU training by two full BLEU points~\citep{ucam-wmt18}. \citet{nmt-train-batch-size-decay} argued that increasing the batch size during training can have a similar effect as learning rate decay. For a thorough and insightful discussion of large batch training we refer the reader to \citep{train-largebatch-model}.

Previous studies~\citep{nmt-train-trick-bag,nmt-train-batch-size} on batch size were limited by the hardware since -- in vanilla SGD -- the training batch has to fit into the GPU memories. \citet{danielle-syntax} presented a technique called {\em delayed SGD} which sidesteps these limitations by decoupling the batch size limit from the available hardware.

\subsection{Reinforcement Learning}
\label{sec:risk-training}

\citet{nmt-train-reinforce} pointed out two weaknesses of standard MLE training\index{Maximum likelihood training} in neural sequence models. First, there is a discrepancy between NMT training and decoding. During training, the correct target label $y_{j-1}$ is used in the $j$-th time step. Obviously, during decoding, the correct labels are not available, so the previous (potentially wrong) output is fed back to the model. This is called `exposure bias'~\citep{nmt-train-reinforce}\index{Exposure bias} as the model is never exposed to its own mistakes during training. The exposure bias can be tackled by feeding back the ground-truth labels only at early training stages, but gradually switching to feeding back the previously produced target tokens instead as training progresses~\citep{nmt-train-feedback-mistakes}. 

The second issue in NMT training pointed out by~\citet{nmt-train-reinforce} is the mismatch between training loss function and evaluation metric. Training uses cross-entropy loss on the word-level, whereas the final evaluation metric is usually BLEU~\citep{bleu} which is defined on sentence- or document-level.  Both of these problems can be tackled with reinforcement learning~\citep{nmt-train-reinforce,nmt-train-deep-rl}.\index{Reinforcement learning|textbf} In the standard terminology of reinforcement learning, an {\em agent} interacts with an {\em environment} via {\em actions}. A {\em policy} determines the action to pick depending on the environment. The goal is to learn a policy which maximizes the expected {\em reward}.\index{Reward} In NMT, the agent is the NMT model that interacts with the environment consisting of the source sentence $\mathbf{x}$ and the translation history $y_1^{j-1}$ by picking actions (words) according the policy $P(y_j|y_1^{j-1},\mathbf{x})$. 

The advantage of casting NMT as reinforcement learning problem is that the reward does not need to be differentiable, and thus can be any quality measure such as BLEU or GLEU~\citep{production-gnmt}. However, training is computationally very expensive as it requires sampling or decoding during training~\citep{nmt-train-diff-bleu-lower-bound}. Therefore, reinforcement learning is usually used to refine a model trained with cross-entropy~\citep{production-gnmt}.\index{Cross-entropy} However, even though reinforcement learning has yielded some gains in the past in isolated experiments, it is difficult to improve over stronger baselines with recent NMT architectures and back-translation~\citep{nmt-train-rl-study}\index{Back-translation}. \citet{production-gnmt} reported that their gains in BLEU from reinforcement learning were not reflected in the human evaluation.
Other possible applications for reinforcement learning in neural sequence prediction include architecture search~\citep{nn-architecture-search}, adequacy-oriented learning~\citep{nmt-train-adequacy},\index{Adequacy} and simultaneous translation (Sec.~\ref{sec:sim-trans}).\index{Simultaneous translation} An alternative way to incorporate the BLEU metric into NMT training is via a minimum risk formulation~\citep{nmt-train-risk,nmt-train-fb-seqlevel}.\index{Minimum risk training}

\subsection{Dual Supervised Learning}

Recall that NMT networks are trained to model the distribution $P(\mathbf{y}|\mathbf{x})$ over translations $\mathbf{y}$ for a given source sentence $\mathbf{x}$. This training objective takes only one translation direction into account -- from the source language to the target language. However, the chain rule gives us the following relation:
\begin{equation}
\label{eq:dual-chain}
P(\mathbf{y}|\mathbf{x})P(\mathbf{x}) = P(\mathbf{x},\mathbf{y}) = P(\mathbf{x}|\mathbf{y})P(\mathbf{y}).
\end{equation}
Eq.~\ref{eq:dual-chain} is often not satisfied when the two translation models $P(\mathbf{y}|\mathbf{x})$ and $P(\mathbf{x}|\mathbf{y})$ are trained independently. The dual supervised learning~\index{Dual learning!Dual supervised learning} loss $\mathcal{L}_\text{DSL}$ aims to correlate both translation directions as follows~\citep{sys-ms-parity,nmt-train-dual-supervised}:
\begin{equation}
\mathcal{L}_\text{DSL} = {\big(\log P(\mathbf{x}) + \log P(\mathbf{y}|\mathbf{x}) - \log P(\mathbf{y}) - \log P(\mathbf{x}|\mathbf{y}) \big)}^2.
\end{equation}
An alternative way to incorporate both translation directions is the agreement-based approach of~\citet{nmt-train-srctrg-trgsrc-agreement}.

\subsection{Adversarial Training}

Generative adversarial networks~\citep[GANs]{nn-gan}\index{Generative adversarial networks}\index{Adversarial training} have recently become extremely popular in computer vision.\index{Computer vision} GANs were originally proposed as framework for training generative models. For example, in computer vision, a generative model $G$ would generate images that are similar to the ones in the training corpus. The input to a classic GAN is noise\index{Noise} which is sampled from a noise prior. The key idea of adversarial training is that $G$ is trained to fool a discriminative model $D$. The discriminator $D$ takes an image as input and outputs the probability of the image coming from the real training corpus as opposed to being generated by $G$. $G$ and $D$ are jointly trained with opposing objectives: $G$ tries to drive up the probability of $D$ making a mistake whereas $D$ aims to discriminate between real and fake images generated by $G$. GANs are particularly useful when they condition on some input (conditional GANs). For example, a GAN which conditions on a textual description of an image is able to synthesize an image for an unseen description at test time.

In computer vision,\index{Computer vision} it is possible to back-propagate\index{Backpropagation} gradients through the synthetic image and thus train $G$ and $D$ jointly without approximations. The main challenge for applying GANs to text is that this is no longer possible since text consists of a variable number of discrete symbols. Therefore, most work on adversarial training in NLP\index{Natural language processing} relies on reinforcement learning to generate synthetic text samples~\citep{nmt-train-adversarial-bidirectional,nmt-train-adversarial-conditional,nmt-train-adversarial,nn-seqgan,nn-gan-dialog} or directly operates on the hidden activations\index{Neuron activity} in $G$~\citep{nn-gan-hidden}. Besides some exploratory efforts~\citep{nmt-train-adversarial-bidirectional,nmt-train-adversarial-conditional,nmt-train-adversarial}, adversarial training for NLP and particularly NMT is still in its infancy and rather brittle~\citep{nmt-train-adversarial-conditional,nn-gan-lang-falling-short,nn-adversarial-nlp-survey,nn-adversarial-eval}.

\section{Explainable Neural Machine Translation}
\label{sec:explainable-nmt}

\subsection{Post-hoc Interpretability}

Explaining the predictions of deep neural models is hard because they consist of tens of thousands of neurons and millions of parameters. Therefore, explainable and interpretable\index{Interpretability}\index{Explainability} deep learning is still an open research question~\citep{xai-blackbox,xai-survey,xai-mythos,xai-methods,xai-blackbox-nlp-review}. {\em Post-hoc interpretability}\index{Interpretability!Post-hoc interpretability} refers to the idea of sidestepping the model complexity by treating it as a black-box and not trying to understand the inner workings of the model. \citet{xai-methods} defines post-hoc interpretability as follows: ``A trained model is given and our goal is to understand what the model predicts (e.g.\ categories) in terms what is readily interpretable (e.g.\ the input variables)''. In NMT, this means that we try to understand the target tokens (``what the model predicts'') in terms of the source tokens (``the input variables''). Post-hoc intepretability methods such as layer-wise relevance propagation~\citep{xai-lrp} are often visualized with heat maps\index{Visualization} representing the importance of input variables -- pixels in computer vision or source words in machine translation. 

Applying post-hoc interpretability methods to sequence-to-sequence prediction has received some attention in the literature~\citep{nmt-train-sort-explain}. \citet{causal-seq2seq} proposed a causal model which finds related source-target pairs by feeding in perturbed versions of the source sentence. \citet{nmt-interpret-zeroout-src} derived relevance scores for NMT by comparing the predictive probability distributions before and after zeroing out a particular source word. See \citep{nmt-interpret-posthoc-limit} for some general limitations of such post-hoc analyses in NLP.\index{Natural language processing}

\subsection{Model-intrinsic Interpretability}

Unlike the black-box methods for post-hoc interpretability, another line of research tries to understand the functions of individual hidden neurons or layers in the NMT network. Different methods have been proposed to visualize\index{Visualization} the activities\index{Neuron activity} or gradients in hidden layers~\citep{visual-rnn,visual-nlp,visual-nmt,visual-rnnbow}.\index{Decoder state}
\citet{nmt-interpret-enc2pos} shed some light on NMT's ability to handle morphology\index{Morphology} by investigating how well a classifier can predict part-of-speech or morphological tags from the last encoder hidden layer. \citet{nmt-interpret-neuron-regression,nmt-interpret-neuron-regression-dalvi-tool,nmt-interpret-neuron-regression-dalvi} found individual neurons that capture certain linguistic properties with different forms of regression analysis. \citet{nmt-interpret-neuron-regression} were even able to alter the translation (e.g.\ change the gender) by manipulating the activities in these neurons. Other researchers have focused on the attention\index{Attention} layer. \citet{nmt-att-word-disambig} suggested that attention at different layers of the Transformer\index{Transformer} serves different purposes. They also showed that NMT does not use the means of attention for word sense disambiguation.\index{Word sense disambiguation}\index{Ambiguity} \citet{nmt-att-where} provide a detailed analysis of how NMT uses attention to condition on the source sentence.

\subsection{Confidence Estimation in Translation}

Obtaining word level or sentence level confidence scores for translations is not only very useful for practical MT, it also improves the explainability and trustworthiness of the MT system. An obvious candidate for confidence scores from an NMT system are the probabilities the model assigns to tokens or sentences. However, there is some disagreement in the literature on how well NMT models are calibrated~\citep{conf-ana-uncertainty,nmt-length-calibration}.\index{Calibration} Poorly calibrated models do not assign probabilities according to the true data distribution. Such models might still assign high scores to high quality translations, but their output distributions are no reliable source for deriving word-level confidence\index{Confidence} scores. While confidence estimation has been explored for traditional SMT~\citep{conf-smt-ngram-ucam,conf-smt-goodness,conf-smt-ueffing}, it has received almost no attention since the advent of neural machine translation. The only work on confidence in NMT we are aware of is from~\citet{conf-through-att,conf-debugging-nmt} who aim to use attention to estimate word-level confidences.

In contrast, the related field of Quality Estimation\index{Quality estimation} for MT enjoys great popularity, with well-attended annual WMT\index{WMT} evaluation campaigns -- by now in their seventh edition~\citep{wmt18-qa}. Quality estimation aims to find meaningful quality metrics which are more accepted by users and customers than abstract metrics like BLEU~\citep{bleu}, and are more correlated to the usefulness of MT in a real-world scenario. Possible applications for quality estimation include estimating post-editing\index{Post-editing} efficiency~\citep{qa-postedit} or selecting sentences in the MT output which need human revision~\citep{conf-smt-goodness}.

\subsection{Word Alignment in Neural Machine Translation}
\label{sec:hard-attention}

\begin{figure}[t!] 
\centering    
\includegraphics[scale=0.43]{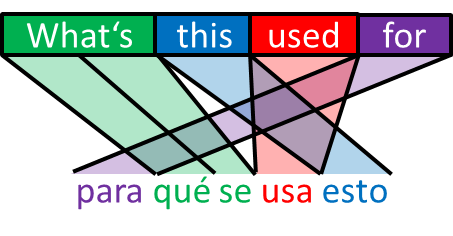}
\caption{Word alignment from the English sentence ``What's this used for'' to the Spanish sentence ``para que se usa esto''.}
\label{fig:word-align}
\end{figure}

Word alignment is one of the fundamental problems in \index{Alignments!Word alignments}traditional phrase-based SMT\index{Statistical machine translation}\index{Phrase-based translation}. SMT constructs the target sentence by matching phrases in the source sentence, and combing their translations to form a fluent sentence~\citep{pb-koehn,hiero-hiero}.\index{Fluency} This approach does not only yield a translation, it also produces a word alignment along with it since each target phrase is generated from a unique source phrase. Thus, a word alignment can be seen as an explanation for the produced translation: each target phrase is explained with a link into the source sentence (Fig.~\ref{fig:word-align}). Unfortunately, vanilla NMT does not have the notion of a hard word alignment. It is tempting to interpret encoder-decoder attention matrices\index{Attention!Attention weight matrix} in neural models (Sec.~\ref{sec:attention}) as (soft) alignments,\index{Alignments!Soft alignment} but previous work has found that the attention weights in NMT are often erratic and differ significantly from traditional word alignments:
\begin{itemize}
    \item ``The  attention  model  for  NMT  does  not  always  fulfill  the  role  of  a word alignment model, but may in fact dramatically diverge.''~\citep{nmt-overview-six-challenges}
    \item ``We perform extensive experiments across a variety of NLP tasks that aim to assess the degree to which attention weights provide meaningful `explanations' for predictions. We find that they largely do not.''~\citep{nmt-interpret-att-no-explanation}
    \item ``Attention weights are only noisy predictors of even intermediate components' importance, and should not be treated as justification for a decision.''~\citep{nmt-interpret-att-inter}
    \item ``Although attention is very useful for under-standing the connection between source and target words, only using attention is not sufficient for deep interpretation of target word generation.''~\citep{visual-nmt}
    \item ``Attention agrees with traditional alignments to a high degree in the case of nouns. However, it captures other information rather than only  the  translational  equivalent  in  the  case  of verbs.''~\citep{nmt-att-where}
    \item ``Attention visualizations are misleading and should be treated with care when explaining the underlying deep learning system.''~\citep{nmt-interpret-att-validity}
\end{itemize}


Despite considerable consensus about the importance of word alignments for practical machine translation~\citep{nmt-overview-six-challenges}, e.g.\ to enforce constraints on the output~\citep{eva-constrained-decoding}\index{Constrained decoding} or to preserve text formatting, introducing explicit alignment information to NMT is still an open research problem. Word alignments have been used as supervision signal for the NMT attention model~\citep{nmt-align-supervised-att1,nmt-align-supervised-att2,nmt-align-supervised-att3,nmt-align-supervised-att-rwth}. \citet{nmt-align-ibm-like-attention} showed how to reintroduce concepts known from traditional statistical alignment models~\citep{word-ibm} like fertility\index{Fertility} and agreement over translation direction to NMT.

Hard attention~\citep{att-image-caption}\index{Attention!Hard attention} is a discrete version of the usual soft attention and is thus closer to the concept of a hard alignment. Similar ideas have been explored for speech recognition~\citep{att-hard-variational}, morphological inflection~\citep{att-hard-morph}, text summarization~\citep{nmt-sim-online-linear-att,nmt-sim-seg-to-seg}, and image caption generation~\citep{att-image-caption}. Some approaches to simultaneous translation\index{Simultaneous translation} presented in Sec.~\ref{sec:sim-trans} explicitly control for reading source tokens and writing target tokens and thereby generate monotonic\index{Alignments!Monotonic alignments} hard alignments on the segment level~\citep{nmt-decode-sim-rl-nmt,nmt-sim-noisy-channel}. Hybrids between soft and hard attention have been proposed by~\citet{nmt-att-softhard1,nmt-att-softhard2}. However, the usefulness of hard attention for generic offline machine translation is often limited since it usually can only represent monotonic alignments.

\citet{nmt-align-alignment-based-nmt} used separate alignment and lexical models and thus were able to hypothesize explicit alignment links during decoding. Alignment-based NMT has been extended to multi-head attention by using an additional alignment head~\citep{nmt-alignment-based-transformer}. A similar idea was pursued by \citet{nmt-alignment-transformer-interpretable-att} who added an additional alignment layer to the Transformer and trained it -- unlike \citet{nmt-alignment-based-transformer} -- in an unsupervised way. The neural operation sequence model of \citet{osnmt} is another way of generating an alignment along with the translation in NMT.

\section{Alternative NMT Architectures}
\label{sec:alt-archs}

\subsection{Extensions to the Transformer Architecture}
\label{sec:transformer-ext}

The Transformer\index{Transformer} model architecture~\citep{nmt-transformer} introduced in Sec.~\ref{sec:transformer} has become the de facto standard architecture for neural machine translation because of its superior translation quality on a variety of language pairs~\citep{sys-wmt18,sys-wmt19}.\footnote{The only contrary evidence we are aware of is from \citet{nmt-arch-recurrent-vs-trans} who found that recurrent models can better model hierarchical structure than the Transformer.} The Transformer comes with a number of techniques which sets it apart from previous architectures such as multi-head attention,\index{Attention!Multi-head attention} self-attention,\index{Attention!Self-attention} large batch training, etc. Some ablation studies in the literature aim to factor out or explain the contributions of these different techniques~\citep{nmt-why-self-att,nmt-rnmt,nmt-how-much-att,nmt-att-word-disambig}. Several attempts have been made to improve different aspects of the vanilla model for machine translation, but none has been widely adopted. Most notably, \citet{nmt-transformer-relative} proposed to embed relative positions rather than absolute ones. A disadvantage of the relative Transformer is the increased computational complexity. The memory keys and values with absolute positions are the same in each decoding step. With relative positioning, however, both have to be recomputed in each time step since the relative positions change over time. The model of \citet{nmt-arch-hysan} works with attention masks (Sec.~\ref{sec:mask-attention}) to narrow down context.\index{Attention!Masked attention}\index{Context} \citet{nmt-arch-weighted-transformer} proposed to weight the output of attention heads inside multi-head attention. The Star-Transformer~\citep{nmt-arch-star-transformer} thins out inter-layer connections of the standard model to reduce computational complexity. With a similar outset, \citet{nmt-arch-parallel-enc-transformer} reported speed-ups\index{Decoding speed} by replacing the single deep encoder with multiple shallow encoders.

Some recent research has focused on large scale language modelling with the Transformer~\citep{nlm-openai-gpt2,nlm-transformer-xl,nlm-transformer-xl-eval,embed-context-openai,nlm-transformer-aws}\index{Language models}. 
The Transformer is also the starting point for neural architectures for contextualized word embeddings (see Sec.~\ref{sec:word-embeddings})\index{Word embeddings!Contextualized word embeddings} such as BERT~\citep{embed-context-bert}.

\subsection{Advanced Attention Models}
\label{sec:advanced-attention}

As shown in Sec.~\ref{sec:attention}, the vast majority of current NMT architectures are based on one of three attention\index{Attention} types: additive, (scaled) dot-product, or multi-head attention~\citep{nmt-dot-product-att,nmt-transformer,nmt-bahdanau}. In this section, we will outline attempts to improve upon these standard models.

Sec.~\ref{sec:sentence-length} discussed the problem of over- and under-translation,\index{Over-translation}\index{Under-translation} and how coverage models can mitigate this problem by controlling the attention weights with fertilities.\index{Fertility} Alternatively, researchers have tried to equip the attention layer itself with additional components like a memory~\citep{nmt-att-mem}\index{Memory} or a recurrent network~\citep{nmt-att-recurrent,nmt-att-recurrent2} to enable it to keep track of the attention history. \citet{nmt-att-finegrained} proposed an attention model that is able to learn different attention weights for each dimension in the values, not only one weight for each value vector.

One potential weakness of the standard models is that they are token-based: the attention output is a weighted average of the values, and the attention weights tend to focus on a single key-value pair. Therefore, there is no explicit mechanism to attend to full phrases rather than subwords\index{Subword units} or characters.\footnote{This does not mean that the source sentence context is always reduced to a single input token since the encoder hidden states are by themselves context-sensitive.} Phrase-based NMT\index{Phrase-based translation!Phrase-based NMT} which equips the model with the ability to attend to full phrases or multi-word expressions has been studied by~\citet{nmt-phrase-multiword,nmt-phrase-chunk,nmt-phrase-p2p,nmt-phrase1,nmt-phrase-att,nmt-phrase-att2}.

On the other side of the spectrum, it has been noted that regular attention sometimes spreads out over too many elements, especially when applied over long sequences. The attention output in this case is an average of many values which is naturally more noisy than with sharp attention, and which impedes the propagation of information through the network. Hard attention\index{Attention!Hard attention} (Sec.~\ref{sec:hard-attention}) removes this sort of noise, but is often restricted to monotonic alignment.\index{Alignments!Monotonic alignments} \citet{nmt-att-temperature} proposed to explicitly learn to set the temperature of attention weights to control the softness of attention. Another potential solution has been suggested by~\citet{nmt-att-gate} who used GRU gates\index{Gated activation} rather than weighted linear combinations\index{Linear combination} to compute the attention output from the values.

\subsection{Memory-augmented Neural Networks}
\label{sec:memory-networks}

\begin{figure}[t!] 
\centering    
\includegraphics[scale=0.3]{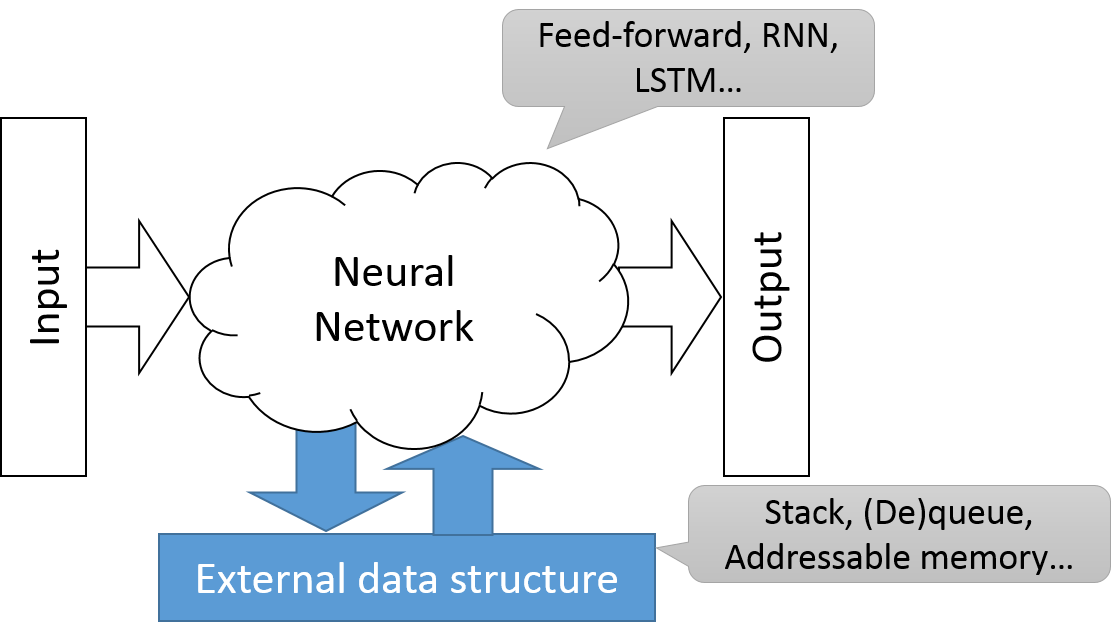}
\caption{Neural networks with external memory.}
\label{fig:external-memory}
\end{figure}

RNNs\index{RNN} are theoretically Turing-complete~\citep{nn-rnn-turingcomplete} and thus potentially very powerful models of computation. However, since training is still a challenge (see Sec.~\ref{sec:nmt-training}), even advanced RNN architectures like LSTMs~\citep{nn-lstm} fail to solve certain basic sequence-to-sequence tasks like (repeated) copying or reversal in practice~\citep{nnmem-ntm,nnmem-nnpda-deepmind}. This observation motivated researchers to add external memory structures like a memory\index{Memory|textbf} tape~\citep{nnmem-turing-org} or a stack~\citep{nnmem-stack-org,nnmem-stack-org2} to the neural network. The basic idea is illustrated in Fig.~\ref{fig:external-memory}. Besides producing the output sequence, the neural network learns to operate an external data structure. The external memory is not part of the neural network but the network learns to communicate with it through conceptually discrete operations like PUSH and POP. However, in order to train the whole system with a gradient-based optimizer, these discrete operations are often approximated with continuous versions~\citep{nnmem-ntm,nnmem-nnpda-deepmind,nnmem-nnpda-mikolov}. Various data structures have been used in combination with neural networks such as (inter alia) stacks~\citep{nnmem-nnpda-mikolov},\index{Neural stack}\index{Neural Turing machine} (double-ended) queues~\citep{nnmem-nnpda-deepmind}, addressable memory cells~\citep{nnmem-ram,nnmem-ntm,nnmem-dnc}, and hierarchical memory structures~\citep{nnmem-hierarchical}. \citet{nnmem-nnpda-deepmind} suggested that even simple data structures like dequeues help to solve linguistically motivated tasks like bigram flipping or Inversion Transduction Grammar~\citep[ITG]{fst-itg}\index{Grammars!Inversion transduction grammars} tasks.
Research on these kinds of neural network operated data structures still mainly focuses on synthetic tasks like relatively simple algorithmic problems. Initial efforts to apply this line of research to real world problems are limited to neural machine translation~\citep{nnmem-nmt,nnmem-nmt2,nnmem-nmt-constrained,nmt-ntm-multichannel}, sentence simplification~\citep{nnmem-simplification}, and text normalization~\citep{nnmem-textnorm}.\index{Text normalization}

\subsection{Beyond Encoder-decoder Networks}

All NMT architectures which we have discussed in the previous sections fall in the category of encoder-decoder networks:\index{Encoder-decoder networks} An encoder network computes a fixed or variable length continuous hidden representation of the source sentence, and a separate decoder network defines a probability distribution over target sentences given that representation. There are some initial efforts in the literature to depart from this overall structure. For example, variational methods\index{Variational neural machine translation} that define a {\em distribution} over (a part of) the hidden representations have been explored by \citet{nmt-variational1,nmt-variational2,nmt-variational-latent-structure,nmt-variational-generative}. Non-autoregressive NMT\index{Non-autoregressive neural machine translation} which aims to reduce or remove the sequential dependency on the translation prefix\index{Translation prefix} inside the decoder for enhanced parallelizability\index{Parallelization} has been studied by~\citet{nmt-arch-semiautoregressive,nmt-arch-nonautoregressive1,nmt-arch-nonautoregressive2,nmt-arch-nonautoregressive3,nmt-arch-ctc,nmt-arch-nonauto-iter-refine,nmt-syntax-speed}. \citet{nmt-arch-2d,nmt-arch-active-mem} recomputed the encoder state after each time step and thus effectively expanded the hidden representation into a 2D structure. The architecture proposed by \citet{nmt-arch-layer-wise-encdec} does not only use the last encoder layer as hidden representation, but instead connects encoder and decoder layers at the same depth via attention.

\section{Data Sparsity}

Deep learning methods are notoriously data hungry. For example, traditional statistical machine translation still often outperforms neural machine translation when training data is scarce~\citep{nmt-overview-six-challenges,nmt-direction-ins-lowres}. In this section we will look at the problem of training data sparsity from different angles such as reducing noise in training data (Sec.~\ref{sec:corpus-filtering}), using data from a different domain, or making use of less or no parallel data.

\subsection{Corpus Filtering}
\label{sec:corpus-filtering}

Unfortunately, MT training data is usually inherently noisy as it is often extracted (semi-) automatically by crawling the web~\citep{data-mt-web-crawl1,data-mt-web-crawl2} and therefore commonly contains sentence fragments, wrong languages, misaligned sentence pairs~\citep{nmt-noise-impact}, or MT output rather than genuine parallel text~\citep{data-web-is-mt1,data-web-is-mt2}. In the previous sections we discussed several instances of the use of synthetic noise in NMT. For example, adding noise to the synthetic sentences in back-translation\index{Back-translation} can be beneficial (Sec.~\ref{sec:monolingual-data}). Noise can also be used to generate diverse translations (Sec.~\ref{sec:diverse-decoding}) or as regularizer (Sec.~\ref{sec:regularization}).\index{Regularization}  However, when discussing the role of noise\index{Noise} in NMT it is imperative to carefully differentiate between the various kinds of noise and the ways it impacts NMT. Studies have shown that NMT is not robust against naturally occurring noise at training~\citep{nmt-noise-impact}\index{Robustness} and test~\citep{nmt-char-noise,data-nmt-robustness-mtnt,data-nmt-robustness,nmt-vs-smt-asr} time. Robustness at test time can be improved by training on synthetic noise~\citep{nmt-train-noise-robustness,nmt-train-noise-robustness2}. Corpus filtering\index{Filtering}\index{Data selection} to reduce the amount of noise in the training data has been widely studied for traditional SMT~\citep{data-smt-filtering,data-smt-filtering-graph},\index{Statistical machine translation} often in context of domain adaptation~\citep{data-smt-filtering-adaptation1,data-smt-filtering-adaptation2}.\index{Adaptation} More recent research on data filtering focuses on NMT since \citet{data-filtering-smt-neq-nmt} had shown that filtering techniques developed for SMT are less useful for NMT. One of the first approaches to NMT corpus filtering was the method of \citet{data-filtering-nmt-first} based on semantic analysis. The most effective approaches in the WMT18\index{WMT} shared task on corpus filtering for NMT~\citep{data-filtering-wmt18} used a combination of likelihood scores from neural translation models and neural language models\index{Language models!Neural language models} which have been trained on clean data~\citep{data-ms-filtering,data-rwth-filtering,sys-microsoft-wmt18}.\index{Filtering!Dual filtering} These criteria prefer sentence pairs which are likely translations of one another according the translation model~\citep{data-filtering-nmt-static-dynamic}. \citet{data-nmt-boosting} proposed the exact opposite, arguing that NMT training should concentrate on ``difficult'' training samples, i.e.\ samples with low translation probability. An alternative to hard data filtering called {\em curriculum learning}\index{Curriculum learning}~\citep{data-cl} that controls the order of training samples has been applied to NMT by~\citet{data-cl-trusted-online,data-filtering-smt-neq-nmt,data-cl-nmt1,data-cl-nmt2}.

\subsection{Domain Adaptation}
\label{sec:domain-adaptation}

There is a robust body of research on domain adaptation\index{Adaptation|textbf}\index{Domain|textbf} for machine translation~\citep{nmt-adaptation-survey1,nmt-adaptation-survey2}. Popular domain adaptation techniques for both SMT and NMT aim to select~\citep{data-smt-filtering-adaptation-kit,data-smt-filtering-adaptation1,data-smt-filtering-adaptation2,nmt-adaptation-select-weight,nmt-adaptation-select-embed}\index{Filtering} or weight~\citep{nmt-adaptation-select-weight,nmt-adaptation-weight,nmt-adaptation-weight2}\index{Instance weighting} samples in a large out-of-domain corpus. Back-translation\index{Back-translation} (Sec.~\ref{sec:monolingual-data}) can also be used for domain adaptation by back-translating sentences from an in-domain monolingual corpus.\index{Monolingual data} Another simple yet very effective method is to jointly train on in-domain and out-domain sentences, possibly with domain-tags to help learning~\citep{nmt-adaptation-concat-tag,nmt-adaptation-concat-tag2,nmt-adaptation-concat-tag-discriminative}. \citet{nmt-adaptation-concat-wins} showed that a simple concatenation of in-domain and out-domain corpora can already increase the robustness\index{Robustness} and generalization\index{Generalization} of NMT significantly. \citet{nmt-adaptation-sgnmt} studied domain adaptation by constraining\index{Constrained decoding} an NMT system to SMT\index{Statistical machine translation} lattices.\index{Lattices} \citet{nmt-adaptation-ensembling} ensembled separately trained general-domain and in-domain models.

Another widely used technique is to train the model on a general domain corpus, and then fine-tune\index{Fine-tuning|textbf}\index{Pre-training} it by continuing training\index{Continued training} on the in-domain corpus~\citep{nmt-adaptation-finetuning,nmt-mono-backtrans}. Fine-tuning bears the risk of two negative effects: catastrophic forgetting~\citep{nmt-adaptation-cf,nmt-adaptation-cf-investigation}\index{Catastrophic forgetting} and over-fitting.\index{Over-fitting} Catastrophic forgetting occurs when the performance on the specific domain is improved after fine-tuning, but the performance of the model on the general domain has decreased drastically.\index{Generalization} The risk of over-fitting is connected to the fact that the in-domain corpus is usually very small. Both effects can be mitigated by artificially limiting the learning capabilities of the fine-tuning stage, e.g.\ by freezing sub-networks~\citep{nmt-adaptation-freeze} or by only learning additional scaling factors for hidden units rather than full weights~\citep{nmt-adaptation-lhuc-org,nmt-adaptation-lhuc}. A very elegant way to prevent over-fitting and catastrophic forgetting is to apply regularizers\index{Regularization} (Sec.~\ref{sec:regularization}) to keep the adapted model weights close to their original values. \citet{nmt-adaptation-secretly-kd,nmt-adaptation-kd} regularized the output distributions using techniques inspired by knowledge distillation\index{Knowledge distillation} (Sec.~\ref{sec:nmt-model-size}). \citet{nmt-adaptation-l2} applied standard L2 regularization\index{L2 regularization} and a variant of dropout\index{Dropout} to domain adaptation. Elastic weight consolidation~\citep{nn-ewc}\index{Elastic weight consolidation} can be seen as generalization of L2 regularization that takes the importance of weights (in terms of Fisher information) into account, and has been applied to NMT domain adaptation by~\citet{nmt-adaptation-ewc,nmt-adaptation-ewc-danielle}. In particular, \citet{nmt-adaptation-ewc-danielle} showed that EWC does not only reduce catastrophic forgetting but even yields gains on the general domain when used for fine-tuning on a related domain.

\subsection{Low-resource NMT}
\label{sec:low-res-nmt}

One of the areas in which traditional SMT\index{Statistical machine translation} still often outperforms NMT is low-resource translation~\citep{nmt-overview-six-challenges,nmt-direction-ins-lowres}.\index{Low-resource machine translation} However, several techniques have been proposed to improve the performance of NMT under low-resource conditions. In general, the methods discussed in Sec.~\ref{sec:monolingual-data} to leverage monolingual data such as back-translation\index{Back-translation} are particularly effective for low-resource MT. \citet{lowres-triangular} proposed a scheme that could make use of translations from/into the source/target language into/from a third resource-rich language. The transfer-learning\index{Transfer learning} approach of \citet{lowres-transfer} first trains a {\em parent} model on a resource-rich language pair (e.g.\ French-English), and then continues training\index{Continued training} on the low-resource pair of interest (e.g.\ Uzbek-English). The effectiveness of transfer-learning depends on the relatedness of the languages~\citep{lowres-transfer,lowres-transfer2,lowres-transfer-relatedness,lowres-transfer-empirical}. The rapid adaptation\index{Adaptation} of multilingual NMT\index{Multilingual NMT} systems to new low-resource language pairs has been studied by~\citet{lowres-multiling-adapt}. Approaches that do not rely on resources from a third language include \citet{nmt-direction-ins-lowres} who supervised the generation order of an insertion-based\index{Insertion-based translation} low-resource translation model with word alignments.\index{Insertions} 

A series of NIST evaluation campaigns called LoReHLT~\citep{lowres-lorehlt}\index{LoReHLT} focuses on low-resource MT, and recent WMT\index{WMT} editions also contain low-resource language pairs~\citep{sys-wmt19,sys-wmt18,sys-wmt17}.

\subsection{Unsupervised NMT}
\label{sec:nmt-unsupervised}

Unsupervised\index{Unsupervised learning}\index{Unsupervised NMT|textbf} NMT is an extreme case of the low-resource scenario in which not even small amounts of cross-lingual data is available, and the translation system learns entirely from (unrelated) monolingual data.\index{Monolingual data} Unsupervised NMT often starts off from an unsupervised cross-lingual word embedding\index{Word embeddings!Cross-lingual word embeddings} model~\citep{nmt-unsupervised-crossling-embed-lample,nmt-unsupervised-crossling-embed-artetxe,nmt-unsupervised-crossling-embed-nonadversarial} that maps word embeddings from the source and the target language into a joint embedding space~\citep{nmt-unsupervised-lample,nmt-unsupervised-artetxe}. The translation model is then further refined by iterative back-translation~\citep{nmt-unsupervised-neural-phrase,nmt-unsupervised-smt-regularization}.\index{Back-translation} The extract-edit scheme of \citet{nmt-unsupervised-extract-edit} is an alternative to back-translation for unsupervised NMT that edits a sentence in the monolingual corpus rather than synthesize it from scratch. Unsupervised NMT has been targeted in recent WMT\index{WMT} evaluation campaigns~\citep{sys-wmt18,sys-wmt19}.

\section{Multilingual NMT}
\label{sec:nmt-multilingual}

NMT is usually trained to translate a single fixed source language into another fixed target language. Multilingual NMT\index{Multilingual NMT|textbf} aims to cover translation directions between multiple languages with a single model. This does not only have the potential of exploiting similarities across language pairs, it also reduces the number of systems required for all-way translation between a set of languages from quadratic to linear or even one. Multilingual NMT systems can be largely categorized by the components they share between language directions.\index{Parameter sharing} On one side of the spectrum, the entire neural architecture (both encoder and decoder) can be shared, and source and target languages can be specified by annotating sentences \citep{multiling-google-zeroshot} or words~\citep{multiling-kit-shared-encdec,multiling-effective-zero-shot} with language ID tags or embeddings. On the other side of the spectrum, \citet{multiling-multi-encdec} used a separate encoder for each source language and a separate decoder for each target language. \citet{multiling-shared-att,multiling-shared-att-csl} extended the work of \citet{multiling-multi-encdec} to attentional NMT by sharing the attention\index{Attention} mechanism across language directions. \citet{multiling-shared-enc} studied one-to-many translation with a single encoder but separate decoders for each target language.
A potential benefit of multilingual systems is zero-shot translation,\index{Zero-shot translation} i.e.\ the translation between two languages for which no direct training data is available.\footnote{The difference between zero-shot and unsupervised NMT (Sec.~\ref{sec:nmt-unsupervised}) is that unsupervised NMT does not rely on {\em any} cross-lingual data whereas zero-shot NMT uses cross-lingual data in other language directions.} \citet{multiling-google-zeroshot} reported reasonable Portuguese$\rightarrow$Spanish translation performance of their multilingual system that has been trained on Portuguese$\leftrightarrow$English and Spanish$\leftrightarrow$English, although pivoting through English (translate Spanish to English, and then English to Portuguese) worked better. Pivot-based\index{Pivot-based translation} zero-shot translation can be further improved by fine-tuning\index{Fine-tuning} on a pseudo parallel corpus~\citep{multiling-multisrc-ensembling} or by jointly training some components of the source-pivot and pivot-target systems like word embedding matrices~\citep{multiling-zero-shot-embed}.\index{Word embeddings} \citet{multiling-zero-shot-il} reported gains in zero-shot settings by adding a boldly named ``neural interlingual'' component between the encoder and the decoder which is shared\index{Parameter sharing} across language directions. For an assessment of the current capabilities of multilingual and zero-shot translation systems see~\citep{multiling-transformer-vs-recurrent,multiling-massive,sys-iwslt17}.
Another form of multilingual NMT is multi-source NMT~\citep{multiling-multisrc,smt-multi-source},\index{Multi-source NMT|textbf} in which the system tries to generate a single translation given sentences in two source languages simultaneously. A problem with this approach is data sparsity as missing source sentences have to be synthesized~\citep{lowres-multisrc,multiling-multisrc-data} if the training corpus does not provide sentences in all source languages. In a wider context, multi-source architectures can be used for multimodal NMT (Sec.~\ref{sec:multimodal-nmt}), morphological inflection~\citep{multiling-multisrc-inflection},\index{Morphology} zero-shot translation~\citep{multiling-multisrc-ensembling}, low-resource MT~\citep{lowres-multisrc}, syntax-based NMT~\citep{nmt-syntax-multisrc}, document-level MT~\citep{nmt-doc-eval}, or bidirectional decoding~\citep{nmt-direction-draft}.
\citet{multiling-survey} provide an overview of recent trends in multilingual NMT.

\section{NMT Model Size}
\label{sec:nmt-model-size}

NMT models usually have hundreds of millions of parameters (Tab.~\ref{tab:rnnsearch-params}). Such large models cause a number of practical issues. GPUs\index{GPU} are usually required to run such big models efficiently, but GPUs are expensive and their memory\index{Memory} is limited. Smaller models would not only reduce the computational complexity but could also make better use of GPU parallelism\index{Parallelization} by increasing batch sizes.\index{Batch size} Furthermore, model files require large amounts of disk space which is a problem on mobile platforms.
One way to increase the space efficiency of neural models is neural architecture search~\citep{nlm-transformer-aws,nn-architecture-search}.\index{Neural architecture search} For example, \citet{arch-search-evolved-transformer} found computationally efficient Transformer\index{Transformer} hyper-parameters by systematic neural architecture search. Rather than optimizing the dimensionality of layers, it is also possible to significantly speed up translation by departing from the usual 32 bit floating point arithmetics by reducing the precision\index{Precision} to 8 or 16 bits~\citep{precision-nmt-16bit,precision-nmt-16bit2,precision-nmt-8bit,precision-nmt-16bit-cpu} or by using vector quantization~\citep{production-gnmt,precision-cnn-quantized}.\index{Quantization}
The idea of pruning neural networks to improve the compactness of the models dates back almost 30 years~\citep{sparsify-obd}. The literature is therefore vast~\citep{sparsify-review}. One line of research aims to remove unimportant network connections. The connections can be selected for deletion based on the second-derivative of the training error with respect to the weight~\citep{sparsify-obd,sparsify-obs}, or by a threshold criterion on its magnitude~\citep{sparsify-threshold}. \citet{sparsify-nmt} confirmed a high degree of weight redundancy in NMT networks.\index{Over-parameterization} \citet{sparsify-prune-not-prune} demonstrated that large sparse models outperform smaller dense networks with the same memory footprint. \citet{sparsify-datafree} proposed to remove neurons which are very similar to another neuron and have small outgoing weights. \citet{unfolding} generalized their method to linear combinations of neurons. \citet{sparsify-noiseout} combined pairs of neurons with similar activities\index{Neuron activity} during training.
Using low rank matrices\index{Low rank matrix factorization} for neural network compression,\index{Compression} particularly approximations via Singular Value Decomposition (SVD),\index{Singular value decomposition} has been studied widely in the literature~\citep{sparsify-lowrank,sparsify-exploitlin,sparsify-svd-replace,sparsify-svd-replace2,sparsify-svd-asr}.
Another approach, known as {\em knowledge distillation},\index{Knowledge distillation|textbf} uses a large model (the teacher) to generate soft training labels for a smaller student network~\citep{kd-original,kd-hinton}. The student network is trained by minimizing the cross-entropy to the teacher. This idea has been applied to sequence modelling tasks such as machine translation and speech recognition~\citep{kd-asr,kd-nmt,kd-nmt-ensemble,kd-nmt-ana,kd-asr2,kd-speech2speech}.\index{Speech recognition}

\section{NMT with Extended Context}

\subsection{Multimodal NMT}
\label{sec:multimodal-nmt}

Machine translation is usually framed as the isolated transformation of the textual representation of a single sentence in one language into another. Since language is inherently ambiguous,\index{Ambiguity} researchers have searched for ways to provide the translation system with more context.\index{Context|textbf} For example, if the source sentence describes an image, the image itself potentially carries valuable clues to help the translation process. Multimodal machine translation~\citep{nmt-multimodal-org1,nmt-multimodal-org2}\index{Multimodal machine translation} aims to generate an image caption in the target language given both the source language caption and the image itself. The core of most multimodal MT models is a normal text-to-text system which integrates visual information by using global image features extracted with a separate computer vision\index{Computer vision} model~\citep{nmt-multimodal-org1,nmt-multimodal-org2} or via visual attention~\citep{nmt-multimodal-att}.\index{Attention} Multimodality in translation was the subject of a series of WMT\index{WMT} shared tasks~\citep{nmt-multimodal-wmt16,nmt-multimodal-wmt17,nmt-multimodal-wmt18}. \citet{nmt-multimodal-analysis} demonstrated the usefulness of visual clues in translation.

\subsection{Tree-based NMT}
\label{sec:tree-based-nmt}

The prevalent choice for modeling units in NMT are characters are subword-units\index{Subword units} (Sec.~\ref{sec:subword-nmt}). This design decision is not linguistically motivated but rather stems from the difficulty of extending NMT to an open vocabulary.\index{Vocabulary size}\index{Closed vocabulary} From the linguistic perspective, however, translation is better viewed as the transformation of larger elements in the sentence such as words, phrases, or even syntactic structures.

Various attempts have been made to introduce structures such as syntactic constituency trees\index{Trees!Syntactic constituency trees}\index{Syntax} or dependency trees\index{Trees!Dependency trees} both on the source and the target side of NMT. A popular approach is to retain the sequence-to-sequence architecture and linearize the tree structures,\index{Trees!Lineraized trees}\index{Linearization} for example using bracket expressions~\citep{syntax-grammar-foreign,nmt-syntax-multisrc,nmt-syntax-str2tree,syntax-det-att-parsing}, sequences of rules~\citep{danielle-syntax}, or CCG supertags~\citep{nmt-syntax-ccg}. \citet{nmt-syntax-packed-forest,nmt-syntax-packed-forest2} developed a linearization of a packed forests that represented multiple source sentence parses. \citet{danielle-syntax} reported gains by ensembling different linearization strategies of target-side syntax trees. Recurrent neural network grammars~\citep{syntax-rnng}\index{Grammars!Recurrent neural network grammars} that represent syntactic parse trees as sequence of actions were applied to machine translation by~\citet{syntax-rnng-nmt,syntax-rnng-nmt2}. Using actions to build target side tree structures is also central to the tree-based decoders of~\citet{syntax-treebased-dec,syntax-treebased-dec2}. \citet{nmt-syntax-speed} used syntax to speed up decoding by first predicting a parse tree, and then predicting all target tokens in parallel.
Tree-LSTMs~\citep{syntax-treelstm}\index{Tree-based NMT}\index{Tree-LSTMs} make it possible to represent a tree structure directly with the neural network architecture. They are a generalization of recurrent LSTM\index{LSTM} cells (Sec.~\ref{sec:nmt-att-recurrent}) that replaces the single input of a standard LSTM cell (usually from the previous time step) with multiple input connections, one from each child node. Thus, each Tree-LSTM cell represents a node in the tree, and the root node contains a fixed-length vector encoding of the whole tree structure. Tree-LSTMs have been applied to syntax-based NMT~\citep{syntax-treelstm-nmt,syntax-treelstm-nmt2,syntax-treelstm-nmt3}. An alternative to Tree-LSTMs was proposed by \citet{tree-ordered-neurons} who rearranged neurons in an LSTM network to resemble a block representation of the tree. \citet{nmt-syntax-conv-enc,nmt-syntax-conv-enc2} used convolutional encoders to represent a dependency graph in the source sentence. \citet{syntax-directed-att} biased encoder-decoder attention weights with syntactic clues. Unsupervised tree-based methods have been studied by~\citet{tree-unsupervised-rnng,tree-unsupervised-treelstm,tree-unsupervised-meaningful}.

\subsection{NMT with Graph Structured Input}

As a generalization of the tree-based approaches discussed in the previous section, lattice-based NMT\index{Lattice-based NMT} allows more general graph structures on the input side to provide a richer description of the source sentence. Lattices can represent uncertainty of upstream components such as speech recognizers~\citep{nmt-lattice-asr}\index{Speech recognition} or tokenizers~\citep{nmt-lattice-tok,nmt-lattice-tok-journal}.\index{Tokenization} Lattices have also been used to augment the input with external knowledge sources such as knowledge graphs~\citep{nmt-lattice-knoweldge-graph,nmt-lattice-knoweldge-graph2}\index{Knowledge graphs} or semantic predicate-argument structures~\citep{nmt-lattice-sem}.
Factors\index{Factored machine translation} are another way of providing more information to the translation system. Factors describe a word by a tuple consisting of its lemma and various linguistic information (prefix, suffix, part-of-speech etc.) rather than its surface form. This technique is popular for traditional statistical machine translation~\citep{pb-koehn,pb-factored},\index{Statistical machine translation} and has been applied to neural machine translation both on the input~\citep{nmt-factors-input} and the output~\citep{nmt-factors-output1,nmt-factors-output2} side.

\subsection{Document-level Translation}
\label{sec:document-level}

MT systems usually translate sentences in isolation. However, there is evidence that humans also take context into account, and rate translations from humans with access to the full document higher than the output of a state-of-the-art sentence-level machine translation system~\citep{nmt-doc-parity}. Common examples of ambiguity\index{Ambiguity} which can be resolved with cross-sentence context are pronoun prediction or coherency in lexical choice.\index{Lexical choice}

Various techniques have been proposed to provide the translation system with inter-sentential context,\index{Context!Document-level context}\index{Document-level NMT} for example by initializing encoder or decoder states~\citep{nmt-doc-init}, using multi-source\index{Multi-source NMT} encoders~\citep{nmt-doc-eval,nmt-doc-multisrc-src}, as additional decoder input~\citep{nmt-doc-init}, with memory-augmented\index{Memory} neural networks~\citep{nmt-doc-cache,nmt-doc-mem,nmt-doc-cache-trg}, a document-level LM~\citep{ucam-wmt19}, hierarchical attention~\citep{nmt-doc-han,nmt-doc-selective-att},\index{Attention} deliberation networks~\citep{nmt-doc-delib}, or by simply concatenating multiple source and/or target sentences~\citep{nmt-doc-concat,nmt-doc-eval}. Context-aware extensions to Transformer\index{Transformer} encoders have been proposed by~\citet{nmt-doc-transformer-enc1,nmt-doc-transformer-enc2}. Techniques also differ in whether they use source context only~\citep{nmt-doc-multisrc-src,nmt-doc-init,nmt-doc-transformer-enc1,nmt-doc-transformer-enc2,ucam-wmt19}, target context only~\citep{nmt-doc-cache,nmt-doc-cache-trg}, or both~\citep{nmt-doc-eval,nmt-doc-mem,nmt-doc-han,nmt-doc-concat,nmt-doc-selective-att}. Several studies on document-level NMT indicate that automatic and human sentence-level evaluation metrics often do not correlate well with improvements in discourse level phenomena~\citep{nmt-doc-eval,nmt-doc-parity,nmt-doc-test-set}.

\section{NMT-SMT Hybrid Systems}
\label{sec:nmt-smt-hybrids}

\begin{table}[t!]
\centering
\footnotesize
\begin{tabular}{p{0.4\textwidth}@{\hspace{0.1\textwidth}}p{0.4\textwidth}}
\toprule
Neural machine translation & Statistical machine translation\index{Statistical machine translation} \\
\midrule
\begin{itemize}[leftmargin=*]
    \item[$+$] Much better overall translation quality than SMT with enough training data~\citep{nmt-overview-six-challenges,nmt-vs-smt-9-langs,nmt-vs-smt-case-bentivogli1,nmt-vs-smt-case-bentivogli2,nmt-vs-smt-professional-translators,nmt-tool-marian,nmt-vs-smt-swiss-post}.
    \item[$+$] More fluent\index{Fluency} than SMT~\citep{nmt-vs-smt-case-bentivogli1,nmt-vs-smt-9-langs,nmt-vs-smt-professional-translators,nmt-vs-smt-bengali,nmt-vs-smt-new-sota}.
    \item[$+$] Better handles a variety of linguistic phenomena than SMT~\citep{nmt-vs-smt-case-bentivogli1,nmt-vs-smt-case-bentivogli2,nmt-vs-smt-challenge-set}.
    \item[$-$] Adequacy\index{Adequacy} issues due to lack of explicit coverage mechanism~\citep{nmt-coverage-tu,nmt-coverage-otem-utem,nmt-train-adequacy,nmt-vs-smt-bengali,nmt-vs-smt-new-sota}.
    \item[$-$] Lack of hypothesis diversity\index{Diversity} (Sec.~\ref{sec:diverse-decoding}).
    \item[$-$] Neural models perform not as well as specialized symbolic models on several monotone seq2seq tasks~\citep{nmt-vs-smt-monotone}.
\end{itemize} & \begin{itemize}[leftmargin=*]
    \item[$+$] Outperforms NMT in low-resource\index{Low-resource machine translation} scenarios~\citep{nmt-overview-six-challenges,nmt-vs-smt-dead,nmt-vs-smt-irish,nmt-vs-smt-basque,nmt-vs-smt-bengali,nmt-vs-smt-wat}.
    \item[$+$] Produces richer output lattices~\citep{sgnmt}\index{Lattices}.
    \item[$+$] More robust against noise
    \index{Noise}~\citep{nmt-vs-smt-asr,nmt-noise-impact}.
    \item[$+$] Translation quality degrades less on very long sentences\index{Sentence length} than NMT~\citep{nmt-vs-smt-9-langs,nmt-vs-smt-case-bentivogli1}.
    \item[$+$] Less errors in the translation of proper nouns~\citep{nmt-vs-smt-case-bentivogli2}.
    \item[$\circ$] NMT and SMT require comparable amounts of (document-level)\index{Context!Document-level context} post-editing\index{Post-editing}~\citep{nmt-vs-smt-postedit,nmt-vs-smt-professional-translators}.
\end{itemize} \\
\bottomrule
\end{tabular}
\caption{\label{tab:nmt-vs-smt} Summary of studies comparing traditional statistical machine translation and neural machine translation.}
\end{table}

Neural models were increasingly used as features in traditional SMT\index{Statistical machine translation} until NMT evolved as new paradigm. Without question, NMT has become the prevalent approach to machine translation in recent years. There is a large body of research comparing NMT and SMT (Tab.~\ref{tab:nmt-vs-smt}). Most studies have found superior overall translation quality of NMT models in most settings, but complementary strengths of both paradigms. Therefore, the literature about hybrid NMT-SMT systems is also vast. We distinguish between two categories of approaches for blending SMT and NMT. 

Approaches in the first category do not employ a full SMT system but borrow only key ideas or components from SMT to address specific issues in NMT. It is straight-forward to combine NMT scores with other features normally used in SMT (like language models)\index{Language models} in a log-linear model~\citep{nmt-mono-rnnlm,hybrid-nmt-with-smt-features}.
\footnote{Note that this is still different from using neural features in an SMT system as the standard left-to-right NMT decoder is used.}\index{Log-linear models} 
Conventional symbolic SMT-style lexical translation tables\index{Lexical translation probability}\index{Phrase table} can be incorporated into the NMT decoder by using the soft alignment weights of the standard NMT attention model\index{Attention}~\citep{hybrid-nmt-with-smt-features,hybrid-lex,hybrid-dict,sys-neubig-wat16,hybrid-external-phrase-mem}. \citet{nmt-align-ibm-like-attention} proposed to enhance the attention model in NMT by implementing basic concepts from the original word alignment models~\citep{word-ibm,word-hmm} like fertility\index{Fertility} and relative distortion\index{Distortion model}. 

The second category of hybrid systems is related to system combination. The idea is to combine a fully trained SMT system with an independently trained NMT system. Popular examples in this category are rescoring\index{Rescoring} and reranking\index{Reranking} methods~\citep{hybrid-nbest,sgnmt,nmt-adaptation-sgnmt,gec-jd-hybrid,hybrid-reranking-german,hybrid-smorgasbord,hybrid-pb-forced}, although these models may be too constraining\index{Constrained decoding} if the neural system is much stronger. \citet{edit-dist-wmt16} proposed a finite state transducer based loose combination scheme\index{Loose combination} that combines NMT and SMT translations via an edit distance based loss. The minimum Bayes risk (MBR) based approach of \citet{mbr-nmt} biases an unconstrained NMT decoder towards $n$-grams which are likely according the SMT system, and therefore also does not constrain the system to the SMT search space. MBR-based combination of NMT and SMT has been used in WMT evaluation systems~\citep{ucam-wmt18,ucam-wmt19} and in the industry~\citep{mbr-nmt-sdl}. NMT and SMT can also be combined in a cascade, with SMT providing the input to a post-processing\index{Post-processing} NMT system~\citep{hybrid-pre-translation,hybrid-nmt-multisource} or vice versa~\citep{hybrid-neural-pre-translation}.  \citet{hybrid-smt-advise,hybrid-smt-advise2} interpolated NMT posteriors with word recommendations from SMT and jointly trained NMT together with a gating function\index{Gated activation} which assigns the weight between SMT and NMT scores dynamically. The AMU-UEDIN submission to WMT16\index{WMT} let SMT take the lead and used NMT as a feature in phrase-based MT~\citep{sys-amu-wmt16}.\index{Phrase-based translation} In contrast, \citet{hybrid-technical-terms} translated most of the sentence with an NMT system, and just used SMT to translate technical terms in a post-processing\index{Post-processing} step. \citet{hybrid-nmt-smt-search} proposed a hybrid search algorithm in which the neural decoder expands hypotheses with phrases from an SMT system.\index{Phrase table} SMT can also be used as regularizer\index{Regularization} in unsupervised NMT~\citep{nmt-unsupervised-smt-regularization}.\index{Unsupervised NMT}

\section{Conclusion}

Neural machine translation (NMT) has become the de facto standard for large-scale machine translation in a very short period of time. This article traced back the origin of NMT to word and sentence embeddings and neural language models. We reviewed the most commonly used building blocks of NMT architectures -- recurrence, convolution, and attention -- and discussed popular concrete architectures such as RNNsearch, GNMT, ConvS2S, and the Transformer. We discussed the advantages and disadvantages of several important design choices that have to be made to design a good NMT system with respect to decoding, training, and segmentation. We then explored advanced topics in NMT research such as explainability and data sparsity.

\bibliography{references}

\end{document}